\documentclass[journal,twoside]{IEEEtran}
\usepackage{times}
\usepackage{epsfig}
\usepackage{graphicx}
\usepackage{amssymb}

\usepackage{tikz}
\usepackage[T1]{fontenc}
\usepackage[font=small,labelfont=bf,tableposition=top]{caption}
\usepackage[font=footnotesize]{subcaption}
\usepackage{color}
\usepackage{bm}
\usepackage{epstopdf}
\usepackage{multicol}
\usepackage{multirow}
\usepackage[numbers,sort&compress]{natbib}
\usepackage[section]{placeins}
\usepackage{url}
\usepackage{verbatim}
\usepackage{algorithm}
\usepackage{algpseudocode}
\usepackage{amsmath}
\usepackage{amssymb}
\usepackage{mathrsfs}

\usepackage{longtable}
\usepackage{rotating}
\usepackage{array}
\usepackage{multicol}
\usepackage{multirow}

\usepackage{arydshln} 

\ifCLASSINFOpdf
\else
\fi
\PassOptionsToPackage{hyphens}{url}\usepackage[pagebackref=true,breaklinks=true,letterpaper=true,bookmarks=false]{hyperref}

\DeclareMathOperator*{\argminA}{arg\,min} 

\begin{document}
%
\title{Multi-Resolution Multi-Modal Sensor Fusion \\ For Remote Sensing Data With Label Uncertainty}
%
%
%


\author{Xiaoxiao~Du,~\IEEEmembership{Member,~IEEE,}
        and~Alina~Zare,~\IEEEmembership{Senior Member,~IEEE}
\thanks{X. Du was with the Department of Electrical and Computer Engineering, University of Missouri, Columbia, MO 65211 and is currently with the Department of Naval Architecture and Marine Engineering, University of Michigan, Ann Arbor, MI 48109 USA. E-mail: xiaodu@umich.edu.}
\thanks{A. Zare is with the Department
of Electrical and Computer Engineering, University of Florida, Gainesville, FL 32611 USA. Email: azare@ece.ufl.edu.}
\thanks{This material is based upon work supported by the National Science Foundation under Grant IIS-1723891-CAREER: Supervised Learning for Incomplete and Uncertain Data.}
}


%
%


\markboth{IEEE TRANSACTIONS ON GEOSCIENCE AND REMOTE SENSING. Preprint Version. Accepted
November 2019}
{Du and Zare: Multi-Resolution Multi-Modal Sensor Fusion For Remote Sensing Data With Label Uncertainty} 
%



\maketitle

\begin{abstract}
In remote sensing, each sensor can provide complementary or reinforcing information. It is valuable to fuse outputs from multiple sensors to boost overall performance. Previous supervised fusion methods often require accurate labels for each pixel in the training data. However, in many remote sensing applications, pixel-level labels are difficult or infeasible to obtain. In addition, outputs from multiple sensors often have different resolution or modalities. For example, rasterized hyperspectral imagery presents data in a pixel grid while airborne Light Detection and Ranging (LiDAR) generates dense three-dimensional (3D) point clouds. It is often difficult to directly fuse such multi-modal, multi-resolution data. To address these challenges, we present a novel Multiple Instance Multi-Resolution Fusion (MIMRF) framework that can fuse multi-resolution and multi-modal sensor outputs while learning from automatically-generated, imprecisely-labeled data. Experiments were conducted on the MUUFL Gulfport hyperspectral and LiDAR data set and a remotely-sensed soybean and weed data set. Results show improved, consistent performance on scene understanding and agricultural applications when compared to
traditional fusion methods.
\end{abstract}

\begin{IEEEkeywords}
sensor fusion, multi-resolution, multi-modal fusion, multiple instance learning, scene understanding, label uncertainty, hyperspectral, LiDAR, Choquet integral
\end{IEEEkeywords}

%
\IEEEpeerreviewmaketitle

\section{Introduction and Motivation}
\label{sec:intro}

\begin{figure}[h]
\centering
\begin{subfigure}[h]{0.45\linewidth}
\centering
\includegraphics[width=\textwidth,trim={46mm 15mm 46mm 12mm},clip]{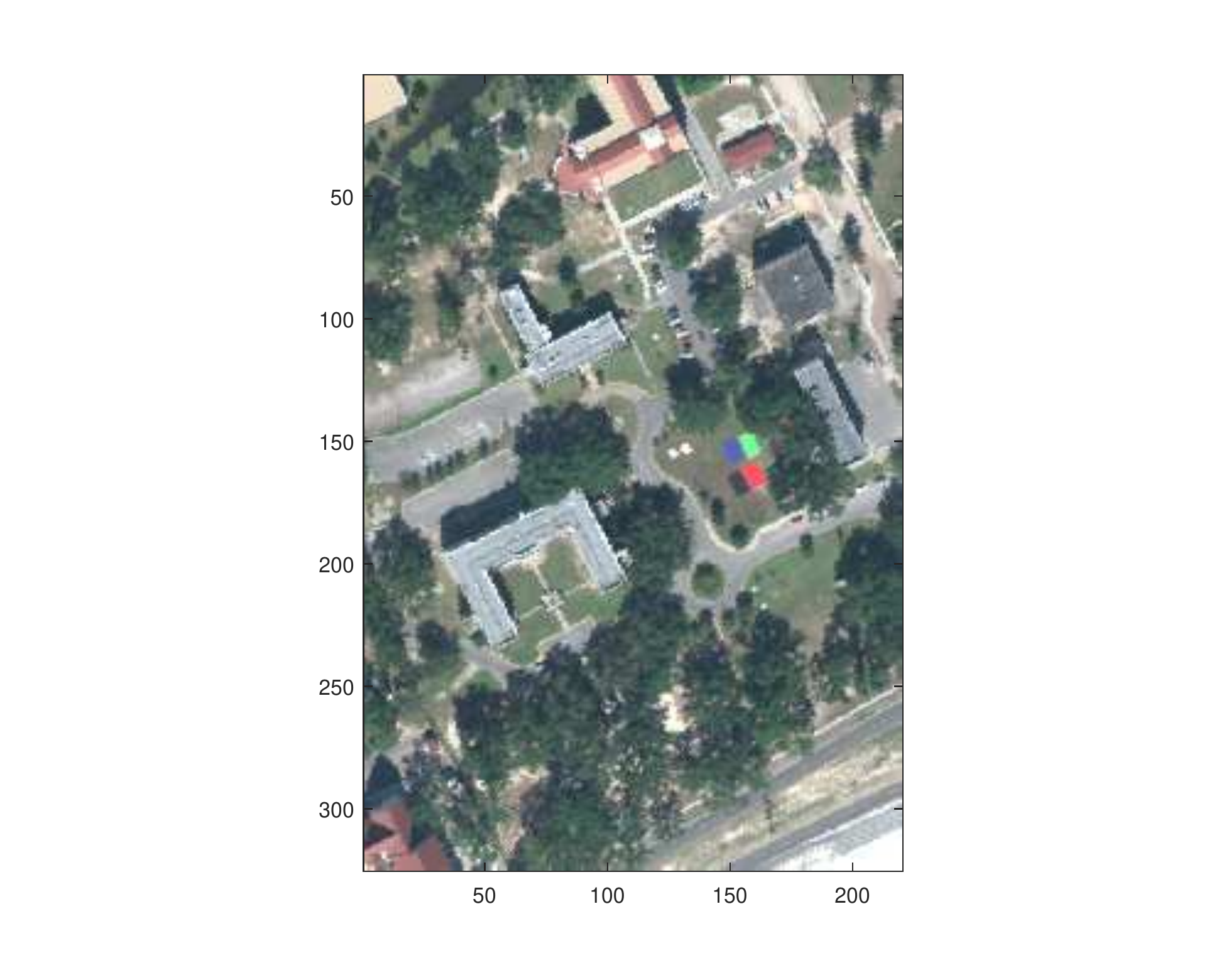}
\caption{}
\label{fig:muuflRGBcampus1_220}
\end{subfigure}
\begin{subfigure}[h]{0.45\linewidth}
\centering
\includegraphics[width=\textwidth,trim={46mm 15mm 46mm 12mm},clip]{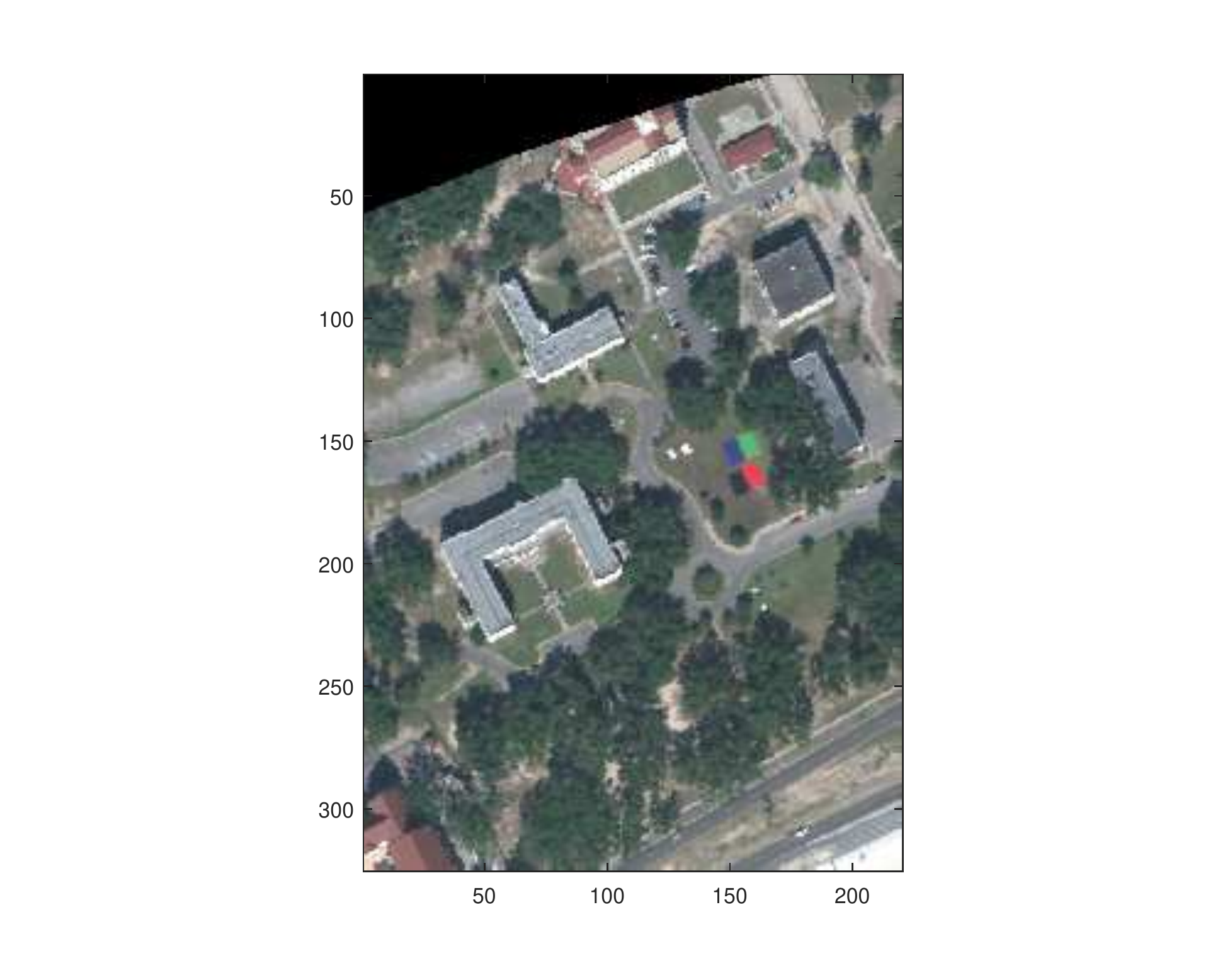}
\caption{}
\label{fig:muuflRGBcampus2_220}
\end{subfigure}
\caption{RGB images of the MUUFL Gulfport data. (a) Campus 1; (b) Campus 2.  The black region in the top left corner of (b) is an invalid region from the data collection and was excluded from any training or testing process in the experiment. The ``campus 1'' and ``campus 2'' refer to data from two flights (more details please see Section~\ref{sec:experiments_muufl}). }
\label{fig:muuflRGBcampus_220}
\end{figure}

\begin{figure}[h]
\centering
\includegraphics[width=0.5\textwidth, height= 65 mm, trim={15mm 5mm 15mm 25mm},clip]{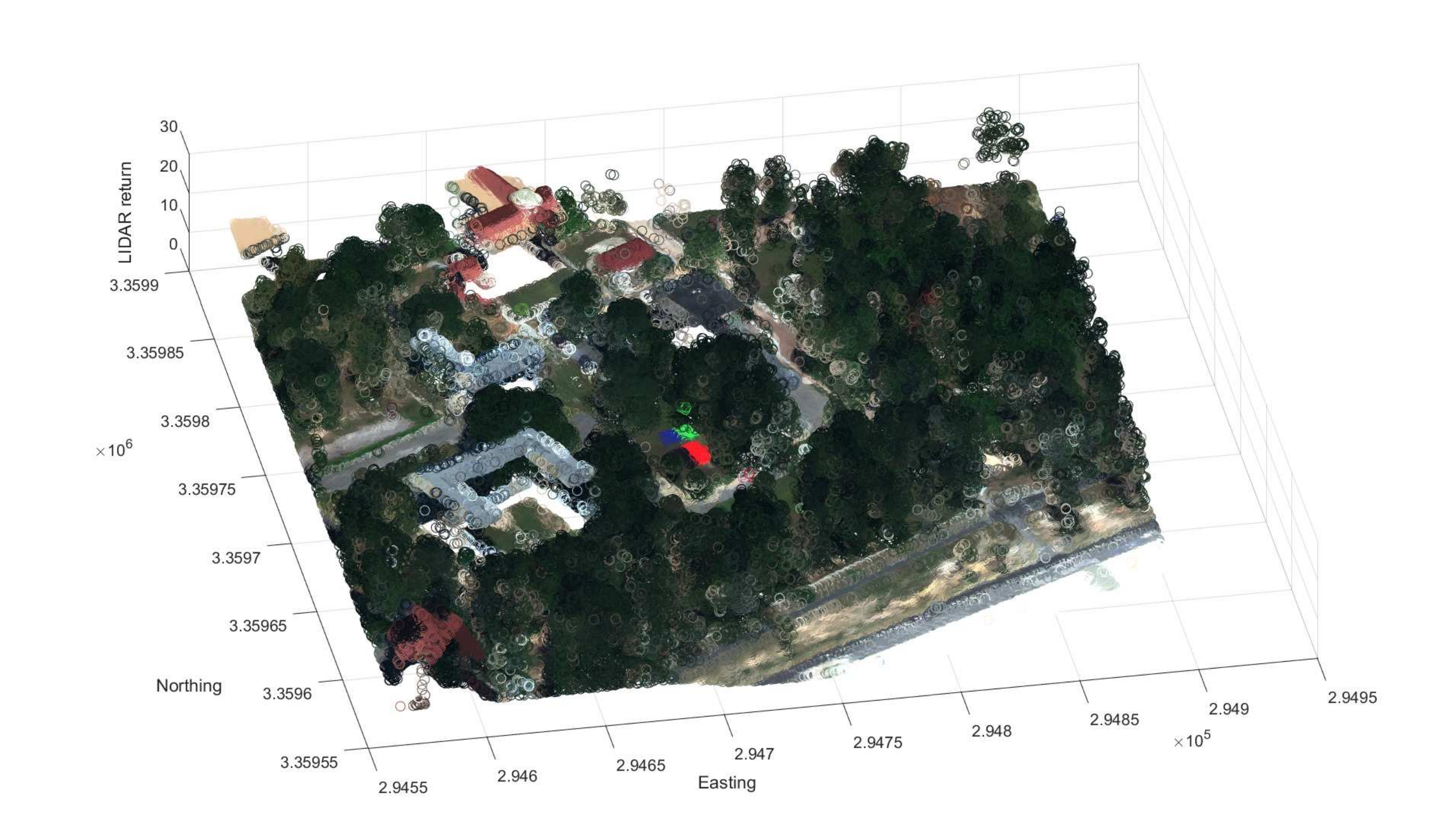}
\caption{An example of 3D scatterplot of LiDAR data over the University of Southern Mississippi - Gulfpark campus. The LiDAR points were colored by the RGB imagery provided by HSI sensors over the scene.  ``X'' and ``Y''-axes are Easting and Northing locations and ``Z''-axis corresponds to the elevation information. }
\label{fig:lidar_full1}
\end{figure}

In multi-sensor fusion, each sensor may provide complementary and reinforcing information that can be helpful in target detection, classification, and scene understanding \cite{gader2004multi}. It is useful to integrate and combine such outputs from multiple sensors to provide more accurate information over a scene.  In this study, hyperspectral (HSI) and LiDAR data collected by us over the University of Southern Mississippi-Gulfpark campus were used \cite{gader2013muufl, du2017technical}. Figure~\ref{fig:muuflRGBcampus_220} shows an RGB image over the scene. Figure~\ref{fig:lidar_full1} shows a scatter plot of the LiDAR point cloud data. As shown, there are a wide variety of materials in the scene, such as buildings, road, etc. The task is to use the HSI and LiDAR data we collected to classify materials in the scene. 

There are several advantages to fuse both HSI and LiDAR data rather than using individual HSI and LiDAR information for this task. If a road and a building rooftop are built with the same material (say, asphalt), hyperspectral information alone may not be sufficient to tell the rooftop and the road apart. However, the LiDAR data provides elevation information and can easily distinguish the two. On the other hand, a highway and a biking trail can be at the same elevation and using LiDAR data alone may not be sufficient to distinguish the two types of roads, but hyperspectral sensor can be very helpful in identifying the distinctive spectral characteristics between a highway, which is likely to be primarily asphalt, and a biking trail, which is likely to be covered in dirt. It would be, thus, valuable to fuse information from both HSI and LiDAR sensors to obtain a better classification result and a more comprehensive understanding of a scene \cite{pohl1998review, liggins2008handbook, zhang2010multi, bioucas2013hyperspectral}.

There are four challenges associated with this fusion task. First, the HSI and LiDAR data we have are of different modalities. The HSI imagery is on a pixel grid and the raw LiDAR data is a point cloud. Many previous HSI and LiDAR fusion methods have been proposed in the literature \cite{li2010extracting, khodadadzadeh2015fusion, rasti2017fusion, luo2017fusion}. However, nearly all of these fusion methods for HSI and LiDAR rely on having a rasterized LiDAR image with accurate image registration. In these methods, the LiDAR point cloud data was mapped in the same grid as the hyperspectral imagery and the fusion methods work with both HSI and LiDAR data in image form. However, the raw dense LiDAR point cloud can offer higher geographic accuracy \cite{sampath2010segmentation} as the raw data does not depend on grid size. Besides, image registration and rasterization may introduce additional inaccuracy \cite{cao2014astable, brigot2016adaptation}. It would be valuable to develop a multi-modal fusion algorithm that can handle directly the HSI imagery and raw LiDAR point cloud.

Second, the HSI and LiDAR data have different resolutions. The hyperspectral imagery we collected have a ground sample distance of $1m$. That is to say, each pixel in the hyperspectral imagery covers $1m^2$ area. The LiDAR data, on the other hand, have a higher resolution with 0.60m cross track and 0.78m along track spot spacing. There is more than one LiDAR data point inside a $1m^2$ area. This means, for each pixel in the hyperspectral imagery, there are more than one LiDAR data point to which it corresponds. It would be, thus, important to develop a fusion method that can handle multi-resolution outputs from multi-modal sensors.


Third, there are inaccuracies of LiDAR measurements in the given data set after rasterization. For example, if we look at building edges on the grey roofs, some of the building points have similar elevation as the neighborhood grass pixels on the ground, which is obviously wrong. Some points on the sidewalks and dirt roads, on the other hand, have high elevation values similar to nearby trees and buildings, which is also wrong. Such incorrect information can be caused by a variety of factors such as poor alignment and rasterization to HSI grid, sensor inaccuracy, or missed edge points from the laser pulse discontinuousness in LiDAR \cite{li2008adaptive}.  Directly using such inaccurate measurements for fusion can cause further inaccuracy or error in classification, detection, or prediction. Therefore, we must develop a fusion algorithm that is able to handle such inaccurate/imprecise measurements.

Finally, to train a supervised fusion algorithm, training labels are needed. However, in this data set, accurate pixel-level labels for both HSI and LiDAR data are not available. There are more than 70,000 pixels in one HSI image and more than 170,000 data points in one LiDAR point cloud. It would be expensive and difficult for humans to manually label each pixel/data point. Even if we have experts manually label each pixel or data point in the scene, there are certain transition areas that are impossible to label. For example, as mentioned above, a pixel in the HSI image corresponds to $1m^2$. It is highly likely that a pixel at the edge of the building is mixed with both building and ground surface materials. In this case, it is impossible to provide accurate labels for each pixel/data point in the training data. However, in this example, one can easily circle a region in the scene that contains a building. That is to say, it is possible to identify a region or a set of pixels that contains the target material in the scene. We call such a region or a set of pixels a ``bag'', based on the Multiple Instance Learning (MIL) framework \cite{dietterich1997solving}. Previous supervised fusion methods often require accurate pixel-level labels and cannot handle ``bag-level'' labels \cite{yang2010decision, dalponte2008fusion}. We aim to develop a trained fusion algorithm that can handle such uncertainty in training labels. 


This paper proposes a Multiple Instance Multi-Resolution Fusion (MIMRF) framework to address the above four challenges. The proposed MIMRF can fuse multi-resolution and multi-modal sensor outputs from multiple sensors while effectively learning from bag-level training labels.

\section{Related Work}  
\label{sec:relatedwork}
The proposed MIMRF algorithm handles label uncertainty in training data by formulating the supervised fusion problem under the Multiple Instance Learning (MIL) framework. The MIMRF is based on the Choquet integral, an effective nonlinear fusion tool widely used in the literature. In this section, related work on multi-sensor fusion and the MIL framework is described. The basis of the proposed MIMRF fusion method, the Choquet integral, is also described in this section.
 
\subsection{Multi-Resolution and Multi-Modal Sensor Fusion}  
\label{sec:relatedworkmultifusion}
Existing optical sensors produce data at varying spatial, temporal and spectral resolutions \cite{shen2016an}. Those sensor outputs may also have different modalities, such as imagery and point clouds. Multi-resolution and multi-modal fusion methods have been studied and developed in the literature in order to better fuse such information from different sensors. Multi-sensor fusion has wide applications in remote sensing such as the extraction of urban road networks \cite{hinz2003automatic}, building detection \cite{stankov2014detection}, precision agriculture \cite{king2000challenge}, and anomaly detection in archaeology \cite{lin2011combining}.

Nearly all previous multi-resolution fusion methods focus on fusing image data only. Pan-sharpening methods, for example, use a panchromatic image with higer spatial resolution to fuse with multi- or hyper-spectral images in order to obtain images with higher spectral and spatial resolution. However, methods like pan-sharpening only handles imagery and mostly focus on fusing only two images (a panchromatic and a multispectral or hyperspectral image) \cite{ehlers2010multi, shi2015learning, liu2016spatial, du2007performance,mckeown1999fusion, licciardi2012fusion,gomez2001wavelet, chen2014fusion}.

More specifically, regarding HSI and LiDAR fusion, previous fusion methods can only work with pre-rasterized LiDAR images \cite{li2010extracting, khodadadzadeh2015fusion, rasti2017fusion, luo2017fusion}. In these methods, the LiDAR point cloud data was mapped in the same grid as the hyperspectral imagery and the fusion methods work with both HSI and LiDAR data in image form. However, as discussed in Introduction, the raw dense LiDAR point cloud can offer higher geographic accuracy and alleviate the necessity to geometrically align the HSI and LiDAR data \cite{de2017fusion, maddern2016real}. 

\subsection{Multiple Instance Learning}  
\label{sec:relatedworkmiclass}

As discussed in Introduction, the proposed MIMRF algorithm aims to handle bag-level training labels rather than pixel-level labels. Here, this label uncertainty problem is formulated under the Multiple Instance Learning (MIL) framework. 

The MIL framework was first proposed in \cite{dietterich1997solving} to deal with uncertainties in training labels. In the MIL framework, training labels are associated with sets of data points (``bags'') instead of each data point (``instance''). In two-class classification/target detection applications, the standard MIL assumes that a bag is labeled positive if at least one instance in the bag is positive (target) and a bag is labeled negative if all the instances in the bag are negative (non-target). The MIL has wide applications in natural scene classification \cite{maron1998multipleicml, zhou2006multi}, human action recognition in videos \cite{ali2010human}, object detection and tracking \cite{zhang2005multiple, babenko2009visual, babenko2011robust}, context identification, and context-dependent learning in remote sensing data \cite{du2015possibilistic, torrione2009multiple}.

The mi-SVM algorithm is a widely cited MIL method for classification \cite{xu2017weakly, zare2017discriminative, cao2017weakly} and will later be used as a comparison method in the experiments. The mi-SVM algorithm was proposed by Andrews et al.  \cite{andrews2002support} as an MIL extension to support vector machine (SVM) learning approaches. The mi-SVM algorithm can work with bag-level labels and learns a linear discriminate function to separate the positive from the negative classes.

\subsection{Fuzzy Measure and Choquet Integral}  
\label{sec:relatedworkci}
The proposed MIMRF algorithm uses the Choquet integral (CI) \cite{choquet1954theory} to perform fusion. The CI is a powerful non-linear fusion and information aggregation framework and has wide applications in the literature  \cite{grabisch1996application, grabisch1995anew, labreuche2003thechoquet, mendezvazquez2008minimum, mendezvazquez2008learning}. In the field of remote sensing, the CI has been applied to multi-sensor fusion in landmine detection \cite{gader2004multi, gader2001recognition}, classifier fusion \cite{du2016multiple}, and target detection using Landsat  and hyperspectral imagery \cite{wang2015integration, du2016multiple} . 

Compared with commonly used aggregation operators such as weighted arithmetic means \cite{fodor1995characterization}, the CI is able to model complex, non-linear relationship amongst the combinations of the sources for fusion.  The Choquet integral depends on ``fuzzy measures'', whose values determine the fusion outcome. Depending on the set of real-valued fuzzy measure $\mathbf{g}$ it learns, the CI can flexibly represent a wide variety of aggregation operators \cite{maron1998phdthesis, marichal2000an, labreuche2003thechoquet, gader2004multi, mendezvazquez2008minimum, du2016multiple}. 

Suppose we are fusing $m$ sources using the discrete Choquet integral. \textcolor{black}{These fusion sources can come from sensory data (data-level fusion) or confidence values from detectors or classifiers on the raw sensor data (decision-level fusion). In the case where a sensor has multidimensional outputs, such as a 3D hyperspectral data cube, each band can be regarded as a ``sensor''.  In this paper, the term ``sensor outputs'' can refer to either sensor data or classifier outputs, and is equivalent with ``fusion sources.'' } Let $S=\left\{s_1,s_2,\dots,s_m\right\}$ denote the $m$ sources to be fused. The power set of all (crisp) subsets of $S$ is denoted by $2^S$.  A monotonic and normalized fuzzy measure, $\mathbf{g}$, is a real valued function that maps $2^S \rightarrow [0,1]$ \cite{choquet1954theory,Sugeno74, fitting2003beyond, du2016multiple}.  The fuzzy measure used in this paper satisfies monotonicity and normalization properties, i.e. $g(A) \leq g(B)$ if $A \subseteq B$ and $A, B \subseteq C$; $g(\emptyset) = 0$; and $g(S) = 1$ \cite{keller2016fundamentals}. Therefore, the power set has $2^m - 1$ non-empty crisp subsets, and each element in the fuzzy measure corresponds to one of the subset. In this paper,  the fuzzy measure $\mathbf{g}$ can be written as a vector of length $(2^m-1)$, where the last measure element that correspond to the full set $g(S) \equiv 1$, and the remaining $(2^m-2)$ element values are unknown and to be learned. Let $h(s_k; \mathbf{x}_n)$ denote the (known) $k^{th}$ sensor output for the $n^{th}$ data point.  \textcolor{black}{If data-level fusion is performed, the function $h$ is an identity function and $h(s_k; \mathbf{x}_n)$ simply means the $n^{th}$ data point from the $k^{th}$ sensor. If decision-level fusion is performed, the function $h$ can represent the detector or classifier used to produce a decision-level output, and the term $h(s_k; \mathbf{x}_n)$ represents the $n^{th}$ value from the $k^{th}$ classifier output or detector map to be fused. Note that each $h$ here is assumed to have higher confidence values for target and lower confidence values for non-target/background data points, since we later assign label ``1'' to the target (positive) points and label ``0'' to non-target (negative) points.} The discrete Choquet integral on instance $\mathbf{x}_n$ given sensor outputs $S$ is then computed as \cite{mendezvazquez2008info, keller2016fundamentals, du2016multiple}
\begin{equation}
C_\mathbf{g}(\mathbf{x}_n) = \sum_{k=1}^{m}\left[ h(s_k; \mathbf{x}_n) - h(s_{k+1}; \mathbf{x}_n)\right]  g(A_k),
\label{eq:cg}
\end{equation}
%
where $S$ is sorted so that $h(s_1; \mathbf{x}_n)\geq h(s_2; \mathbf{x}_n) \geq \dots \geq h(s_m; \mathbf{x}_n)$ and  $h(s_{m+1}; \mathbf{x}_n) \equiv 0$. \textcolor{black}{Note that the order of the sorting can be different for each $\mathbf{x}_n$.} The term  $g(A_k)$ is the fuzzy measure element that corresponds to the subset $A_k = \left\{ s_1, \dots, s_k\right\}$.  \textcolor{black}{In this paper, we normalize all $h(s_k; \mathbf{x}_n)$ values to be between 0 and 1 for each sensor output, and the $g(A_k)$ used in this paper is bounded between 0 and 1, thus the Choquet integral fusion values $C_\mathbf{g}(\mathbf{x}_n)$ in this paper are bounded between 0 and 1 for all $\mathbf{x}_n$. } 

In the supervised fusion problem, we need to learn the $(2^m-2)$ unknown fuzzy measure element values in $\mathbf{g}$, given known training data (the $h(s_\cdot; \mathbf{x}_\cdot)$ terms) and training labels. \textcolor{black}{The  $(2^m-2)$ measure elements represent the non-linear interactions and interdependencies between all subsets of the fusion sources in the aggregation process. When the number of measure elements reduces to $m$ for $m$ sources, the CI is essentially reduced to a simple linear weighted average.} The learned fuzzy measure $\mathbf{g}$ can then be used to determine the fusion results for test data. In the literature, the fuzzy measure values can be learned using either quadratic programming or sampling techniques. The CI-QP (quadratic programming) approach \cite{keller2016fundamentals} learns a fuzzy measure for Choquet integral by optimizing a least squares error objective using quadratic programming \cite{nocedal2006numerical}. However, the CI-QP method requires pixel-level training labels and cannot work for MIL problems. We previously proposed a Multiple Instance Choquet Integral (MICI) classifier fusion framework that extends the standard CI fusion under the MIL framework  \cite{du2016multiple, du2017multiple, du2018multiple}. However, the previous MICI models still assumes that each sensor sources must have the same number of data points and do not support multi-resolution fusion. The goal of this work is to develop a novel trained classifier fusion algorithm that can both handle multi-resolution data and learn from uncertain and imprecise training labels.

\section{The Multiple Instance Multi-Resolution Fusion (MIMRF) Algorithm}
\label{sec:multiressec}
In this section, the proposed Multiple Instance Multi-Resolution Fusion (MIMRF) algorithm is described.  The proposed MIMRF algorithm learns a monotonic, normalized fuzzy measure from training data and bag-level training labels. Then, the learned fuzzy measure is used  with the Choquet integral to perform multi-resolution fusion. 

\subsection{Objective Function}
\label{sec:multiresobjmotiv}

We discussed in the Introduction that in the HSI/LiDAR fusion problem,  due to differences in modality and resolution, there can be more than one LiDAR data point that correspond to each pixel in the HSI imagery. There may also be inaccuracies in the LiDAR data.  Our proposed MIMRF algorithm aims to handle such challenges. We motivate our objective function with a simple example, as follows. 

\begin{figure}[h]
\centering
\includegraphics[width=0.3\textwidth,trim={0mm 0mm 0mm 5mm},clip]{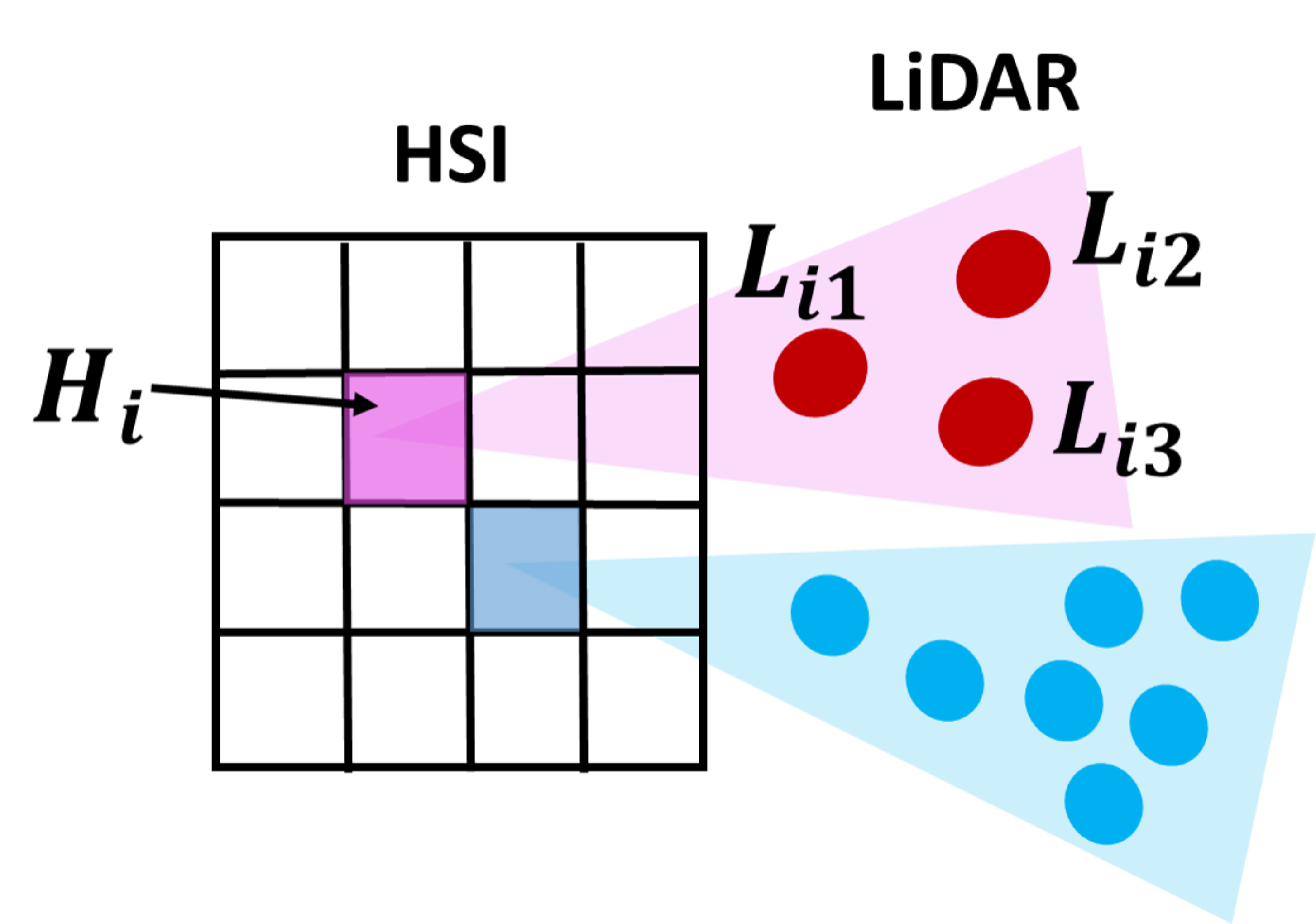}
\caption{Illustration for HSI and LiDAR fusion. All LiDAR data points in the pink shade fell in the same area covered by the pink pixel in the HSI image, and  all LiDAR data points in the blue shade corresponds to the blue pixel in the HSI image. Note that there can be more than one LiDAR points in the area covered by one pixel in the HSI imagery.}
\label{fig:hsi_lidar_illustration1}
\end{figure}

Suppose there are three LiDAR data points that correspond to one pixel in the HSI imagery, as shaded pink in Figure~\ref{fig:hsi_lidar_illustration1}. Let $H_i$ denote the value of the $i^{th}$ pixel in the HSI imagery.  Let $L_{i1}, L_{i2}, L_{i3}$ denote the values of the three LiDAR points. \textcolor{black}{Let $\mathbf{S}_{i}$ denote the set of all possible matching combinations of the sensor outputs for pixel $i$, i.e.,}
\begin{equation}
\textcolor{black}{\mathbf{S}_{i} = \{\{H_i,L_{i1}\},\{H_i,L_{i2}\}, \{H_i,L_{i3}\}\}.}
\label{eq:si}
\end{equation}

%

\textcolor{black}{Our objective function contains two levels of aggregation to solve the aforementioned challenges. We know from the above example that there can be more than one LiDAR data points that correspond to each pixel in the HSI imagery. Here we assume that at least one of the LiDAR points is accurate (has the correct height information), but we do not know which one. Other LiDAR points may carry erroneous or inaccurate information (for example, on the edge of a building). The first level of aggregation was designed to help automatically select the correct LiDAR point.} For the $i^{th}$ negative data point in $\mathbf{S}_{i}$, we write the CI fusion as
%
\begin{equation}
 C_\mathbf{g}(\mathbf{S}_{i}^-)= \min_{ \mathbf{x}_{k}^- \in \mathbf{S}_{i}^- }    C_\mathbf{g}(\mathbf{x}_{k}^- ),
\label{eq:multiresobjneginner}
\end{equation}
%
and for the $j^{th}$ positive data point in $\mathbf{S}_{j}$, we write the CI fusion as
\begin{equation}
 C_\mathbf{g}(\mathbf{S}_{j}^+) = \max_{ \mathbf{x}_{l}^+  \in \mathbf{S}_{j}^+} C_\mathbf{g}(\mathbf{x}_{l}^+),
\label{eq:multiresobjposinner}
\end{equation}
%
where $\mathbf{S}_{i}^-$ is the set of sensor outputs for the $i^{th}$ negative data point and $\mathbf{S}_{j}^+$ is the set of sensor outputs for the $j^{th}$ positive data point as defined in \eqref{eq:si}; $C_\mathbf{g}(\mathbf{S}_{i}^-)$ is the Choquet integral output for $\mathbf{S}_{i}^-$ and $ C_\mathbf{g}(\mathbf{S}_{j}^+)$ is the Choquet integral output for $\mathbf{S}_{j}^+$. \textcolor{black}{In this way, the min and max operators automatically identify and select one most favorable element in terms of CI from the set $\mathbf{S}_i$, which is assumed to be the data point with correct information corresponding to pixel $i$.}

\textcolor{black}{The second level of aggregation was designed to handle bag-level training labels.} We set the desired label for a ``negative'' (non-target) data point to be ``0'' and the desired label for a ``positive'' (target) data point to be ``1'', for two-class classification problems. Recall that the MIL assumes a bag is labeled positive if at least one instance in the bag is positive and a bag is labeled negative if all the instances in the bag are negative. That is to say, we wish to encourage all points in negative bags to have label ``0'' and at least one point in positive bags to have label ``1''. We can write the objective function for negative bags as
\begin{equation}
\begin{split}
J^-
&=  \sum_{a=1}^{B^-} \max_{ \mathbf{S}_{ai}^-  \in \mathscr{B}_a^-} \left(  C_\mathbf{g}(\mathbf{S}_{ai}^-) - 0 \right)^2   ,
\end{split}
\label{eq:multiresobjneg}
\end{equation}
%
Similarly, the objective function for positive bags can be written as
\begin{equation}
\begin{split}
J^+ 
&=\sum_{b=1}^{B^+} \min_{ \mathbf{S}_{bj}^+ \in \mathscr{B}_b^+} \left(  C_\mathbf{g}(\mathbf{S}_{bj}^+)  -1\right)^2 ,
\end{split}
\label{eq:multiresobjpos}
\end{equation}
%
where $B^+$ is the total number of positive bags, $B^-$ is the total number of negative bags, $\mathbf{S}_{ai}^-$ is the set that corresponds to the $i^{th}$ instance in the $a^{th}$ negative bag, $\mathbf{S}_{bj}^+$ is the set that corresponds to  the $j^{th}$ instance in $b^{th}$ positive bag, $C_\mathbf{g}$ is the Choquet integral given fuzzy measure $\mathbf{g}$, $ \mathscr{B}_a^-$ is the $a^{th}$ negative bag, and $\mathscr{B}_b^+$ is the $b^{th}$ positive bag. By minimizing $J^-$ in \eqref{eq:multiresobjneg}, we encourage the fusion output of all the points in the the negative bag to the desired negative label of $0$. By minimizing $J^+$ in \eqref{eq:multiresobjpos}, we encourage the fusion output of at least one of the points in the positive bag to the desired positive label of $1$. This satisfies the MIL assumption and successfully handles label uncertainty. \textcolor{black}{Although this paper focuses on binary classification, the MIL-based formulation can be extended to regression problems where the ground truth labels are of real values \cite{du2018multiple}. The least squares notation that we used above can be generalized to $( C_\mathbf{g}(\mathbf{S}) - d)^2$, where $d$ is the desired binary or real label value. As a side note,  when $d$ is binary (0 or 1), both  \eqref{eq:multiresobjneg} and \eqref{eq:multiresobjpos} could be simplified to $J=\sum_{B} \max_{ \mathbf{S} \in \mathscr{B}} C_\mathbf{g}(\mathbf{S})$. }



 
\textcolor{black}{Thus, the objective function for the proposed Multiple Instance Multi-Resolution Fusion (MIMRF) algorithm is written as 
%
%
\begin{equation}
\begin{split}
\min_\mathbf{g} J &= J^- + J^+  \\
&= \sum_{a=1}^{B^-} \max_{ \mathbf{S}_{ai}^-  \in \mathscr{B}_a^-} \left( \boxed{\min_{ \mathbf{x}_{k}^- \in \mathbf{S}_{ai}^- }    C_\mathbf{g}(\mathbf{x}_{k}^- ) }- 0 \right)^2 \\
&+  \sum_{b=1}^{B^+} \min_{ \mathbf{S}_{bj}^+  \in \mathscr{B}_b^+} \left( \boxed{\max_{ \mathbf{x}_{l}^+  \in \mathbf{S}_{bj}^+} C_\mathbf{g}(\mathbf{x}_{l}^+)}-1\right)^2, 
\end{split}
\label{eq:objlnmultiresminmax}
\end{equation}
where the boxed terms are from the first level of aggregation in \eqref{eq:multiresobjneginner} and \eqref{eq:multiresobjposinner}, and the max and min operators outside the parentheses are from the second level of aggregation in \eqref{eq:multiresobjneg} and \eqref{eq:multiresobjpos}.}




\subsection{Optimization}
\label{sec:optimization}
An evolutionary algorithm (EA) is used to optimize the objective function in \eqref{eq:objlnmultiresminmax} and learn the fuzzy measure $\mathbf{g}$. The evolutionary algorithm used in this paper extends upon the method used in \cite{du2016multiple}. 
First, a population of fuzzy measures was generated randomly between $[0, 1]$ that satisfies monotonicity. Let $P$ denote the measure population size (i.e., how many measures are in the population in one generation). If $P$ is small, there are fewer possibilities to update measure values and only a small part of search space is explored. If $P$ is large, the optimization will slow down given the larger search space and more mutation possibilities. We used $P = 30$ in our experiments as it is a good balance between reasonable computation time and enough population to mutate and converge.


Then, in each iteration, new measure values were updated based on small-scale or large-scale mutations. To update the measure values, the valid interval of each fuzzy measure element in the population was computed. The ``valid interval'' of a fuzzy measure element is defined as how much the measure element value can change while still satisfying the monotonicity property \cite{du2016multiple}. In small-scale mutation, only one measure element value was updated according to their valid interval. In large-scale mutation, all the measure elements were updated. A Truncated Gaussian distribution \citep{johnson1994continuous} was used for sampling new measure values between the lower and upper bound of the valid interval  \textcolor{black}{ since the full Gaussian may get invalid values for the measure elements. In our experiments, the variance of the Truncated Gaussian is set to 0.1 and the mean is equal to the previous measure element value before updating. The lower and upper bounds of the Truncated Gaussian are equal to the lower and upper bounds of the valid intervals of the measure element to be updated. This way, the updated value was bounded within the valid intervals and close to the previous measure element value. We conducted an experiment using our demo MIMRF code\footnote{Available at \url{https://github.com/GatorSense/MIMRF} \cite{xiaoxiao_du_2019_2638382}.} and compared using uniform distribution versus Truncated Gaussian, and using uniform distribution took much longer to converge ($\sim$490 iterations) than the Truncated Gaussian distribution ($\sim$80 iterations). This makes sense as the uniform distribution essentially conducts a random search at every mutation yet the Truncated Gaussian can ``tweak'' the updated measure element values in small scale and thus make the search more efficient.} Denote $\lambda \in [0,1]$ as the rate of small-scale mutation (i.e., how often to sample a single measure element) and $1-\lambda$ is the rate of large-scale mutation (i.e., how often to sample the entire measure). If $\lambda$ is too small, the entire measure will be sampled most of the time and the optimization will be close to a random search. If $\lambda$ is too large, it is likely that the EA will fall into a local minima. We use $\lambda=0.8$ in our experiments so that a new measure element can be updated most of the time, but occasionally the entire measure was updated to expand the search space and avoid local minima.

\begin{algorithm}[t!]
\caption{MIMRF Optimization} 
\label{alg:mici}
\begin{algorithmic}[1] 
\Statex  \textbf{Model Learning Stage}
\Require Training Bags $\mathscr{B}$, Bag-level Labels, Parameters 
\State Initialize a population of fuzzy measures, $\mathbf{\mathscr{W}}$.
\State Compute objective values $\mathbf{J}_\mathbf{\mathscr{W}}^0$.
\State $J^* = min(\mathbf{J}_\mathbf{\mathscr{W}}^0)$, $\mathbf{g}^* = \argminA_{\mathbf{\mathscr{W}}} J^* $.
\While { $t<I$ }   
		\For{ $p:=1 \to P$ }
			\State Evaluate valid intervals of all fuzzy measures.
			\State Generate a random number $z \in [0,1]$.
			\If {$z < \lambda$} 
				\State Update one element in $\mathbf{\mathscr{W}}_p$. (small-scale) 
			\Else
				\State  Update all elements in $\mathbf{\mathscr{W}}_p$. (large-scale)
			\EndIf
		\EndFor
	\State  Compute $ J(\mathbf{\mathscr{W}})$ using \eqref{eq:objlnmultiresminmax}.
	\State  Select and keep the fuzzy measures with low objective function values.  
	\If {$ min(\mathbf{J}_\mathbf{\mathscr{W}}^{t}) < J^*$}
	\State $J^* = min(\mathbf{J}_\mathbf{\mathscr{W}}^t)$, $\mathbf{g}^* = \argminA_{\mathbf{\mathscr{W}}} J^* $
	\EndIf
	\If {$|J^* - min(\mathbf{J}_\mathbf{\mathscr{W}}^{t})| \leq F_T$}
	break;
	\EndIf
	\State $t \gets t+1$.
\EndWhile

\Return Optimal fuzzy measure $\mathbf{g}^* $
\Statex 
\Statex  \textbf{Fusion Stage}
\Require Testing Data, $\mathbf{g}^* $
\State $TestLabels \gets$ CI fusion output computed based on \eqref{eq:cg} using the learned optimal fuzzy measure $\mathbf{g}^*$ 

\Return $TestLabels$
\end{algorithmic}
\end{algorithm}

Once a new generation of measures with updated values were generated, the old and new  measure element values were pooled together (size $2P$) and  their fitness values were computed using \eqref{eq:objlnmultiresminmax}. Then, $P/2$ measures with the top 25\% fitness values are kept and carried over to the next iteration (elitism), and the remaining $P/2$ measures to be carried over were sampled according to a multinomial distribution based on their fitness values from the remaining 75\% of the parent and child population pool, following a similar approach in \citep{du2016multiple}. In the new measure population, the measure with the highest fitness value is kept as the current best measure $\mathbf{g}^*$. The process continues until a stopping criterion is reached, such as when the maximum number of iterations or the change in the objective function value from one iteration to the next is smaller than a fixed threshold, $F_T$. In our experiments, we stop if the difference between the current optimal measure and the previous optimal measure is smaller than $F_T=0.0001$. We used a high maximum number of iteration $I=5000$ and the algorithm will automatically stop if it reaches convergence before the maximum iteration. \textcolor{black}{The number of iterations it takes to converge depends on the random initialization. In our hyperspectral and LiDAR fusion experiments below, the algorithm converged in 165 iterations on average.}

At the end of the model learning process, the best measure $\mathbf{g}^*$ with the optimum fitness value was returned as the learned fuzzy measure. We then use this learned $\mathbf{g}^*$ to compute the pixel-level fusion results using the Choquet integral described in \eqref{eq:cg}. More details about the evolutionary algorithm can be seen in \citep{du2016multiple} and the pseudocode of our learning scheme is shown in Algorithm~\ref{alg:mici}.

Note that in our optimization and model learning process, we only use bag-level imprecise labels. This means, our method is not a standard supervised algorithm in which complete, pixel-wise  training labels are required. Instead, we learn from imprecisely labeled data and produce precise classification maps through the fusion of multi-modal, multi-resolution data. In the fuzzy measure learning (``training'') stage, the algorithm learns the optimal fuzzy measure $\mathbf{g^*}$ from imprecise bag-level labels. In the fusion stage, the algorithm produces complete, pixel-level fusion maps over the scene.  The estimated $\mathbf{g^*}$ represents the relationship among the fusion sources. Thus, the estimated $\mathbf{g^*}$ can be used to fuse any scene as long as the sensor sources in the test scene have similar relationships as in training. 



\section{Experiments}
\label{sec:experiments}

This section presents experimental results of the proposed MIMRF algorithm on two real remote sensing data sets. First, the proposed MIMRF algorithm is used for hyperspectral and LiDAR fusion for scene understanding in the MUUFL Gulfport dataset. Then, the MIMRF algorithm  is used for weed detection on a multi-resolution soybean and weed dataset. \textcolor{black}{At the end of this section, discussions are provided on the assumptions, limitations, and applications of the proposed MIMRF framework.}

\subsection{MUUFL Gulfport HSI and LiDAR Fusion Data Set}
\label{sec:experiments_muufl}
This section desribes the  MUUFL Gulfport hyperspectral and LiDAR fusion data set and presents experimental results. The proposed MIMRF was used to perform multi-resolution fusion on the hyperspectral imagery and raw LiDAR point cloud data. 

\subsubsection{Data Set Description}  
The MUUFL Gulfport hyperspectral imagery and LiDAR data set \cite{gader2013muufl,du2017technical, alina_zare_2018_1186326} was collected over the University of Southern Mississippi - Gulfpark campus in November 2011. The data set contains hyperspectral and LiDAR data from two flights. The RGB images from the HSI imagery of the two flights (named ``campus 1'' and ``campus 2'' data) are shown in Figure \ref{fig:muuflRGBcampus1_220} and Figure \ref{fig:muuflRGBcampus2_220}.  \textcolor{black}{Only the first 220 columns of the hyperspectral data were kept in order to exclude the bright beach sand regions in the lower right corner in the original images.} The first four and last four bands of the data were removed due to noise. The size of the HSI image is $325\times220\times64$ for both flights in this experiment.  

\begin{figure}[h]
\begin{subfigure}[t]{0.38\linewidth}
\centering
\includegraphics[width=\textwidth,trim={40mm 10mm 60mm 12mm},clip]{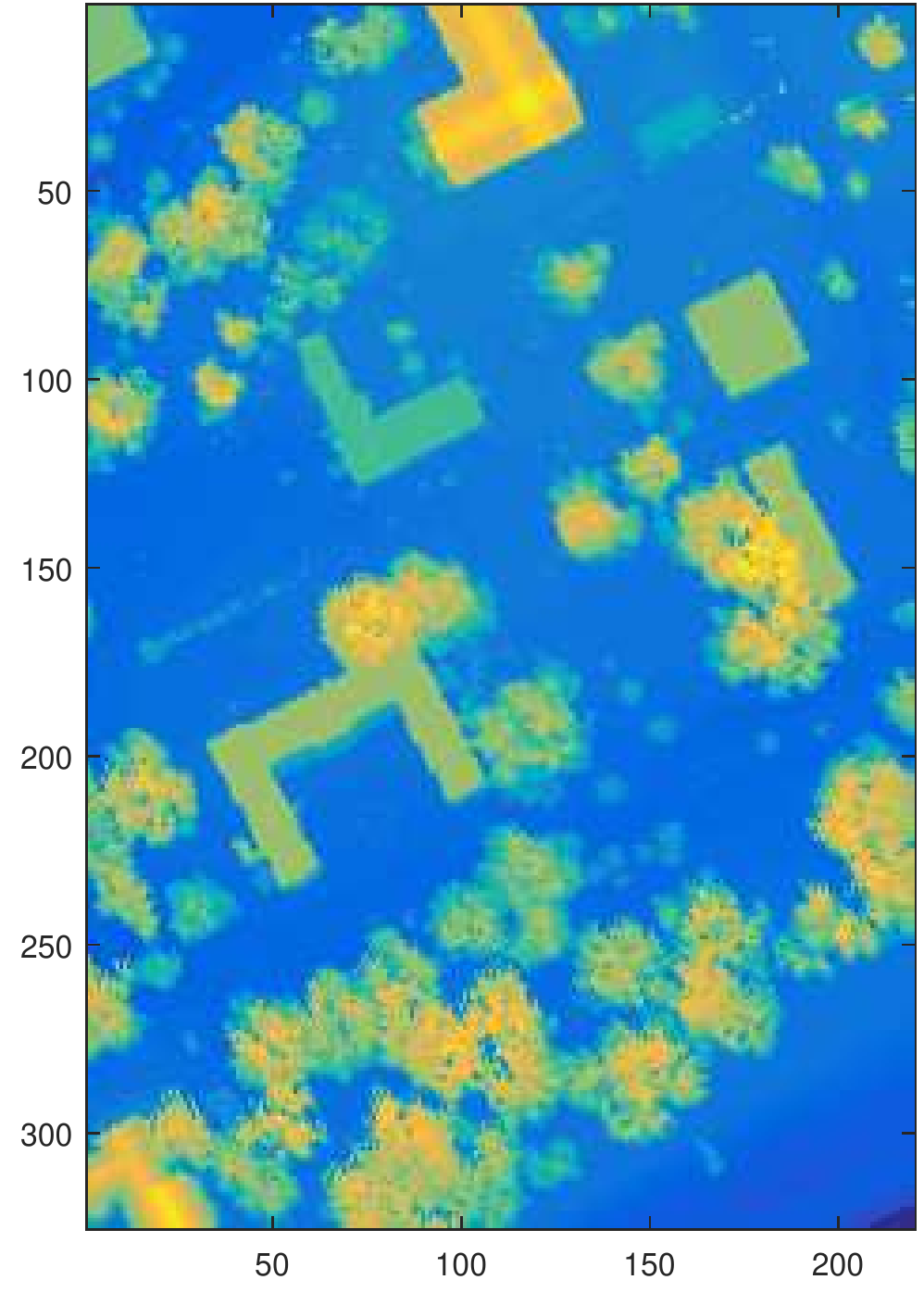}
\caption{}
\label{fig:muuflcampus1raster220}
\end{subfigure}
\begin{subfigure}[t]{0.38\linewidth}
\centering
\includegraphics[width=\textwidth,trim={40mm 10mm 60mm 12mm},clip]{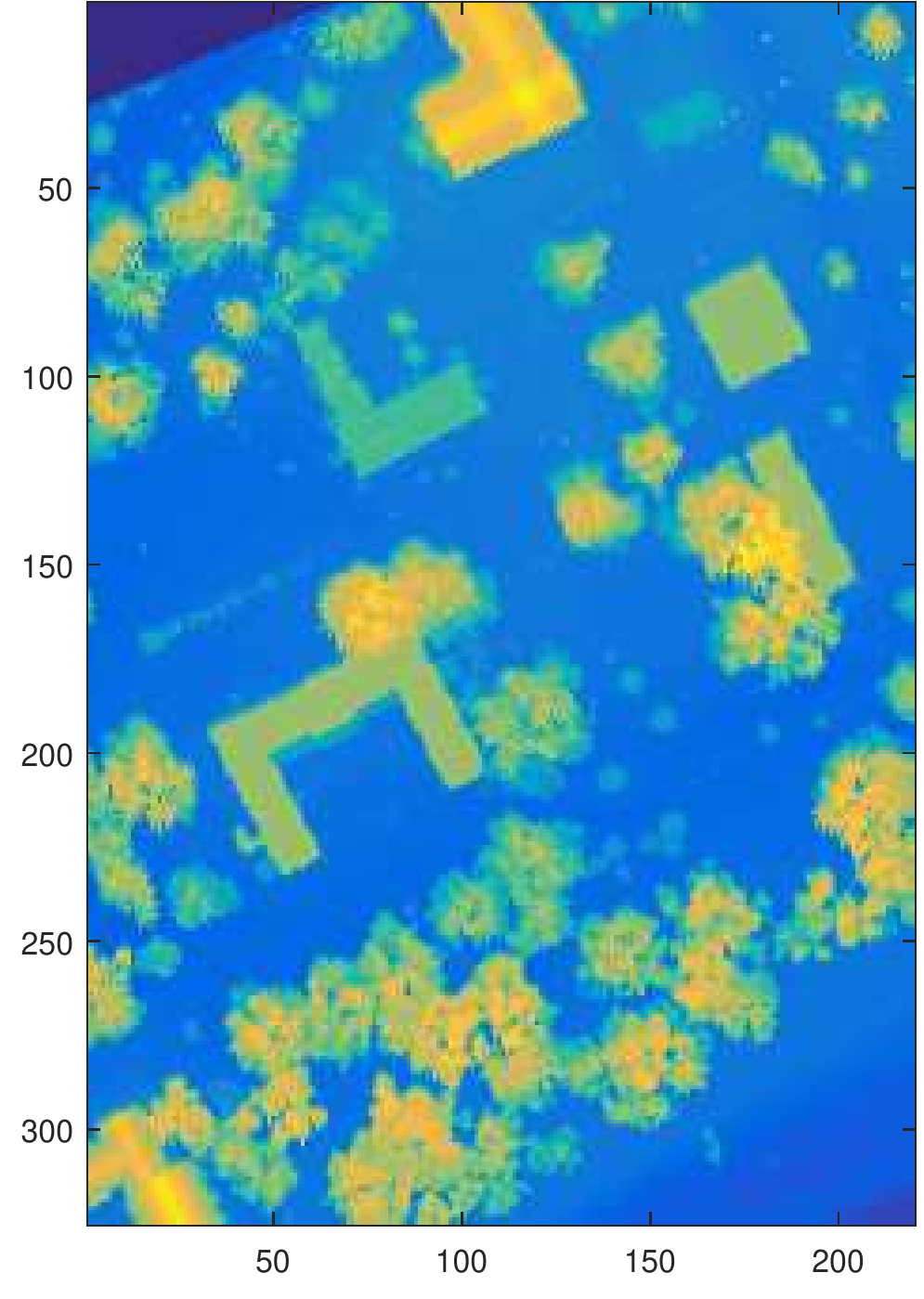}
\caption{}
\label{fig:muuflcampus2raster220}
\end{subfigure}
\begin{subfigure}[t]{0.2\linewidth}
\centering
\includegraphics[height=5cm,trim={50mm 0mm 1mm 1mm},clip]{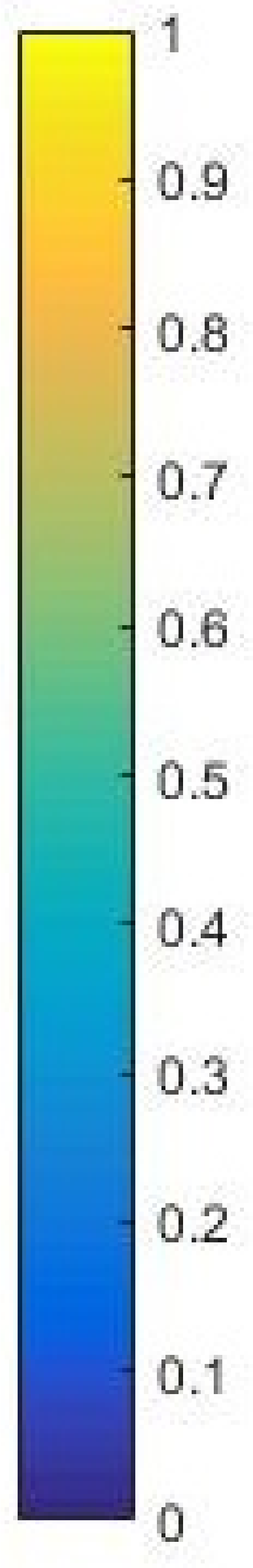}
\caption{}
\label{fig:colorbar}
\end{subfigure}
\caption[Raster image of the first return MUUFL Gulfport LiDAR data.]{Raster image of the first return MUUFL Gulfport LiDAR data.  The color represents the lidar height information (normalized). The rasterized image is in the same size as the hyperspectral imagery. (a) Campus 1 data; (b) Campus 2 data. (c) Color mapping. }
\label{fig:muuflRGBcampus_220raster}
\end{figure}

The LiDAR raw point cloud data was collected by the Gemini LiDAR sensor at 3500 ft over the scene \cite{gader2013muufl}. The scatter plot of the raw LiDAR point cloud data over the entire Gulfpark campus is shown in Figure \ref{fig:lidar_full1}. Rasterized LiDAR imagery (pre-processed by 3001 Inc. and Optech Inc. using nearest neighbor methods) was used to generate results for comparison methods.  The rasterized LiDAR imagery for campus 1 and campus 2 are shown in Figure \ref{fig:muuflcampus1raster220} and Figure \ref{fig:muuflcampus2raster220}.





\subsubsection{Automated Training Label Generation} 
\label{sec:autotrainlabelgen}
As discussed in the Introduction, hand-labeling is tedious, expensive, and prone to error, especially for large data sets.  For transition regions such as edges of a building, it is impossible for humans to visually identify and correctly label each pixel by hand due to mixing. In this section, an automated label generation method is proposed to automatically generate bag-level training labels for this data set for regions where crowd-sourced map data is available. 

\begin{figure}[h]
\begin{subfigure}[h]{0.4\linewidth}
\centering
\includegraphics[width=\textwidth,trim={46mm 15mm 36mm 11mm},clip]{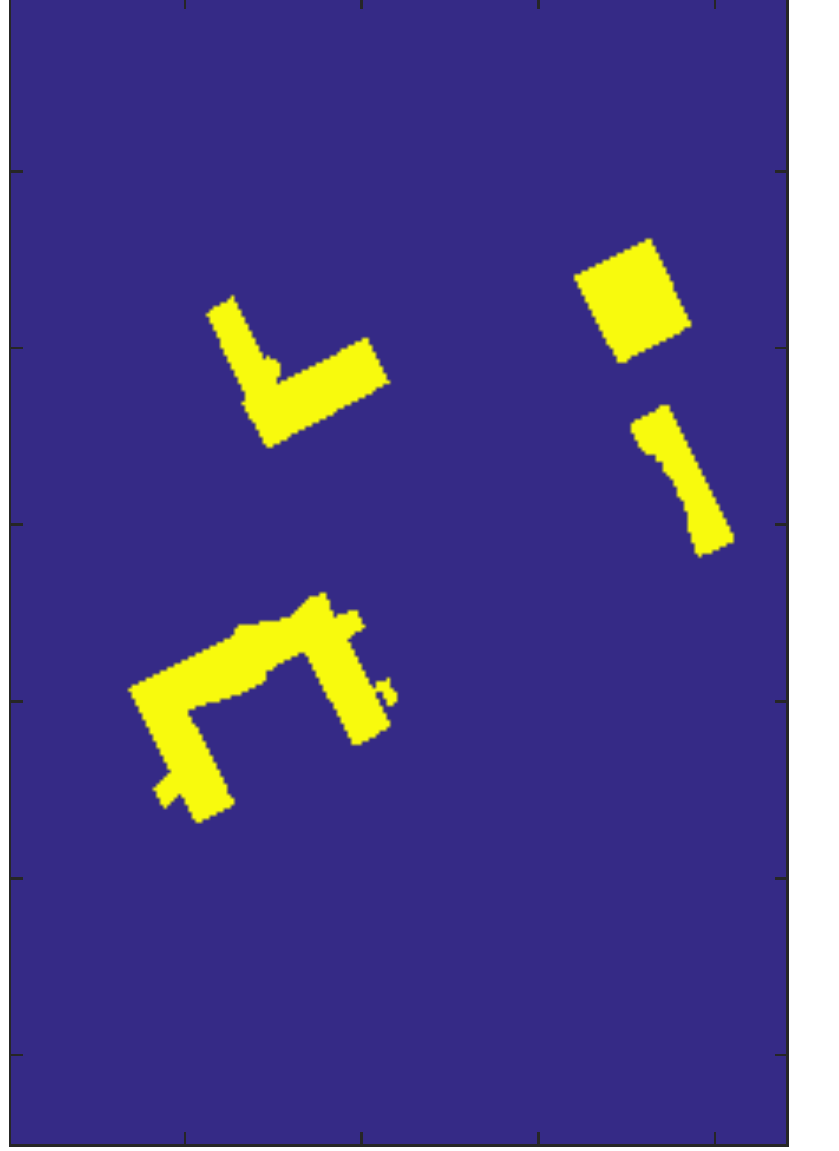}
\caption{}
\label{fig:muufl_building_GT}
\end{subfigure}
\hspace{5mm} 
\begin{subfigure}[h]{0.38\linewidth}
\centering
\includegraphics[width=\textwidth,trim={52mm 18mm 46mm 12mm},clip]{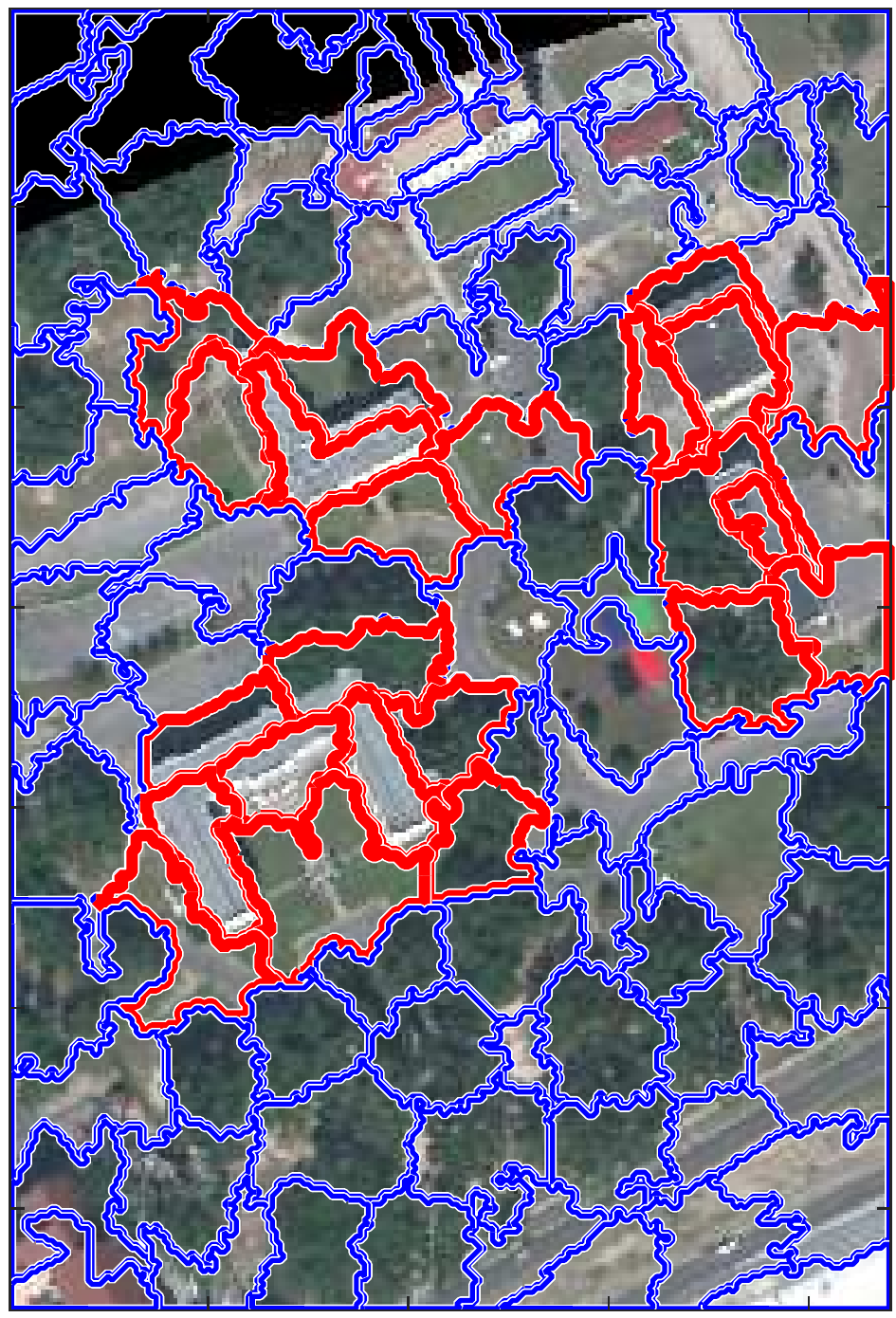}
\caption{}
\label{fig:muufl_slic_1}
\end{subfigure}
\caption[The Ground Truth map and the SLIC segmentation map of the MUUFL Gulfport HSI data for building detection.]{The Ground Truth map and the SLIC segmentation map of the MUUFL Gulfport HSI data for building detection. (a) The Ground Truth map of the buildings in MUUFL Gulfport HSI data. The yellow highlights the ground truth building locations \cite{du2017technical}. (b) The SLIC segmentation result on MUUFL Gulfport HSI data. Red marks positive training bags and blue marks negative bags for building detection experiment.}
\label{fig:muuflGTslic}
\end{figure}

The ``bags'' for this data set were constructed using the simple linear iterative clustering (SLIC) algorithm \cite{achanta2010slictechreport, achanta2012slic}. The SLIC algorithm is a widely used, unsupervised, open-source superpixel segmentation algorithm that can produce spatially coherent regions \cite{achanta2012slic}. \textcolor{black}{The number of superpixels was set to 100 and the compactness factor was set to 20. We chose this parameter setting so that the resulting superpixels are large enough to be spatially coherent and computationally inexpensive, yet small enough to be able to shape the edges along the buildings and also produce entirely non-building (non-target) superpixels. Our parameter choice balances well between computation time and learning performance empirically.} Figure~\ref{fig:muufl_slic_1} shows the SLIC segmentation result on the MUUFL Gulfport hyperspectral campus 1 data. The lines mark the boundaries for each superpixel. Each superpixel is treated as a ``bag'' and the pixels in each superpixel are the instances in the bag.

\begin{figure}[h!]
\centering
\includegraphics[width=0.49\textwidth,trim={0mm 0mm 0mm 0mm},clip]{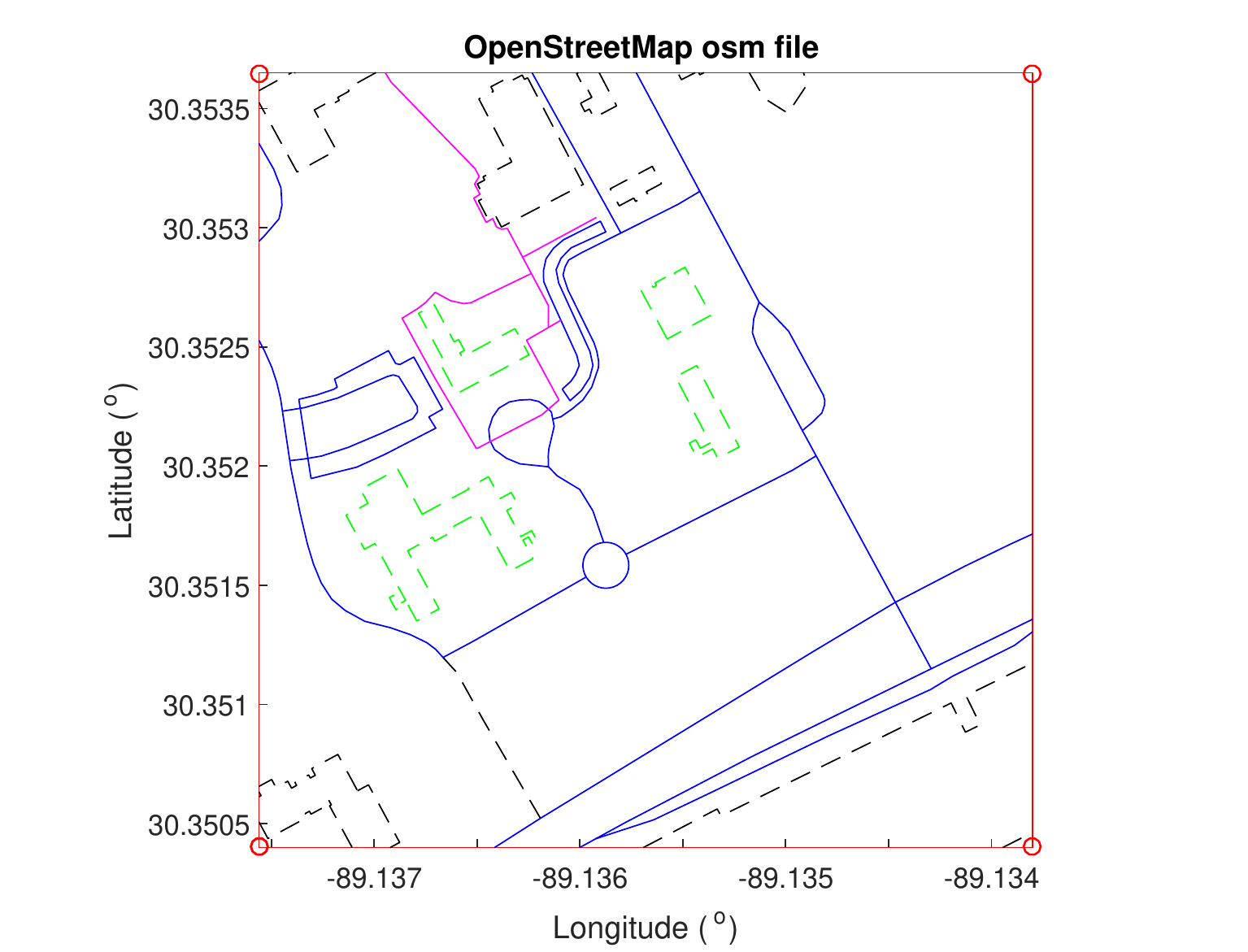}
\caption[Open Street Map imagery over MUUFL Gulfport campus 1.]{Open Street Map imagery over MUUFL Gulfport campus 1. The blue lines correspond to road and highway.  The magenta lines correspond to sidewalk/footway. The green lines mark buildings. Here, the ``building'' tag is specific to the buildings with a grey asphalt roof. The black lines correspond to ``other'' tags. }
\label{fig:muuflosmcampus1}
\end{figure}

Training labels for each bag were automatically generated based on publicly available, crowdsourced map data from the Open Street Map (OSM) \cite{osm1}. Figure~\ref{fig:muuflosmcampus1} shows the map extracted from OSM over the campus area and the available tags, such as ``highway'', ``footway'', ``building'', ``parking'', etc. The blue lines correspond to asphalt, which includes road, highway and parking lot.  The magenta lines correspond to sidewalk/footway. The green lines mark buildings. The black lines correspond to ``other'' tags (regarded as background in our experiments). Keypoints, such as corners of the buildings, were selected manually from both OSM map and HSI RGB imagery and an affine transformation \cite{schalkoff1989digital} was used to map between the OSM Lat/Lon coordinates and the HSI pixel coordinates.  Then, all the superpixels that contain at least one building pixel are labeled positive and all the superpixels that do not contain building pixels are labeled negative. Figure~\ref{fig:muufl_building_GT} shows the ground truth map for buildings with a grey roof and Figure~\ref{fig:muufl_slic_1} shows the bag-level labels for all superpixels. 
 

Buildings are easy to identify in a scene due to their spatial coherence. However, producing an accurate pixel-level classification map for buildings is a challenging task due to many transition areas along the edges of the buildings. The transition areas can be only a few pixels or less from the top of a building to the neighboring ground surface. The edge areas are also very difficult or impossible to label manually. \textcolor{black}{Similar observations can be made in a tree canopy, where pixels along the edge of tree leaves are also highly mixed and can be easily mis-identified as false positives.} In the following, we first present the fusion source generation process in Section~\ref{sec:iva3} and the overall building detection results in Section~\ref{sec:iva4}. Then, we further investigate the performance of the proposed algorithm specifically on those difficult transitional regions, such as building edges, in Section~\ref{sec:iva5}.


\subsubsection{Generation of Fusion Sources for Building Detection} 
\label{sec:iva3}
In this experiment, three sources were generated for fusion  ($m=3$), one from HSI and two from LiDAR. First, building points were manually extracted from the hyperspectral imagery and the mean spectral signature of these point were computed. The adaptive coherence estimator (ACE) detector \cite{scharf1996adaptive,kraut2005adaptive,pulsone2000computationally} was applied to the HSI imagery using spectral signature of the building points. The ACE detection map for buildings is shown in Figure~\ref{fig:TestConfMap_Building_train1test2_ACE}, where the ACE confidence map correctly highlights most buildings, but has some false positives on asphalt roads, which have similar spectral signature as the roof materials. The ACE detector also has low confidence on the top right building, possibly due to the darkness of the roof.

\begin{figure}[h]
\begin{subfigure}[h]{0.49\linewidth}
\centering
\includegraphics[width=\textwidth,trim={10mm 0mm 10mm 0mm},clip]{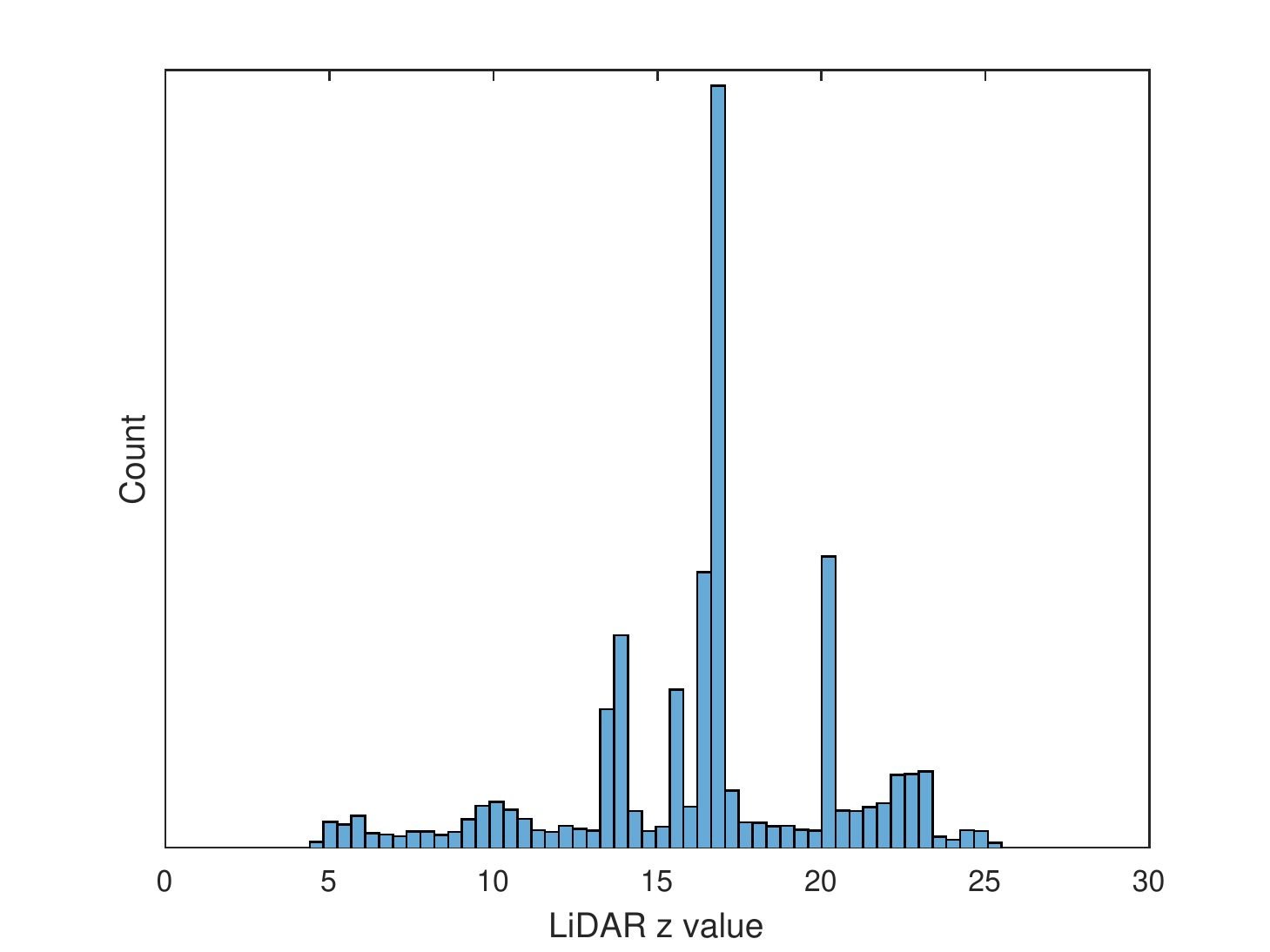}
\caption{}
\label{fig:muufl_lidar_hist_binall_noytick}
\end{subfigure}
\begin{subfigure}[h]{0.49\linewidth}
\centering
\includegraphics[width=\textwidth,trim={10mm 0mm 10mm 0mm},clip]{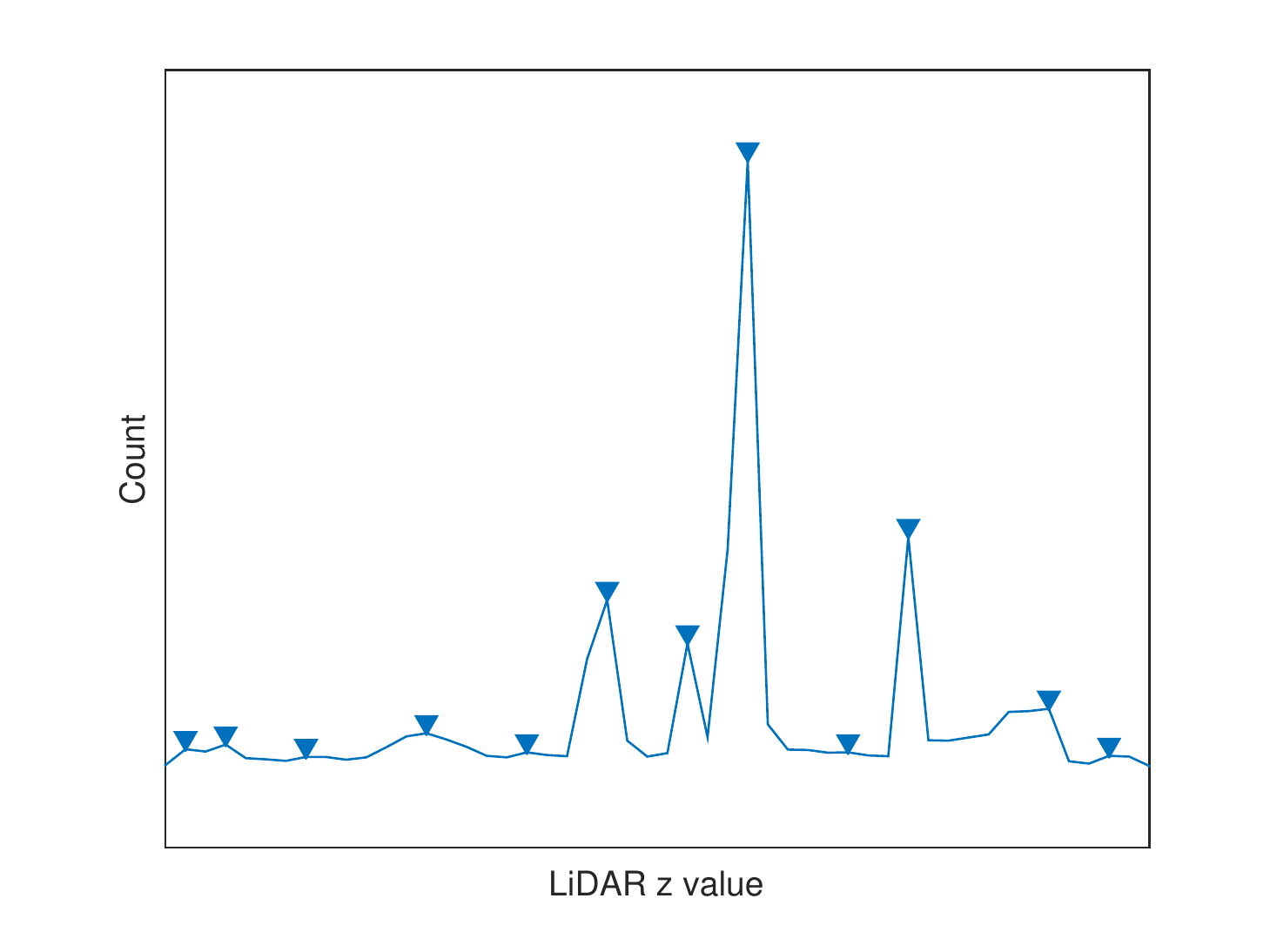}
\caption{}
\label{fig:muufl_lidar_hist_findpeaks_noytick}
\end{subfigure}
\caption{The histogram and peaks of the LiDAR values of building points. (a) The histogram of the LiDAR values of building points. (b) The peaks found based on the histogram in (a). }
\label{fig:muufl_lidar_hist_findpeaks}
\end{figure}

\begin{figure}[h]
\begin{subfigure}[h]{0.32\linewidth}
\centering
\includegraphics[width=\textwidth,trim={46mm 10mm 36mm 7mm},clip]{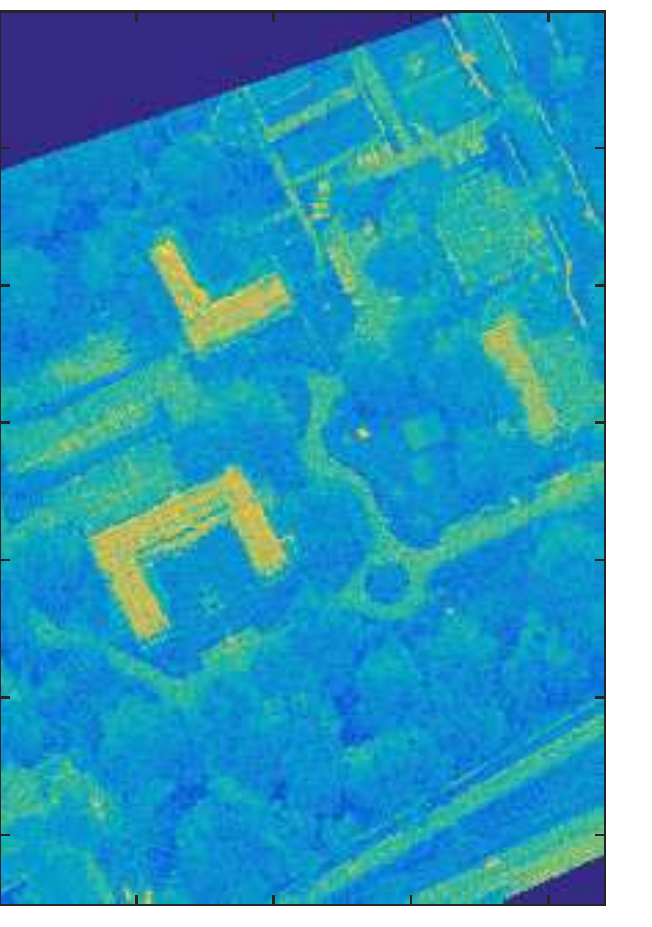}
\caption{}
\label{fig:TestConfMap_Building_train1test2_ACE}
\end{subfigure}
\begin{subfigure}[h]{0.32\linewidth}
\centering
\includegraphics[width=\textwidth,trim={46mm 10mm 36mm 7mm},clip]{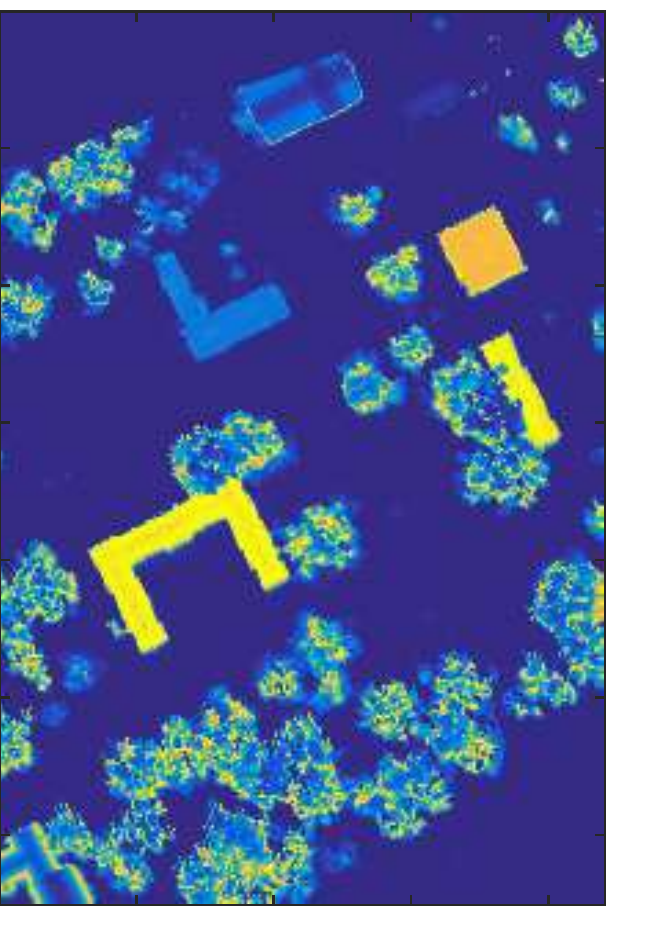}
\caption{}
\label{fig:TestConfMap_Building_train1test2_lidar1}
\end{subfigure}
\begin{subfigure}[h]{0.32\linewidth}
\centering
\includegraphics[width=\textwidth,trim={46mm 10mm 36mm 7mm},clip]{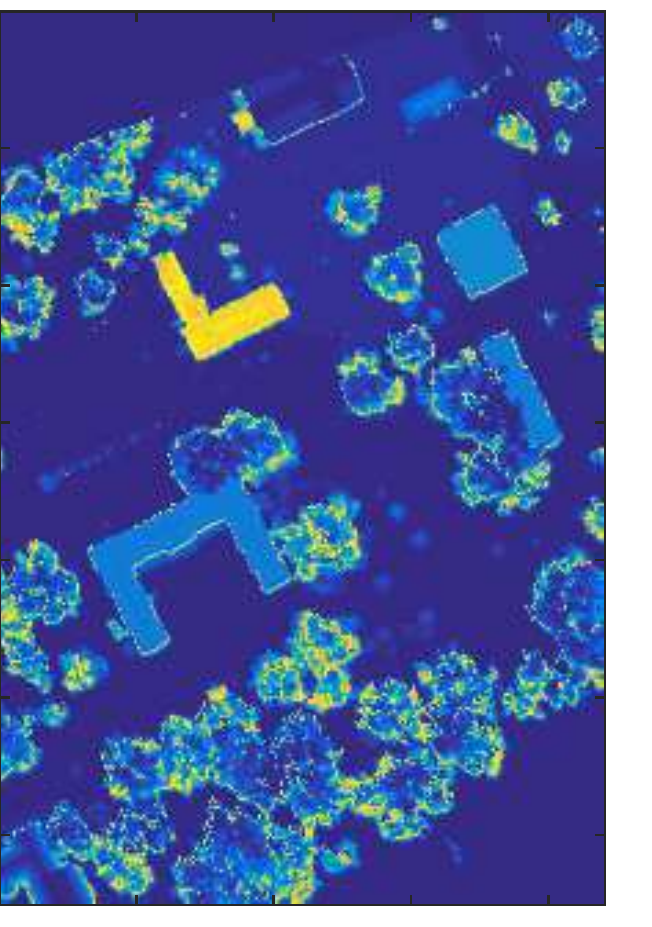}
\caption{}
\label{fig:TestConfMap_Building_train1test2_lidar2}
\end{subfigure}
\caption{The confidence maps from HSI and LiDAR data for building detection in the MUUFL Gulfport. These three sources are used for fusion. (a) ACE detection map from HSI data. (b)(c) The height-based LiDAR confidence maps for building detection in the MUUFL Gulfport data. (b)(c) were distance maps computed against the two peaks found in
Fig.~\ref{fig:muufl_lidar_hist_findpeaks}. The colorbar is the same as in Figure~\ref{fig:colorbar}.}
\label{fig:TestConfMap_Building_train1test2_lidar12}
\end{figure}



\textcolor{black}{Then, two LiDAR confidence maps were generated to be fused with the ACE detection map.} The LiDAR height information of extracted training building points were plotted in a histogram, as shown in Figure~\ref{fig:muufl_lidar_hist_binall_noytick}. The peaks of the histogram were found by using the MATLAB \textit{findpeaks()} function, as shown in Figure~\ref{fig:muufl_lidar_hist_findpeaks_noytick}. \textcolor{black}{The Euclidean distances of all LiDAR points in the scene were computed against two peak height values of the building points, $16.8m$ and $13.9m$, to produce two confidence maps, which matches the two distinct heights of grey-roof buildings in the scene.} Then, we designed an exponential function to compute confidence values based on the distances, written as
\begin{equation}
\textcolor{black}{\mathbf{c}_{n} =e^ {-\mathbf{d}_{n}/2}},
\label{eq:cl}
\end{equation}
\textcolor{black}{where $\mathbf{c}_{n}$ is the confidence value of the $n^{th}$ LiDAR point in the scene and $\mathbf{d}_{n}$ is the Euclidean distance between the $n^{th}$ LiDAR point and the peak height value.  } In this way, the LiDAR maps have high confidence on points with similar height to training building points and have low confidence on points with distinctly different heights than buildings.  \textcolor{black}{Figure~\ref{fig:TestConfMap_Building_train1test2_lidar1} and Figure~\ref{fig:TestConfMap_Building_train1test2_lidar2} show two confidence maps computed from a rasterized LiDAR map against the two peak height values, where Figure~\ref{fig:TestConfMap_Building_train1test2_lidar1} highlights three buildings with height value around $16.8m$ and Figure~\ref{fig:TestConfMap_Building_train1test2_lidar2} highlights the L-shaped building with height value around $13.9m$. Both LiDAR confidence maps also highlight some trees of similar height. On the other hand, the ACE confidence map (Figure~\ref{fig:TestConfMap_Building_train1test2_ACE}) shows high values on some buildings and roads, and low values on trees.  Thus, the purpose of our proposed MIMRF is to fuse all three confidence maps as shown in Figure~\ref{fig:TestConfMap_Building_train1test2_lidar12} to the effect that it highlights all buildings only and suppress other noise objects (such as trees and roads). Note that the purpose of this experiment is not to persuade that this Euclidean-distance-based method is the best or the most general method to produce confidence maps, but rather given such ACE and LiDAR confidence maps that highlight different objects in the scene, our proposed MIMRF algorithm is able to perform fusion and produce a fused map that assigns high confidence on all the target buildings and low confidence on other non-target objects, which is the goal of an effective fusion method. Other feature extraction methods can be explored in future work  for LiDAR data, such as texture or shape-based features. }For the proposed MIMRF algorithm, the raw LiDAR points were  directly used as input instead of the pre-rasterized LiDAR imagery. The rasterized LiDAR confidence maps as shown in Figure~\ref{fig:TestConfMap_Building_train1test2_lidar1} and Figure~\ref{fig:TestConfMap_Building_train1test2_lidar2} were used for non-multi-resolution comparison methods.

\subsubsection{MIMRF Fusion Results on Building Detection} 
\label{sec:iva4}

The proposed MIMRF algorithm was compared with a variety of methods to demonstrate its effectiveness in multi-resolution, multi-modal fusion with label uncertainty. First, the MIMRF was compared with results of individual sources, before fusion (i.e., the ACE confidence map and two LiDAR confidence maps). Then, the MIMRF was compared with widely adopted fusion and decision-making methods, including the Support Vector Machine (SVM) and the min/max/mean aggregation operators. Both SVM and the aggregation operators are applicable to images with pixel-to-pixel correspondence only and cannot handle multi-resolution fusion. The CI-QP \cite{grabisch1996application} approach, as mentioned in Section~\ref{sec:relatedworkci}, was also used as a comparison method. The CI-QP approach uses the Choquet integral to perform fusion, but CI-QP does not support MIL-type learning with bag-level labels. The mi-SVM \cite{andrews2002support} method, also mentioned in Section~\ref{sec:relatedworkmiclass}, was used as another comparison method. The mi-SVM algorithm works under the MIL framework, but does not support multi-resolution fusion. Additionly, the previously proposed MICI classifier fusion algorithm \cite{du2016multiple} was used for comparison. The MICI method is an MIL extension on Choquet integral and can handle label uncertainty, but MICI is only applicable to data with pixel-to-pixel correspondence and does not support multi-resolution fusion either. The CI-QP, mi-SVM, and MICI methods only work with rasterized LiDAR imagery, while our proposed MIMRF can directly handle raw LiDAR point cloud data. Table~\ref{table:featurecompare} shows a comprehensive list of comparison methods and their characteristics.

\begin{table}[t]
  \centering
\caption{The characteristics of comparison methods. ``Fusion'' means fusion method. ``CI'' means the method uses the Choquet integral as a fusion tool. ``MIL'' means the method supports the MIL framework and can handle label uncertainty. ``MR'' means the method supports multi-resolution and multi-modal fusion. The table cell is marked ``\checkmark'' if the method supports the heading conditions, and left blank if it does not work with the heading conditions. The top three rows indicate individual sources before fusion.}
\resizebox{0.45\textwidth}{!}{
\begin{tabular}{|c|c|c|c|c|}
\hline
 \textbf{Comparison Methods} & \textbf{Fusion} & \textbf{CI} & \textbf{MIL} & \textbf{MR}  \\
  \hline \hline
  ACE   &  & & & \\ 
  LiDAR1 &  & & &\\ 
  LiDAR2  &  & & &\\ \hdashline
  SVM/min/max/mean  &  \checkmark & & &\\
  mi-SVM & \checkmark & & \checkmark & \\ 
  CI-QP   &  \checkmark & \checkmark & &\\ 
  MICI &  \checkmark & \checkmark & \checkmark &\\ 
  MIMRF (proposed) & \checkmark  & \checkmark & \checkmark & \checkmark\\ 
     \hline
\end{tabular}}
  \label{table:featurecompare}
\end{table}

\begin{figure}[h]
\begin{subfigure}[t]{0.24\linewidth}
\centering
\includegraphics[width=\textwidth,trim={46mm 15mm 36mm 5mm},clip]{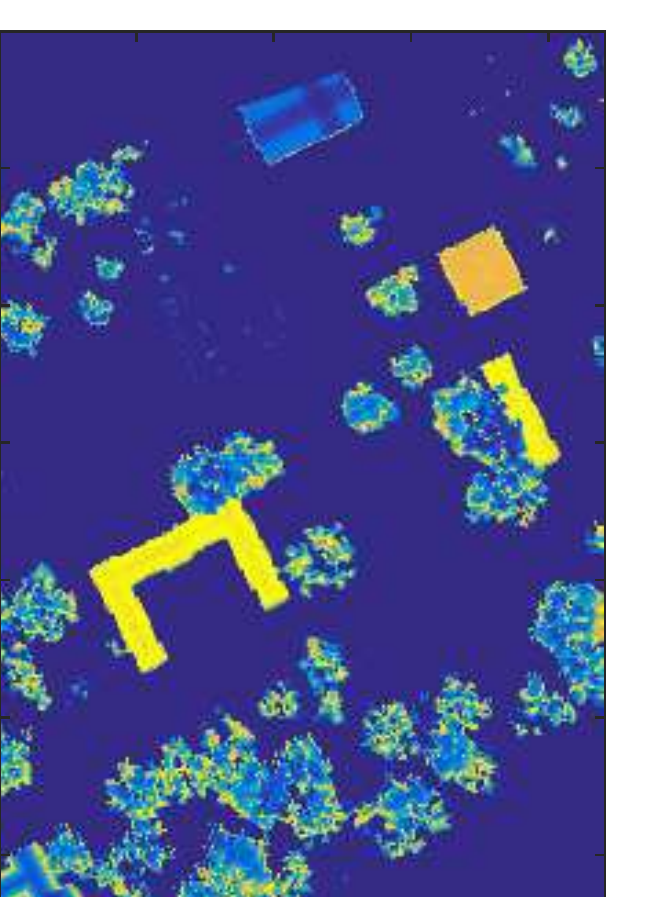}
\caption{}
\label{fig:TestConfMap_Building_train1test2_svm}
\end{subfigure}
\begin{subfigure}[t]{0.24\linewidth}
\centering
\includegraphics[width=\textwidth,trim={46mm 15mm 36mm 5mm},clip]{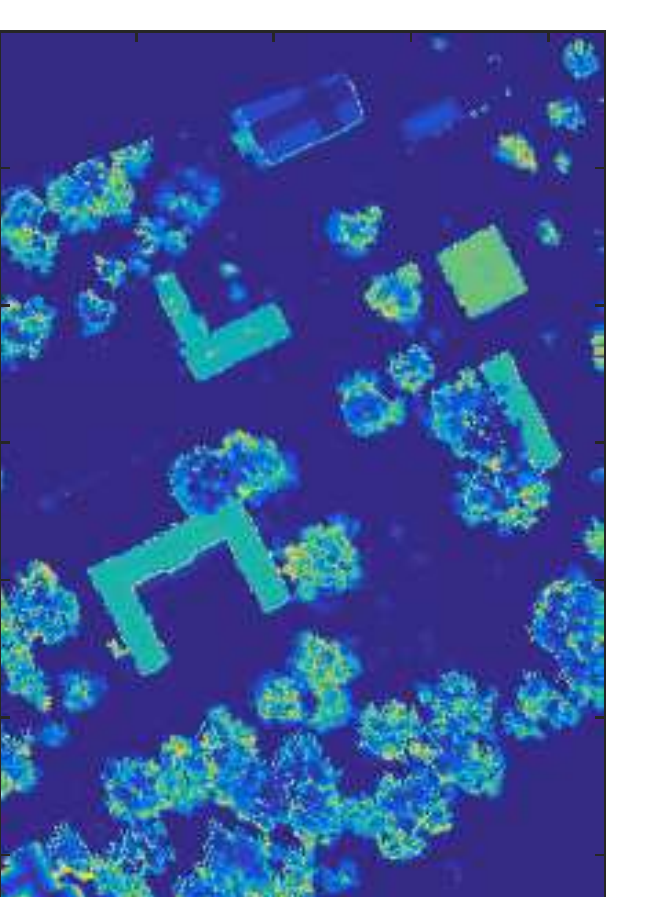}
\caption{}
\label{fig:TestConfMap_Building_train1test2_min}
\end{subfigure}
\begin{subfigure}[t]{0.24\linewidth}
\centering
\includegraphics[width=\textwidth,trim={46mm 15mm 36mm 5mm},clip]{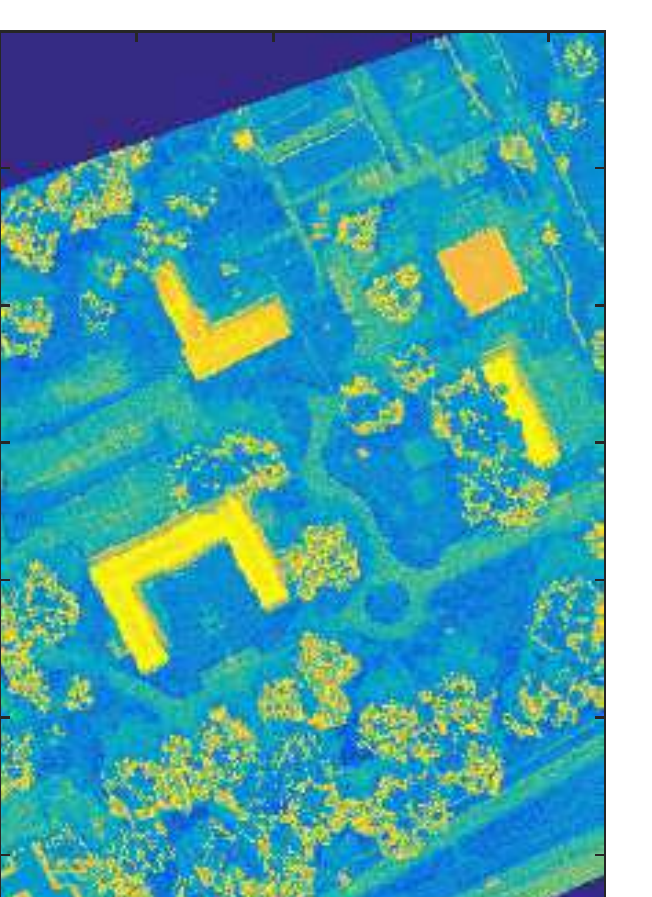}
\caption{}
\label{fig:TestConfMap_Building_train1test2_max}
\end{subfigure}
\begin{subfigure}[t]{0.24\linewidth}
\centering
\includegraphics[width=\textwidth,trim={46mm 15mm 36mm 5mm},clip]{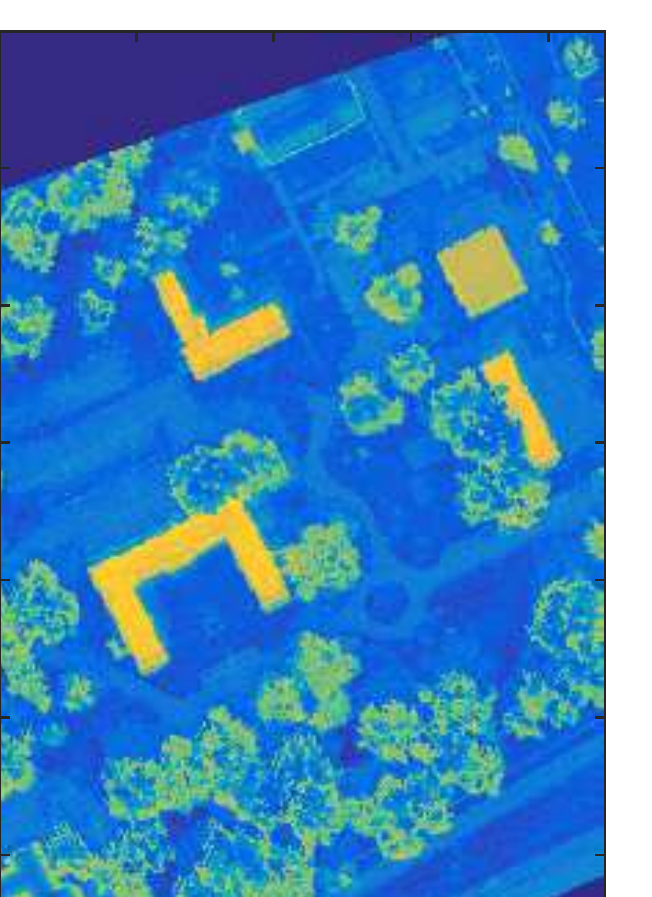}
\caption{}
\label{fig:TestConfMap_Building_train1test2_mean}
\end{subfigure}

\begin{subfigure}[t]{0.24\linewidth}
\centering
\includegraphics[width=\textwidth,trim={46mm 15mm 36mm 5mm},clip]{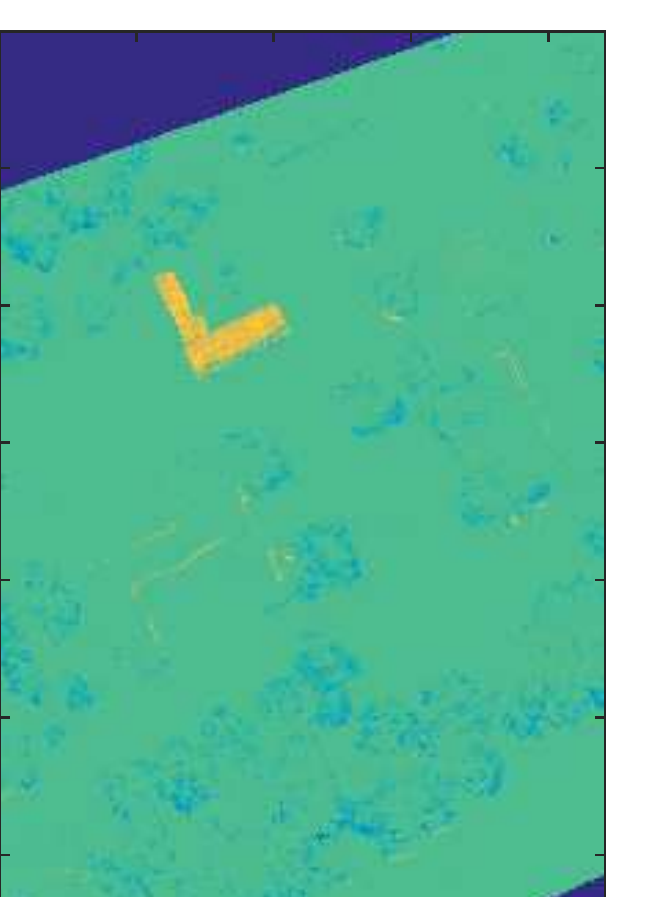}
\caption{}
\label{fig:TestConfMap_Building_train1test2_MISVM}
\end{subfigure}
\begin{subfigure}[t]{0.24\linewidth}
\centering
\includegraphics[width=\textwidth,trim={46mm 15mm 36mm 5mm},clip]{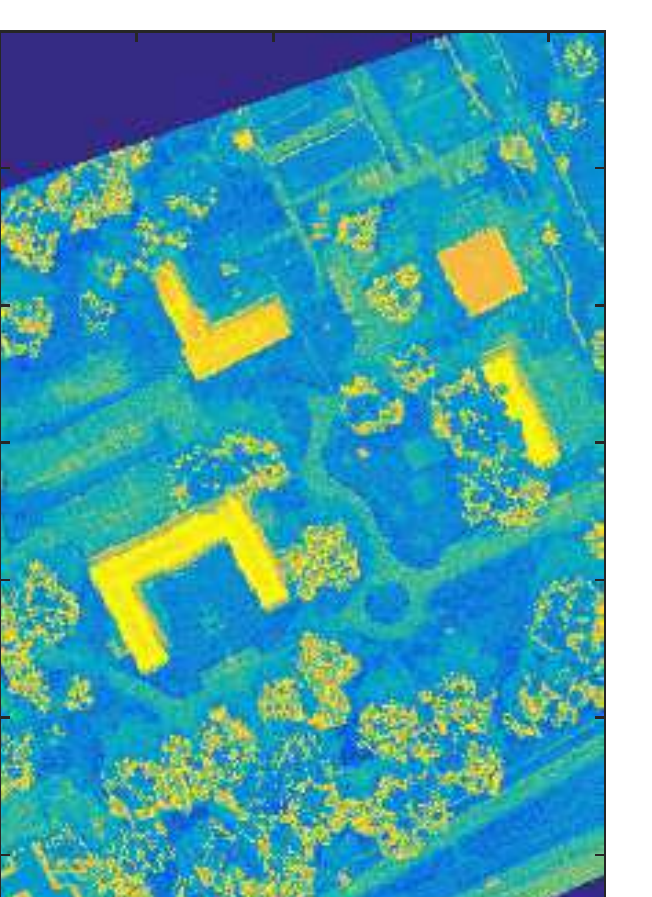}
\caption{}
\label{fig:TestConfMap_Building_train1test2_CIQP}
\end{subfigure}
\begin{subfigure}[t]{0.24\linewidth}
\centering
\includegraphics[width=\textwidth,trim={46mm 15mm 36mm 5mm},clip]{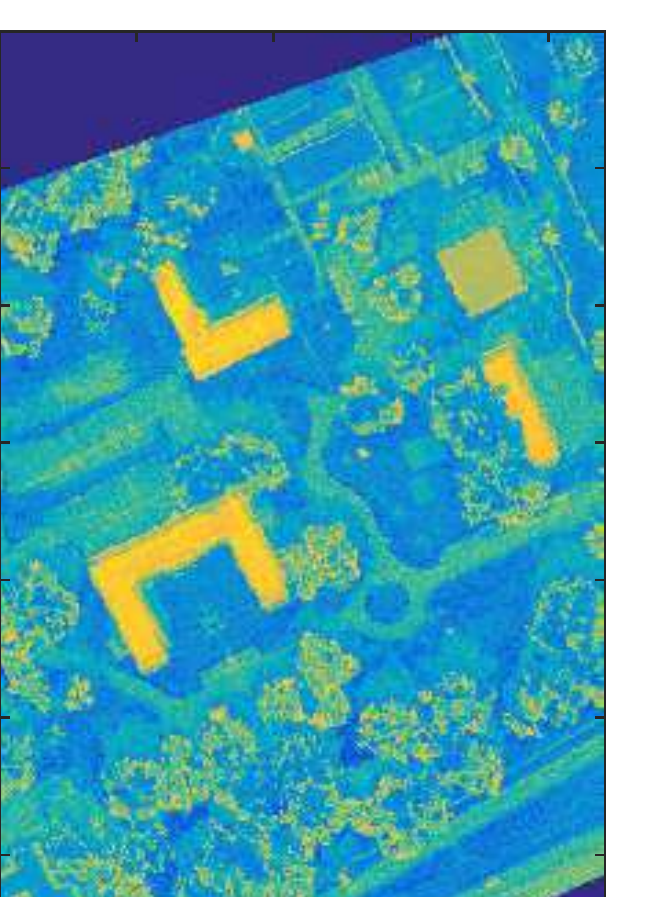}
\caption{}
\label{fig:TestConfMap_Building_train1test2_MICI}
\end{subfigure}
\begin{subfigure}[t]{0.24\linewidth}
\centering
\includegraphics[width=\textwidth,trim={46mm 15mm 36mm 5mm},clip]{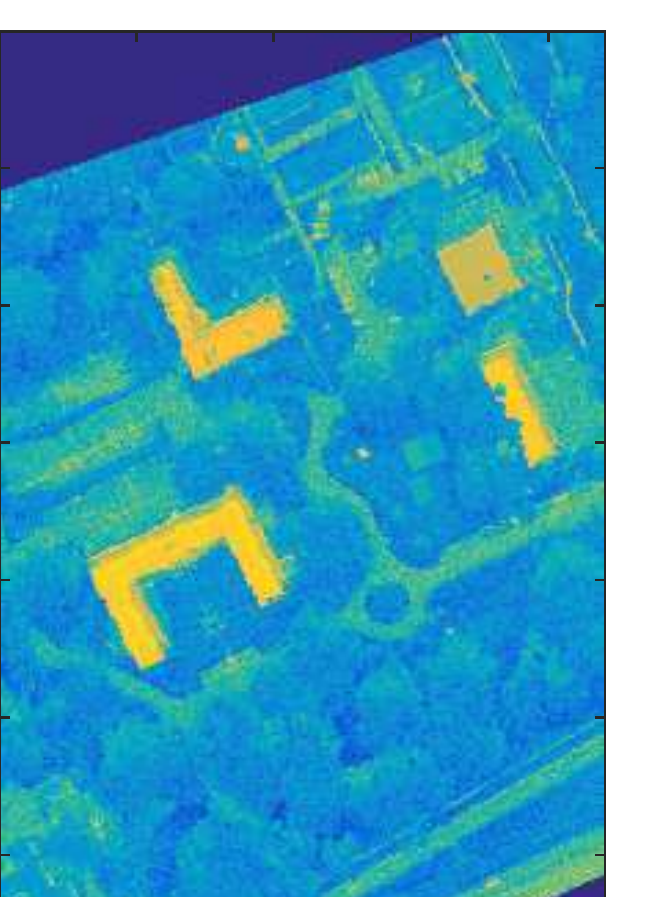}
\caption{}
\label{fig:TestConfMap_Building_train1test2_MRMICI}
\end{subfigure}
\caption{The fusion results for building detection in the MUUFL Gulfport data set. Train on campus 1 and test on campus 2. Fusion results by (a) SVM; (b) Min operator; (c) Max opeator; (d) Mean operator; (e) mi-SVM; (f) CI-QP; (g) MICI; (h) The proposed MIMRF algorithm.}
\label{fig:TestConfMap_Building_train1test2_all}
\end{figure}

Two-fold cross validation was performed on this data set, i.e., training on one flight and testing on another flight. The fusion results when training on campus 1 and testing on campus 2 across all methods are shown in Figure~\ref{fig:TestConfMap_Building_train1test2_all}. As shown, the MIMRF fusion results in Figure~\ref{fig:TestConfMap_Building_train1test2_MRMICI} accurately detects all four grey-roof buildings while having significantly less false positives on other non-building objects, such as trees, in the scene than comparison methods. 

\begin{figure}[h]
\begin{subfigure}[t]{\linewidth}
\centering
\includegraphics[width=0.9\textwidth,trim={0mm 0mm 0mm 0mm},clip]{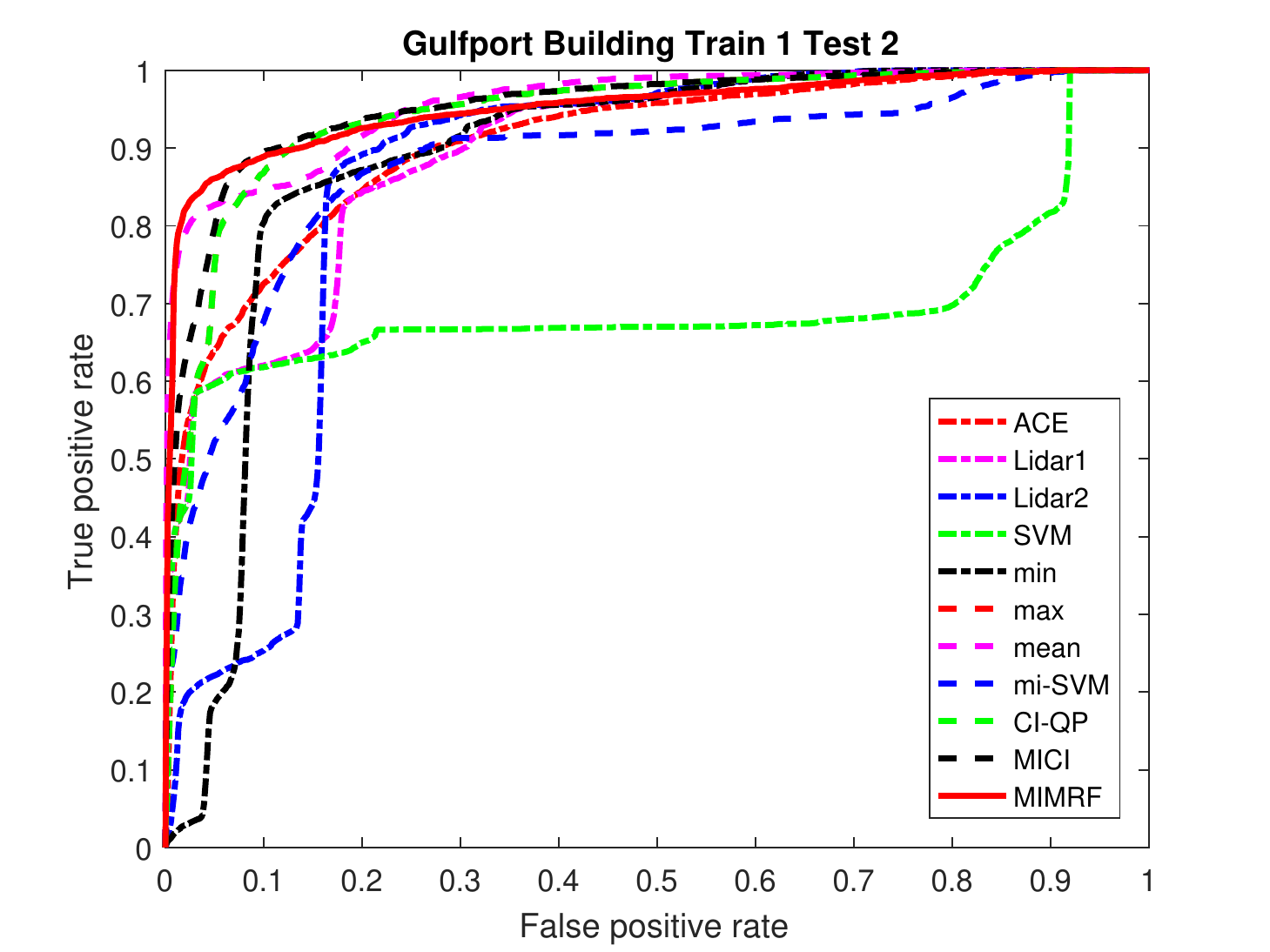}
\caption{}
\label{fig:roc_muufl_building_train1test2_all}
\end{subfigure}
\begin{subfigure}[t]{\linewidth}
\centering
\includegraphics[width=0.9\textwidth,trim={0mm 0mm 0mm 0mm},clip]{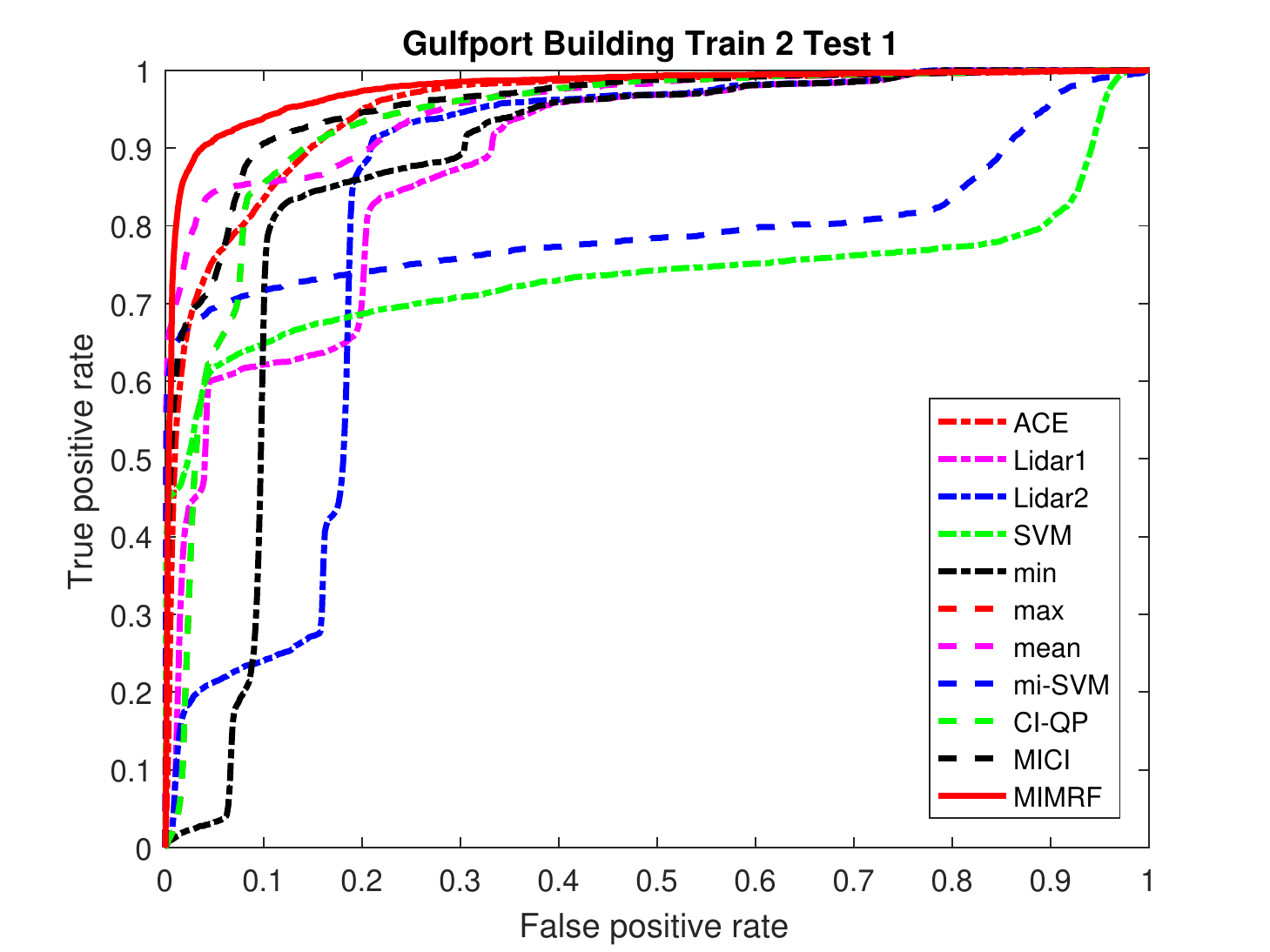}
\caption{}
\label{fig:roc_muufl_building_train2test1_all}
\end{subfigure}
\caption[The overall ROC curve for building detection for MUUFL Gulfport data.]{The overall ROC curve for building detection for MUUFL Gulfport data. (a) Train on campus 1, test on campus 2; (b) Train on campus 2, test on campus 1.}
\label{fig:roc_muufl_building_all}
\end{figure}

Figures~\ref{fig:roc_muufl_building_train1test2_all} and Figures~\ref{fig:roc_muufl_building_train2test1_all} show the overall ROC curve on building detection with cross validation. In addition, the Area Under Curve (AUC) results  from the ROC curves were computed to provide a quantitative comparison. The first two columns of Table~\ref{table:muuflbuildingAUC} shows the AUC results for building detection. The ACE, LiDAR1, and LiDAR2 rows are results from the three individual sources before fusion, and the methods below the dotted line are fusion results from all comparison methods.  As shown, the proposed MIMRF algorithm produces the best or second best ROC curve and AUC results.

Another evaluation metric, the Root Mean Square Error (RMSE), is used for comparison as well. The AUC evaluates how well the method detects the buildings (the higher AUC the better) and the RMSE shows how detection results on both the building and non-building points differ from the ground truth (the lower the RMSE the better). Table~\ref{table:muuflbuildingrmse}  shows the RMSE comparison results between MICI and MIMRF methods (the top methods with high AUC results). The proposed MIMRF has higher AUC performance and lower RMSE than comparison methods.

\subsubsection{MIMRF Results on Edge Areas}
\label{sec:iva5}
As discussed in Section~\ref{sec:autotrainlabelgen}, building detection is a challenging task along the edge areas due to spatial resolution.  The edge areas refer to transition areas and boundaries regions where there is a sudden change in altitude, including but not limited to building edges and edges in a tree canopy. These edge areas are difficult to label and are prone to noise and misalignment in image registration and rasterization. One of the biggest difference between the proposed MIMRF algorithm and previous methods is that the proposed MIMRF can directly use raw LiDAR point cloud data while previous image fusion methods require rasterization. We conducted additional experiments as follows to investigate the performance of all fusion methods on such edge areas, where the rasterization may be noisy, inaccurate, or misaligned. 

\begin{figure}[h]
\begin{subfigure}[t]{0.35\linewidth}
\centering
\includegraphics[width=\textwidth,trim={36mm 12mm 50mm 5mm},clip]{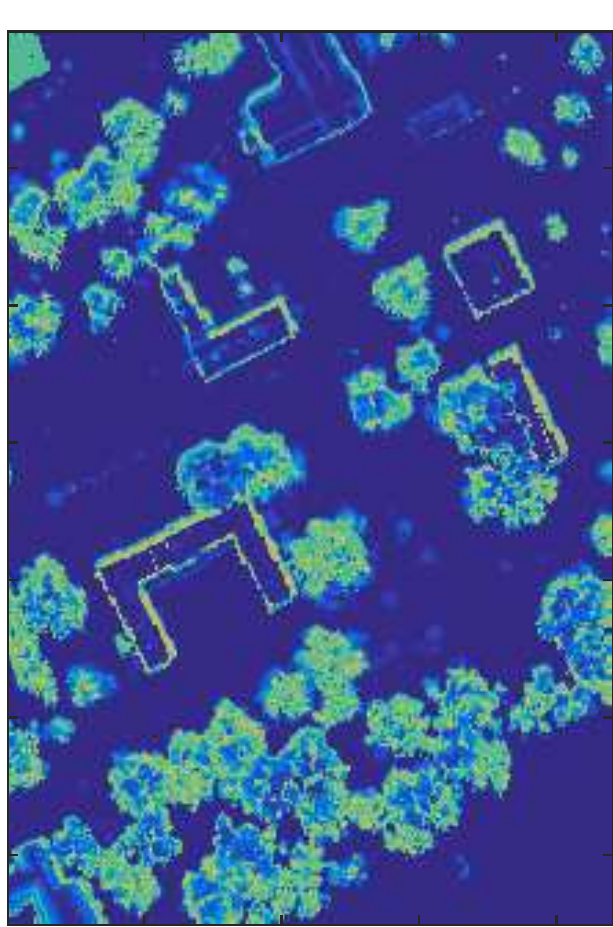}
\caption{}
\label{fig:diffmap_muufl_building_train2test1_lidar}
\end{subfigure}
\begin{subfigure}[t]{0.64\linewidth}
\centering
\includegraphics[width=\textwidth,trim={0mm 0mm 0mm 0mm},clip]{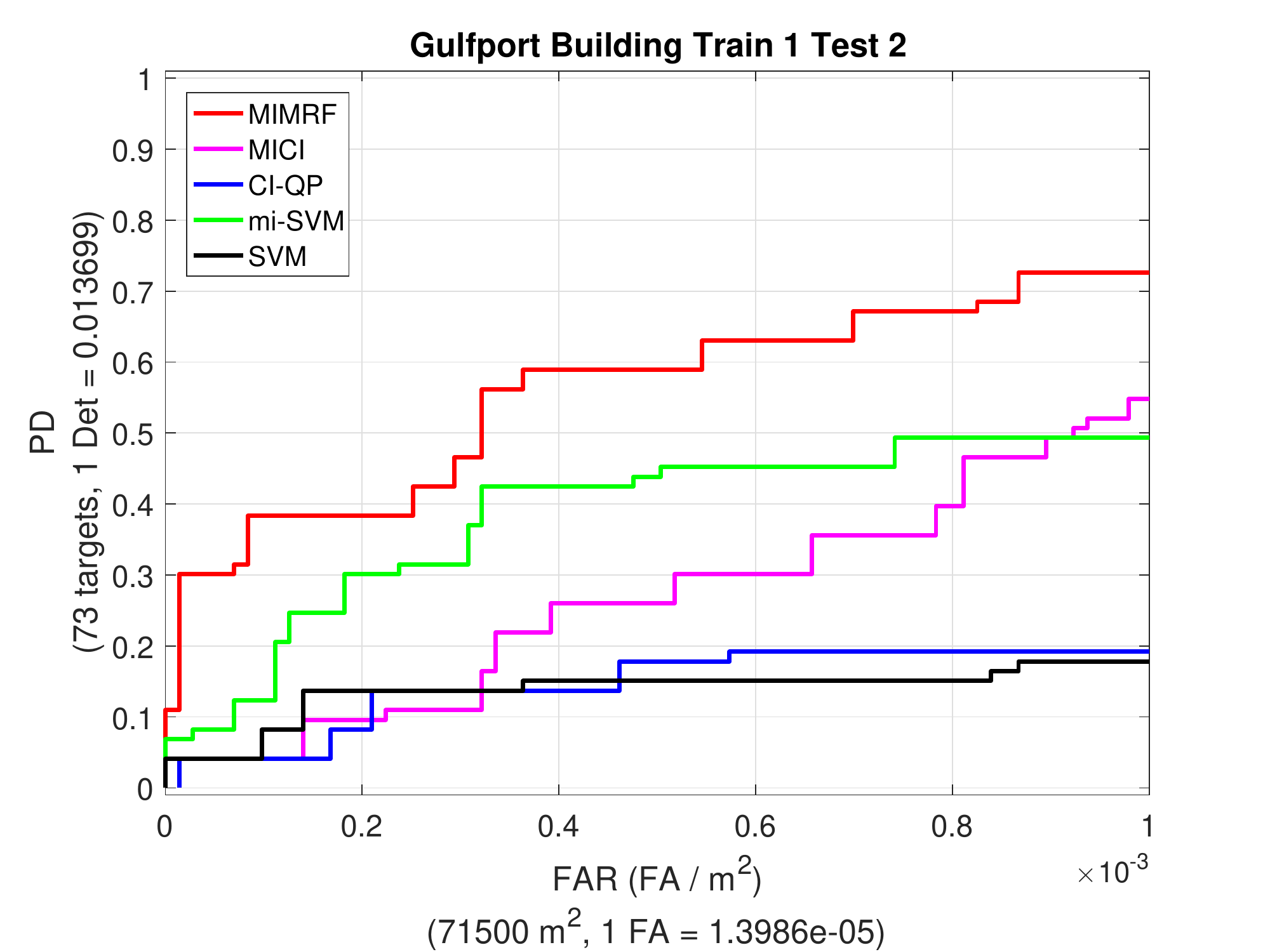}
\caption{}
\label{fig:roc_muufl_building_train1test2_max}
\end{subfigure}
\caption{The LiDAR edge map  and the scoring results on the edge map in the MUUFL Gulfport data set. (a) The differences between the rasterized LiDAR imagery and the selected lidar points by the proposed MIMRF, i.e. the LiDAR edge map. (b) One example of the ROC curve results for building detection scoring on the LiDAR edge map. }
\label{fig:diffmap_muufl_building_train2test1_lidarmean}
\end{figure}

The rasterized LiDAR imagery in our MUUFL data set was pre-processed and provided by third party companies (3001 Inc. and Optech Inc.) with nearest-neighbor-based rasterization techniques. We call the provided LiDAR rasterization ``Optech LiDAR map''. Recall that the proposed MIMRF algorithm can automatically select the ``correct'' LiDAR points for multi-resolution fusion with its objective function. Thus, we can plot the LiDAR points  selected by the proposed MIMRF into a $325\times220$ map. We call this map a ``selected LiDAR map'' by the proposed MIMRF algorithm.  Figure~\ref{fig:diffmap_muufl_building_train2test1_lidar} show the difference between the selected LiDAR map and the Optech rasterized LiDAR map. As shown, the differences are mainly along the edge areas, such as building edges or the edges of tree canopy. These  pixels on the boundary are likely where the rasterization is misaligned, due to the drastic changes in elevation between an object and its surrounding pixels. This difference map is referred to as an ``edge map'' in  the following discussions.

We also generated three alternative rasterized LiDAR maps by taking the min, max, and mean of LiDAR points within neighborhood pixels. We computed the difference between the min/max/mean LiDAR map and the given Optech LiDAR map. Any pixel that has a difference value above a  threshold is determined as ``edge'' pixels. We then look specifically at the detection results on those edge pixels. The purpose of this experiment is to show that the proposed MIMRF algorithm, which uses the raw LiDAR point cloud data, can select the correct LiDAR points from the point cloud and produce better performance especially in those edge areas when compared with other fusion methods that use the (inaccurate) rasterized imagery. 

Figure~\ref{fig:roc_muufl_building_train1test2_max} shows an example of the ROC curve results scored only on the edge pixels.  Table~\ref{table:muuflbuildingAUCedge12} and Table~\ref{table:muuflbuildingAUCedge21} present the AUC results with FAR up to $10^{-3}/m^2$ and the RMSE of the fusion methods, scored only on the edges. In the tables, the ``MIMRF diff map'' refers to results scoring on the edges determined by the difference between the LiDAR points that were selected by the MIMRF method and the Optech rasterized LiDAR imagery. The ``max/min/mean diff map'' refers to results scoring on the edges between aggregating neighborhood LiDAR points using the max/min/mean operators and the Optech rasterized LiDAR imagery. As seen from the AUC and RMSE results, the proposed MIMRF algorithm has superior performance compared with other fusion methods, specifically on the edge areas.

\begin{figure}[h]
\centering
\begin{subfigure}[t]{0.32\linewidth}
\centering
\includegraphics[width=\textwidth,trim={46mm 10mm 36mm 5mm},clip]{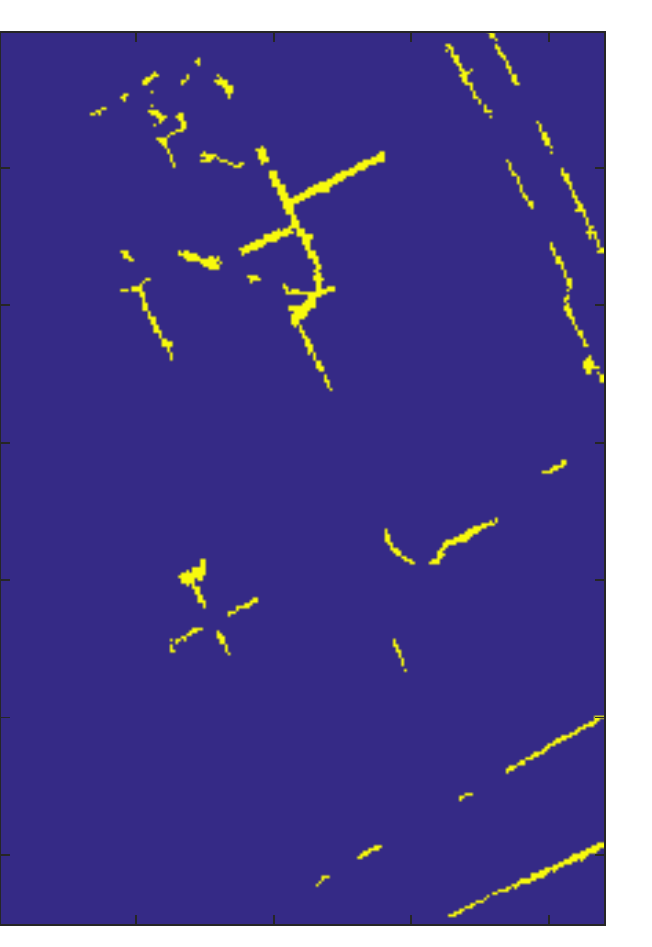}
\caption{}
\label{fig:muufl_sidewalk_GT}
\end{subfigure}
\begin{subfigure}[t]{0.32\linewidth}
\centering
\includegraphics[width=\textwidth,trim={46mm 10mm 36mm 5mm},clip]{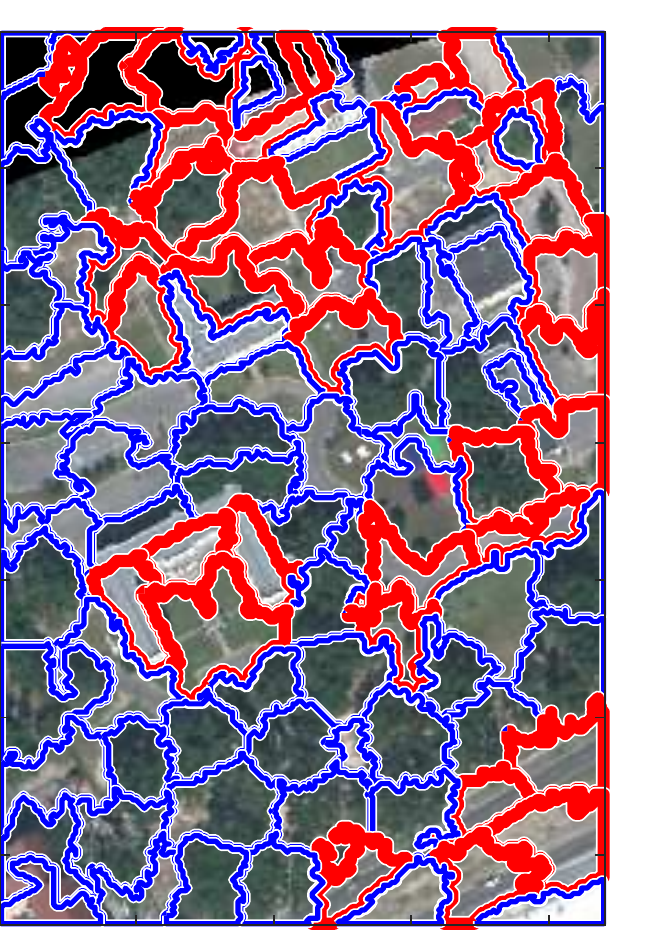}
\caption{}
\label{fig:muufl_slic_sidewalk}
\end{subfigure}
\begin{subfigure}[t]{0.32\linewidth}
\centering
\includegraphics[width=\textwidth,trim={46mm 10mm 36mm 5mm},clip]{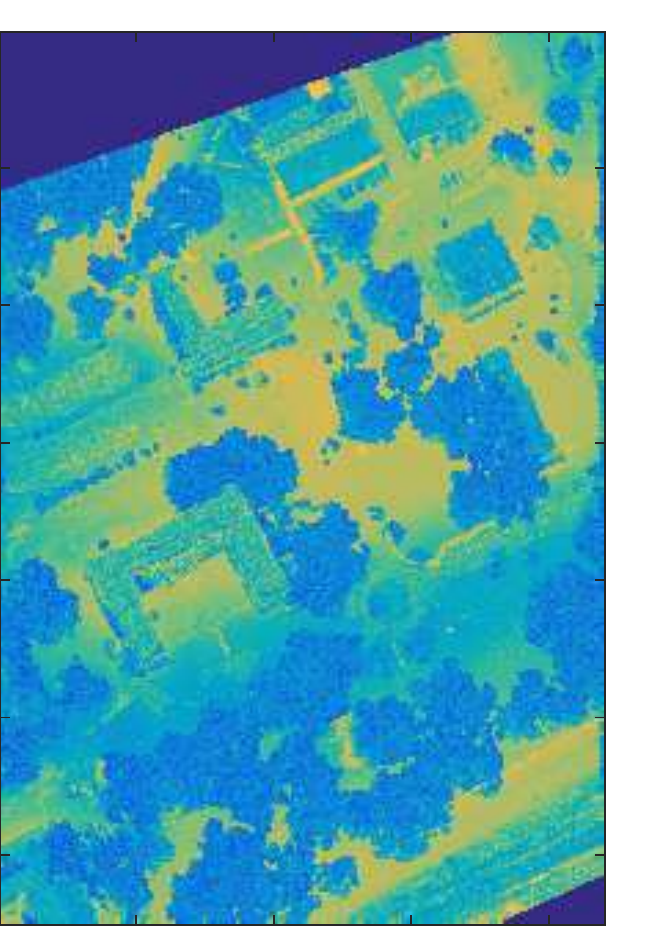}
\caption{}
\label{fig:TestConfMap_sidewalk_train1test2_MRMICI}
\end{subfigure}

\begin{subfigure}[t]{0.32\linewidth}
\centering
\includegraphics[width=\textwidth,trim={46mm 10mm 36mm 5mm},clip]{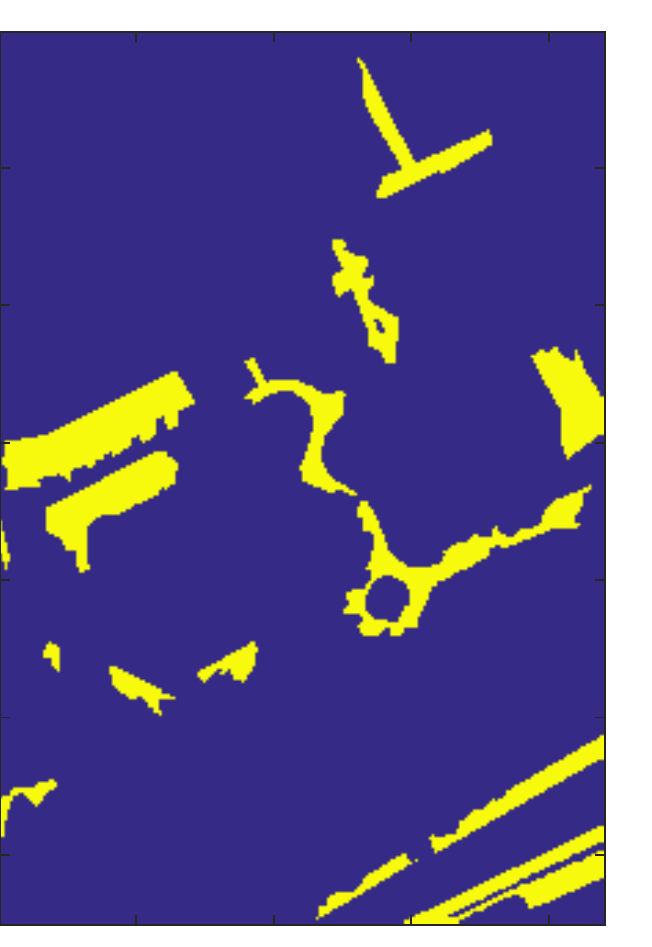}
\caption{}
\label{fig:muufl_road_GT}
\end{subfigure}
\begin{subfigure}[t]{0.32\linewidth}
\centering
\includegraphics[width=\textwidth,trim={46mm 10mm 36mm 5mm},clip]{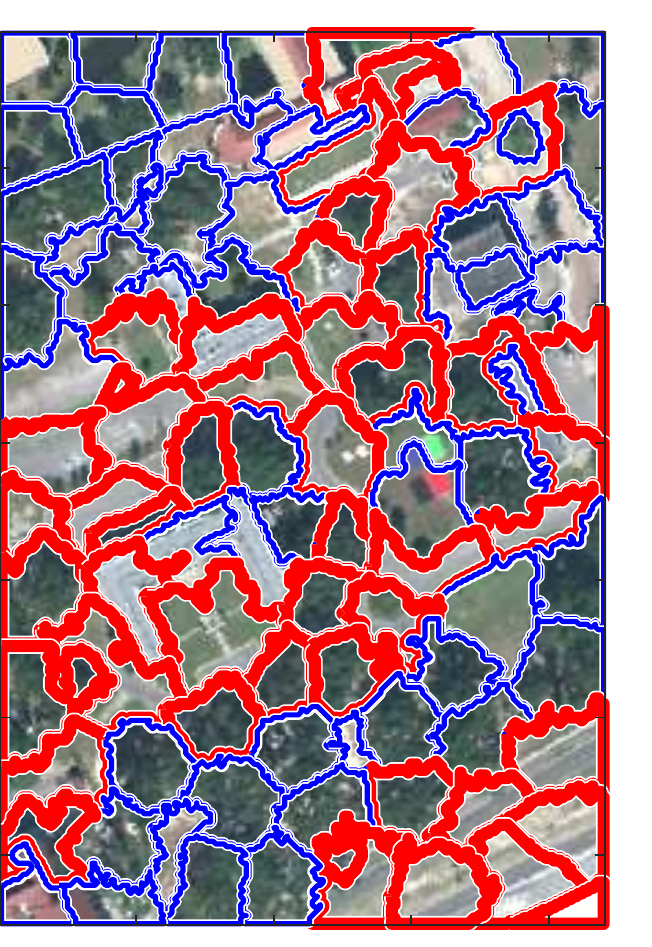}
\caption{}
\label{fig:muufl_slic_road}
\end{subfigure}
\begin{subfigure}[t]{0.32\linewidth}
\centering
\includegraphics[width=\textwidth,trim={46mm 10mm 36mm 5mm},clip]{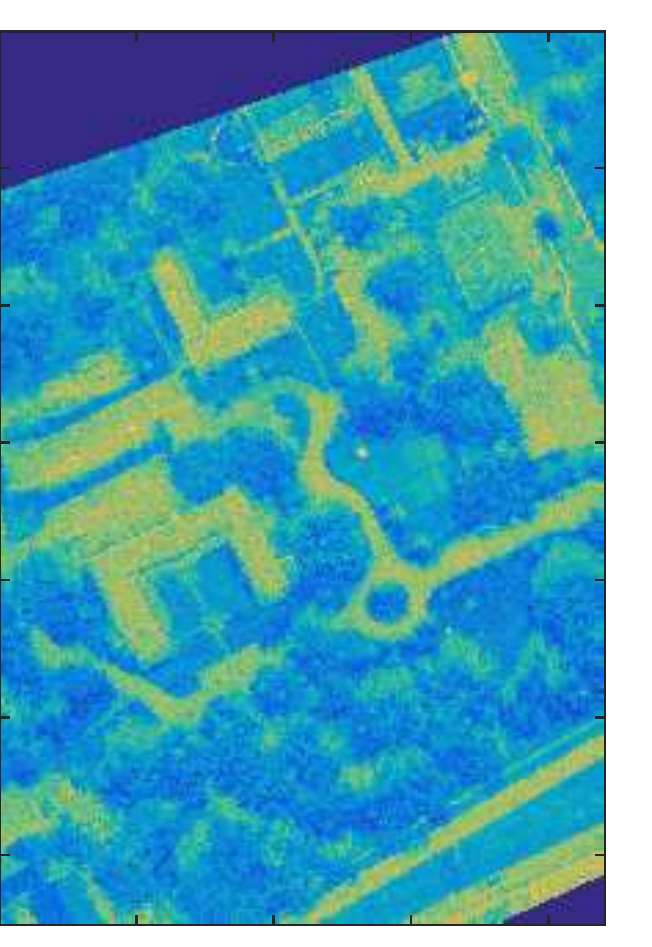}
\caption{}
\label{fig:TestConfMap_road_train1test2_MRMICI}
\end{subfigure}
\caption{Results for the MUUFL Gulfport sidewalk and road detection experiments. (a)(d) The Ground Truth map of sidewalks and roads. (b)(e) The SLIC segmentation results on sidewalks and roads. Red marks positive training bags and blue marks negative bags for road detection experiment. (c)(f) One example of the MIMRF Fusion results for sidewalk and road detection, trained on campus 1 and tested on campus 2. }
\label{fig:muufl_road_sidewalk_all}
\end{figure}

\begin{figure}[h]
\includegraphics[width=0.45\textwidth,trim={0mm 0mm 0mm 0mm},clip]{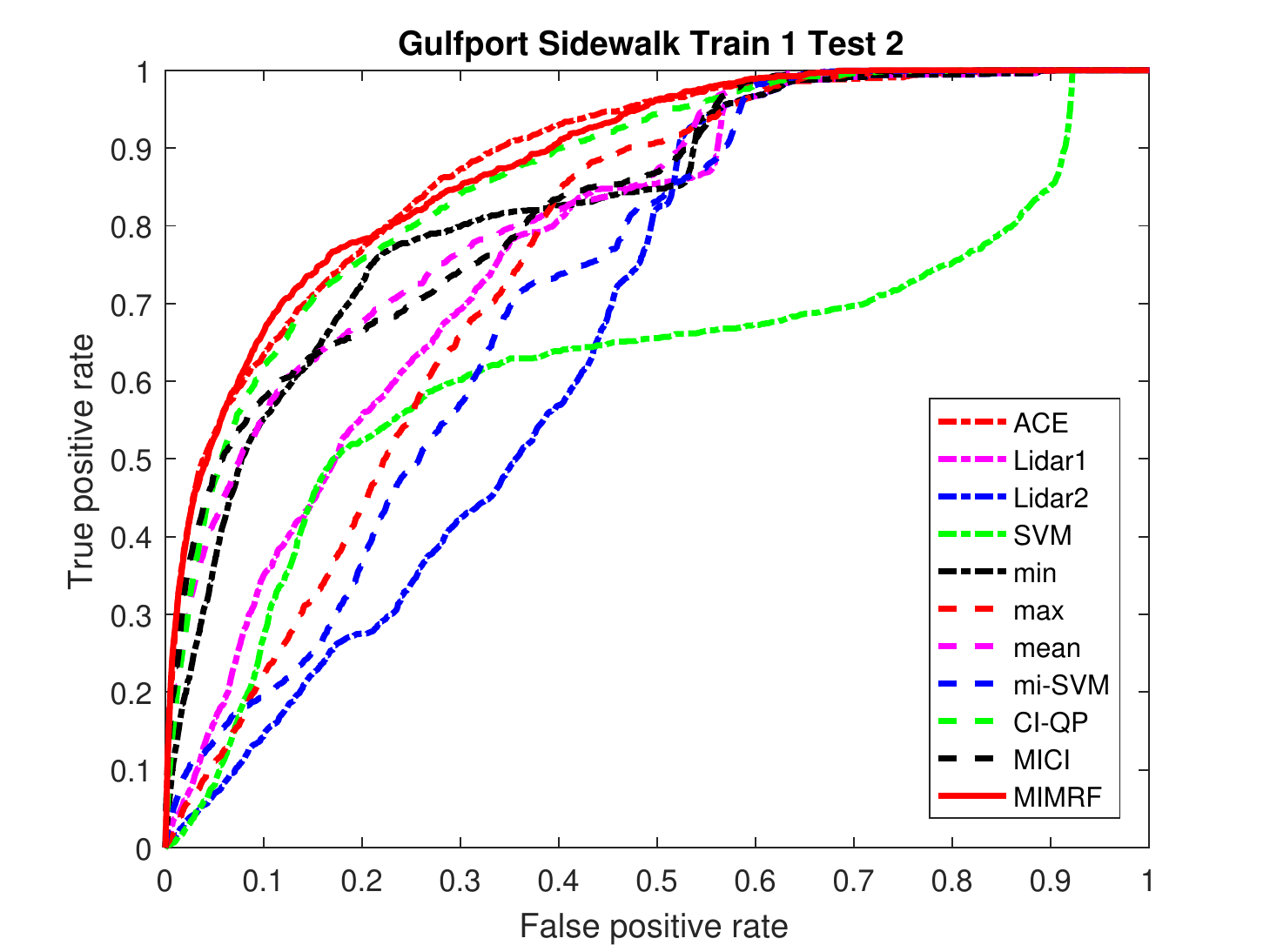}
\caption{The overall ROC curve for sidewalk detection for the MUUFL Gulfport data. Train on campus 1, test on campus 2.}
\label{fig:roc_muufl_sidewalk_train1test2_all}
\end{figure}

\begin{figure}[h]
\includegraphics[width=0.45\textwidth,trim={0mm 0mm 0mm 0mm},clip]{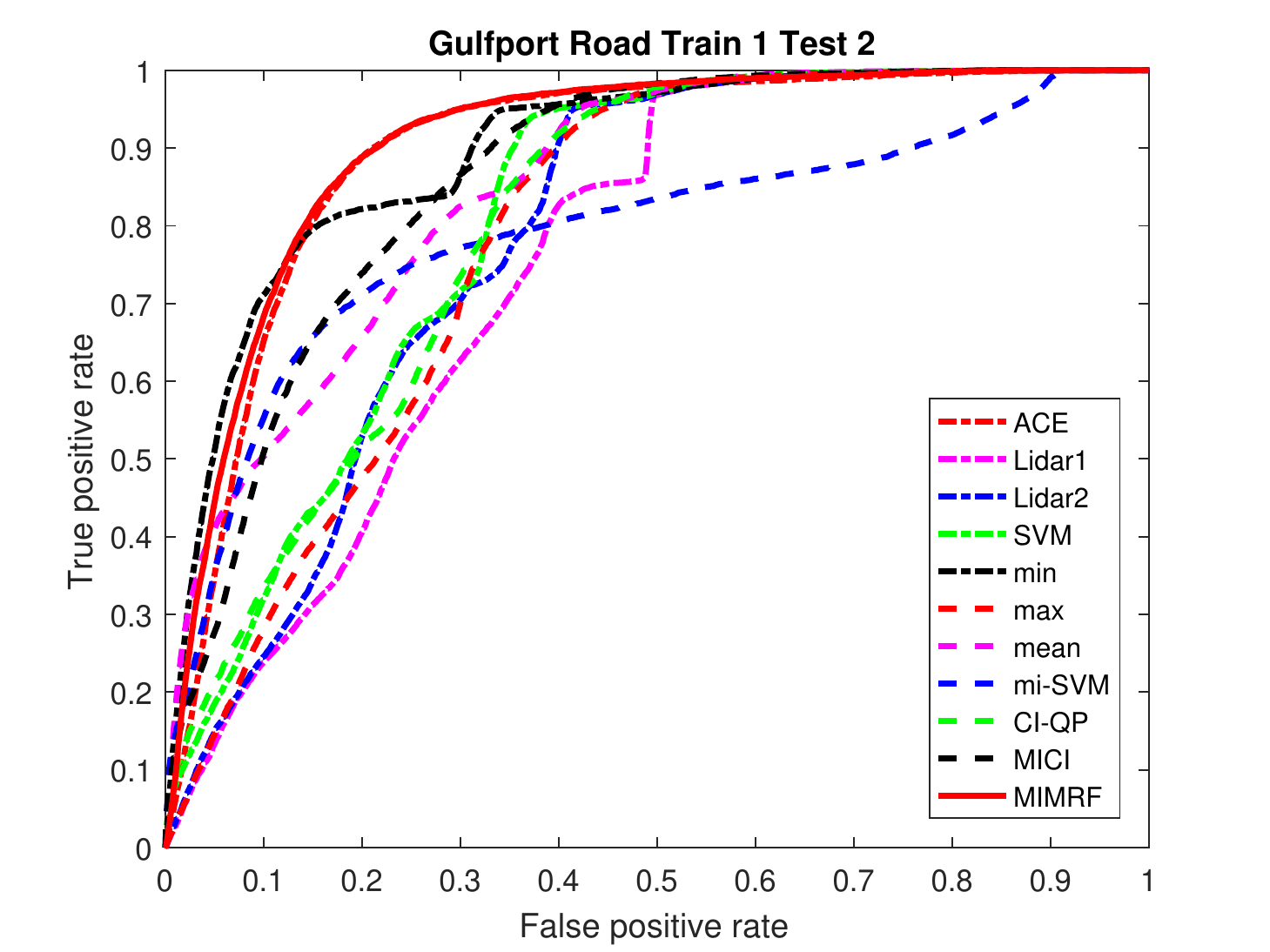}
\caption{The overall ROC curve for road detection for the MUUFL Gulfport data. Train on campus 1, test on campus 2.}
\label{fig:roc_muufl_road_train1test2_all}
\end{figure}

\begin{table*}[t]
  \centering
\caption{The AUC results of building, sidewalk, and road detection using MUUFL Gulfport HSI and LiDAR data. The best two results with the highest AUC were \textbf{bolded} and \underline{underlined}, respectively. The standard deviation is in parentheses.}
\resizebox{0.9\textwidth}{!}{
\begin{tabular}{c|cc|cc|cc}
\hline
 \multirow{2}{*}{ }  &  \multicolumn{2}{c|}{\textbf{Building Detection}}   & \multicolumn{2}{c|}{\textbf{Sidewalk Detection}}  & \multicolumn{2}{c}{\textbf{Road Detection}}\\ \hline
  & {Train1Test2} & {Train2Test1} & {Train1Test2}  & {Train2Test1} & {Train1Test2}  & {Train2Test1} \\
  \hline \hline
  ACE   & 0.906 & 0.952 & \textbf{0.882} & \textbf{0.931} & \underline{0.896} & \underline{0.902}\\ 
  LiDAR1 & 0.897& 0.880  & 0.772 & 0.769 & 0.752 & 0.748\\ 
  LiDAR2  &0.856 & 0.839  & 0.670 & 0.669 & 0.784 & 0.779\\ \hdashline
  SVM  & 0.694 &  0.738 & 0.622 & 0.663 & 0.806  & 0.396\\
  min  & 0.885 & 0.867 & 0.830 & 0.885 & \underline{0.896} & \textbf{0.918}\\
  max  & 0.943 &  0.931 & 0.754 & 0.754 & 0.785 & 0.779\\
  mean  & \textbf{0.957} &  0.953  & 0.831 &0.870   & 0.849 & 0.856\\
  mi-SVM  & 0.881  & 0.800   & 0.721 & 0.904& 0.791 & 0.817 \\ 
  CI-QP   & {0.943} & 0.931   & 0.767 & \underline{0.918} & 0.801 & 0.815\\ 
  MICI & \underline{0.952(0.000)} & \underline{0.956(0.000)}  & {0.838(0.009)}  & {0.908(0.001)} & 0.873(0.011) & 0.824(0.003)\\ 
  MIMRF & \underline{0.952(0.000)}  & \textbf{0.977(0.000)}  & \underline{0.854(0.019)} & 0.861(0.010) & \textbf{0.905(0.002)} & {0.895(0.003)}\\ 
     \hline
\end{tabular}}
  \label{table:muuflbuildingAUC}
\end{table*}

\begin{table*}[h]
\centering
\caption{The RMSE results of MICI and MIMRF on building, sidewalk, and road detection. The best results with the lower RMSE are \textbf{bolded}. The standard deviation is in parentheses. }
\resizebox{0.9\textwidth}{!}{
\begin{tabular}{c|cc|cc|cc}
\hline
 \multirow{2}{*}{ }  &  \multicolumn{2}{c|}{\textbf{Building Detection}}   & \multicolumn{2}{c|}{\textbf{Sidewalk Detection}}  & \multicolumn{2}{c}{\textbf{Road Detection}}\\ \hline
  & {Train1Test2} & {Train2Test1} & {Train1Test2}  & {Train2Test1} & {Train1Test2}  & {Train2Test1} \\
  \hline \hline
  MICI & 0.403(0.002) & 0.382(0.000) &  0.485(0.002) & \textbf{0.466(0.002)} & 0.480(0.009) & 0.514(0.002) \\ 
  MIMRF & \textbf{0.351(0.004)} & \textbf{0.331(0.001)} & \textbf{0.460(0.007)}  & 0.489(0.006)  & \textbf{0.448(0.007)} & \textbf{0.478(0.008)}\\ 
     \hline
\end{tabular}}
\label{table:muuflbuildingrmse}
\end{table*}

\begin{table*}[h]
\centering
\caption{The AUC and RMSE results of MICI and MIMRF on building detection, scored on the edge maps. Train on campus 1 and test on campus 2. The best two results with the highest AUC and lowest RMSE were \textbf{bolded} and \underline{underlined}, respectively. The standard deviation is in parentheses.}
\resizebox{\textwidth}{!}{
\begin{tabular}{c|cc|cc|cc|cc}
\hline
 \multirow{2}{*}{ }  &  \multicolumn{2}{c|}{\textbf{MIMRF diff map}}   & \multicolumn{2}{c|}{\textbf{max diff map}}  & \multicolumn{2}{c|}{\textbf{min diff map}} & \multicolumn{2}{c}{\textbf{mean diff map}}\\ \hline
   & AUC & RMSE  & AUC & RMSE  & AUC & RMSE  & AUC & RMSE\\
  \hline \hline
    SVM & 0.420 & 0.040 & 0.113 & \underline{0.067} & 0.141 & \underline{0.125}  & 0.078 & 0.063\\ 
   mi-SVM  & \underline{0.704}  & \underline{0.031} & \underline{0.327} & 0.076  & \underline{0.448} & 0.140  & \underline{0.490} & 0.071\\ 
    CI-QP & 0.329 & 0.055  & 0.135 & 0.080 & 0.126  & 0.147 & 0.115 & 0.073 \\ 
  MICI & 0.371(0.021) & 0.046(0.000) & 0.311(0.017) & 0.068(0.001)  & 0.190(0.022)& \underline{0.125(0.002)}  & 0.401(0.020)& \underline{0.062(0.001)} \\ 
  MIMRF & \textbf{0.776(0.004)} & \textbf{0.022(0.001)} & \textbf{0.458(0.022)} & \textbf{0.049(0.000)} & \textbf{0.614(0.025)} & \textbf{0.082(0.001)} & \textbf{0.619(0.027)} & \textbf{0.044(0.000)} \\ 
     \hline
\end{tabular}}
\label{table:muuflbuildingAUCedge12}
\end{table*}

\begin{table*}[h]
\centering
\caption{The AUC and RMSE results of MICI and MIMRF on building detection, scored on the edge maps. Train on campus 2 and test on campus 1.}
\resizebox{\textwidth}{!}{
\begin{tabular}{c|cc|cc|cc|cc}
\hline
 \multirow{2}{*}{ }  &  \multicolumn{2}{c|}{\textbf{MIMRF diff map}}   & \multicolumn{2}{c|}{\textbf{max diff map}}  & \multicolumn{2}{c|}{\textbf{min diff map}} & \multicolumn{2}{c}{\textbf{mean diff map}}\\ \hline
    & AUC & RMSE  & AUC & RMSE  & AUC & RMSE  & AUC & RMSE\\
  \hline \hline
    SVM & 0.513 & 0.058 & 0.413 & 0.101  & 0.537 & 0.176  & 0.390 & 0.109 \\ 
   mi-SVM  & \underline{0.695} & \textbf{0.021} & \underline{0.488} & \textbf{0.018} & \underline{0.577} & \textbf{0.031}   & \underline{0.528} & \textbf{0.017} \\ 
    CI-QP &  0.094 & 0.104 & 0.096  & 0.113 & 0.027 & 0.202   & 0.000 & 0.119 \\ 
  MICI & 0.683(0.004)  & 0.077(0.000)  & 0.451(0.008) & 0.091(0.000) & 0.375(0.009) & 0.162(0.000)  & 0.448(0.009) & 0.096(0.000)\\ 
  MIMRF & \textbf{0.794(0.007)}  & \underline{0.035(0.000)} & \textbf{0.529(0.003)} & \underline{0.061(0.000)} & \textbf{0.649(0.007)} & \underline{0.103(0.000)}  & \textbf{0.638(0.005)} & \underline{0.064(0.000)} \\ 
     \hline
\end{tabular}}
\label{table:muuflbuildingAUCedge21}
\end{table*}

\subsubsection{MUUFL Gulfport Sidewalk and Road Detection} 
We conducted similar experiments on other materials in the scene and present sidewalk and road detection results in this section. Sidewalks and roads are on ground surface level and do not have drastic altitude change as building edges.  These additional experiments on sidewalk and road show the effectiveness of the proposed MIMRF algorithm in understanding other materials and objects in the scene.

Figure~\ref{fig:muufl_road_sidewalk_all} shows the ground truth map and SLIC segmentation results for sidewalk and road detection experiments. Figure~\ref{fig:TestConfMap_sidewalk_train1test2_MRMICI} and Figure~\ref{fig:TestConfMap_road_train1test2_MRMICI} show the MIMRF fusion results for sidewalk and road detection. Figure~\ref{fig:roc_muufl_sidewalk_train1test2_all} and Figure~\ref{fig:roc_muufl_road_train1test2_all} show the cross-validated overall ROC curve results on sidewalk and road detection. The complete AUC results can be seen in Table~\ref{table:muuflbuildingAUC}, which shows that the proposed MIMRF algorithm can successfully detect sidewalks and roads in the scene as well. Note that since sidewalks and roads have similar altitude with ground surface, the LiDAR sources do not have significant impact. That is why the ACE map from the HSI data has comparable AUC performance to the fusion results in sidewalk and road. Still, the MIMRF achieved best or second best in half of the experiments.


\subsection{Soybean and Weed Data Set} 

The proposed MIMRF algorithm was originally motivated by the HSI/LiDAR fusion problem for scene understanding. However, the proposed MIMRF can also be used as a general multi-resolution fusion framework for many applications. This section provides additional experimental results of the proposed MIMRF on an agricultural data set.

\begin{figure}[t!]
\centering
\includegraphics[width=0.45\textwidth,trim={65mm 85mm 40mm 70mm},clip]{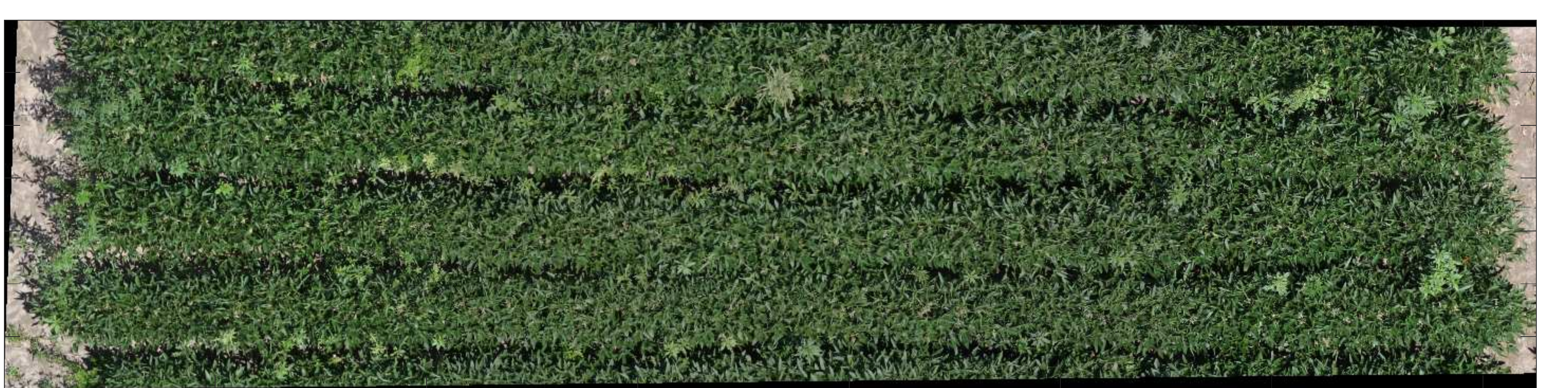}
\caption[The RGB image of weed in the soybean-weed data.]{The RGB image of  the soybean-weed data.}
\label{fig:soybeanweed_RGB}
\end{figure}

The proposed MIMRF was used to detect weed in a remotely sensed, multi-resolution soybean and weed data set. In the data set, a height map and a RGB image are provided over a patch of soybean field and the goal is to detect weed amongst the soybean plants. The height map is $351\times1450$ in size and the RGB map is $1404\times5864$ in size. Figure~\ref{fig:soybeanweed_RGB} shows the RGB map over the scene.  

\begin{figure}[t!]
\centering
\includegraphics[width=0.45\textwidth,trim={65mm 85mm 65mm 70mm},clip]{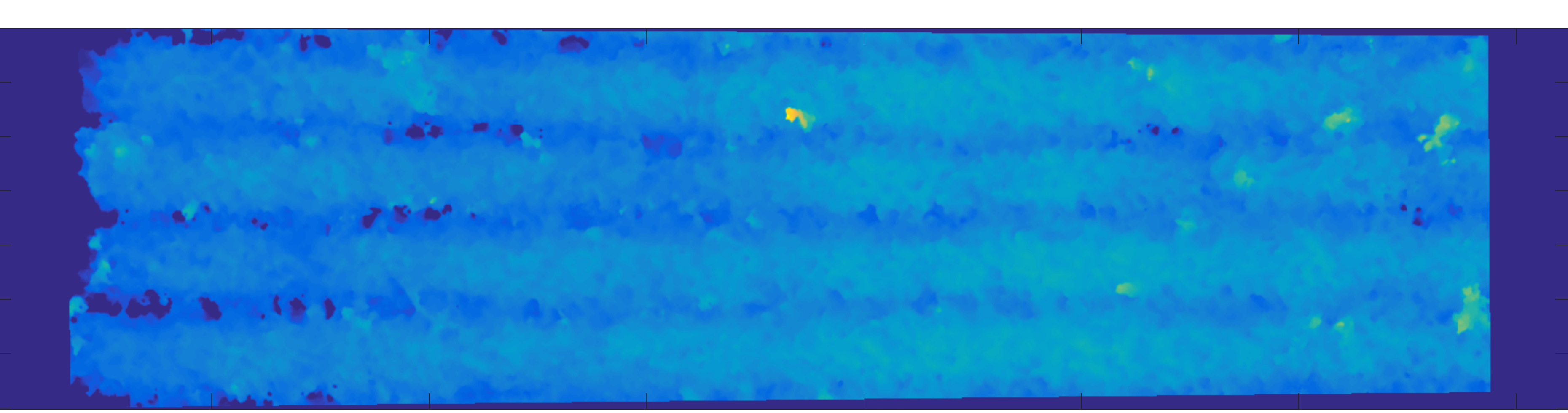}
\caption[The height map of the soybean-weed data.]{The height map of the soybean-weed data.}
\label{fig:soybeanweed_heightmap}
\end{figure}

\begin{figure}[!h]
\centering
\begin{subfigure}[t]{\linewidth}
\centering
\includegraphics[width=0.9\textwidth,trim={65mm 85mm 45mm 70mm},clip]{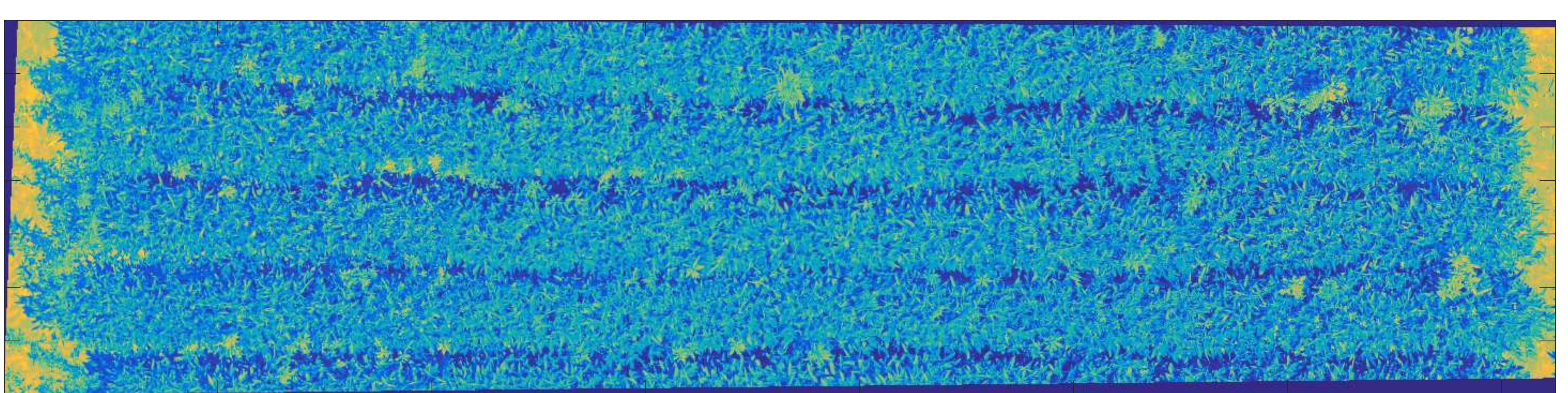}
\caption{ }
\label{fig:soybeanweed_L_feature}
\end{subfigure}
\begin{subfigure}[t]{\linewidth}
\centering
\includegraphics[width=0.9\textwidth,trim={65mm 85mm 45mm 70mm},clip]{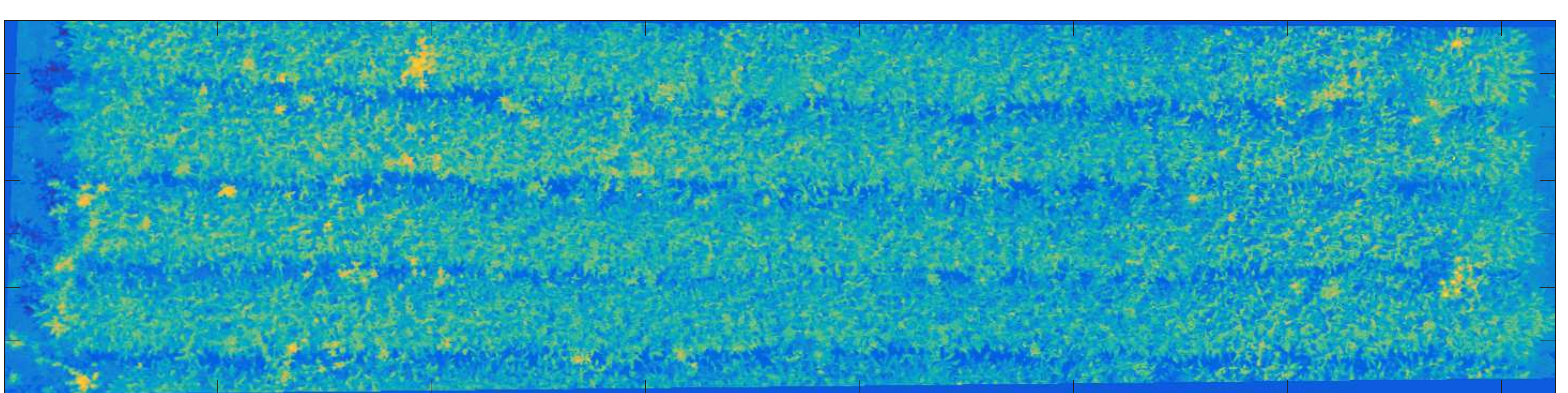}
\caption{ }
\label{fig:soybeanweed_B_feature}
\end{subfigure}
\begin{subfigure}[t]{\linewidth}
\centering
\includegraphics[width=0.9\textwidth,trim={65mm 85mm 45mm 70mm},clip]{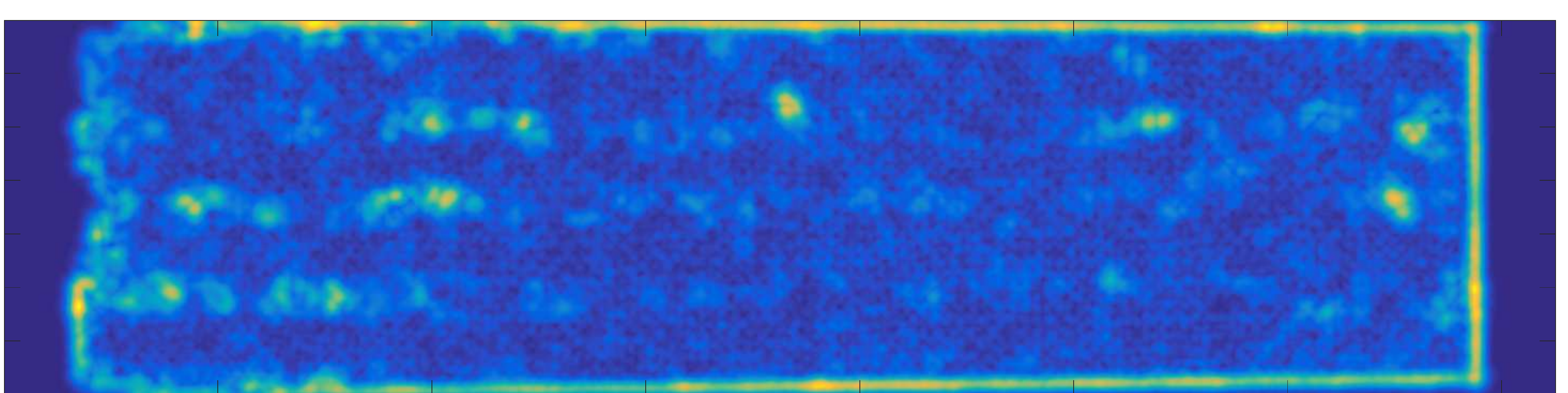}
\caption{ }
\label{fig:soybeanweed_height_gabor}
\end{subfigure}
\caption{Three fusion sources of the soybean-weed data set. (a) The L-band image of the soybean-weed data. (b) The B-band image of the soybean-weed data. (c) The Gabor filtered image of the soybean-weed data height map.}
\label{fig:soybeanweed_sources}
\end{figure}

Three fusion sources were used for the proposed algorithm for weed detection. Figure~\ref{fig:soybeanweed_heightmap} shows the height map over the soybean-weed field. Some weeds in the scene are slightly higher than soybean plants, indicating that height is an important feature for weed detection. To extract height features, four Gabor filters at angles $0^{\circ}$, $45^{\circ}$,  $90^{\circ}$, and  $135^{\circ}$  were applied to the height map and the sum of the filtered images was used as one of the fusion sources, as shown in Figure~\ref{fig:soybeanweed_height_gabor}.

We also observed that, in this data set, the weed plants have lighter-colored pixels than the soybean plants. That is to say, lightness and color are useful sources of weed detection as well. We transformed the RGB values into LAB color space and the L- and B- band imagery were used as the other two sources for fusion. The L dimension provides information about lightness and the B dimension is the color opponent for  the blue-yellow space. Figure~\ref{fig:soybeanweed_L_feature} and Figure~\ref{fig:soybeanweed_B_feature} show the L and B band images, where weed pixels are highlighted. 

\begin{figure}[!h]
\centering
\begin{subfigure}[t]{\linewidth}
\centering
\includegraphics[width=0.9\textwidth,trim={65mm 85mm 45mm 70mm},clip]{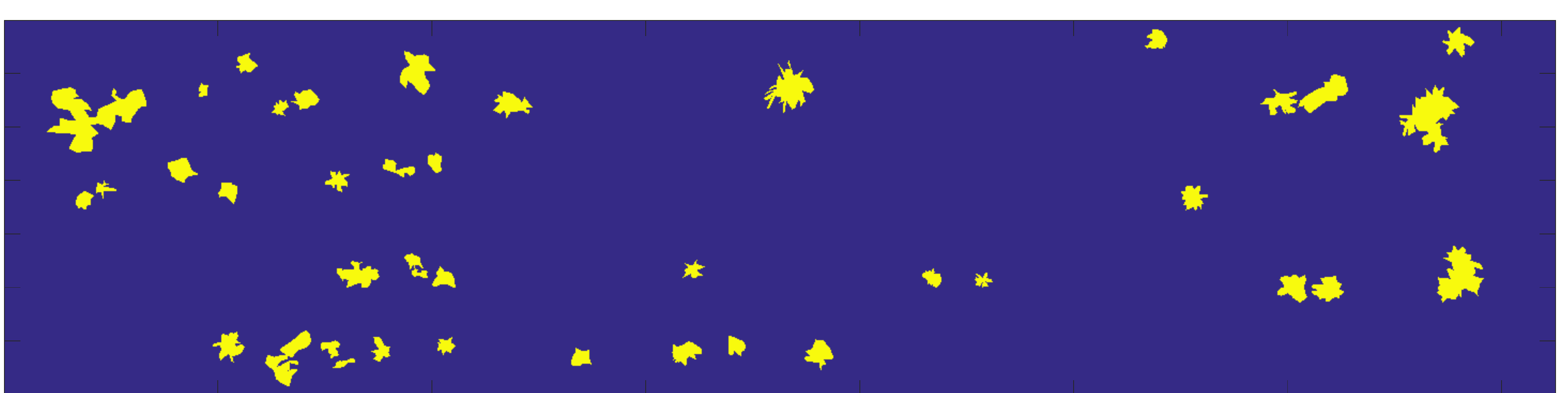}
\caption{ }
\label{fig:soybeanweed_GT}
\end{subfigure}
\begin{subfigure}[t]{\linewidth}
\centering
\includegraphics[width=0.9\textwidth,trim={65mm 85mm 45mm 70mm},clip]{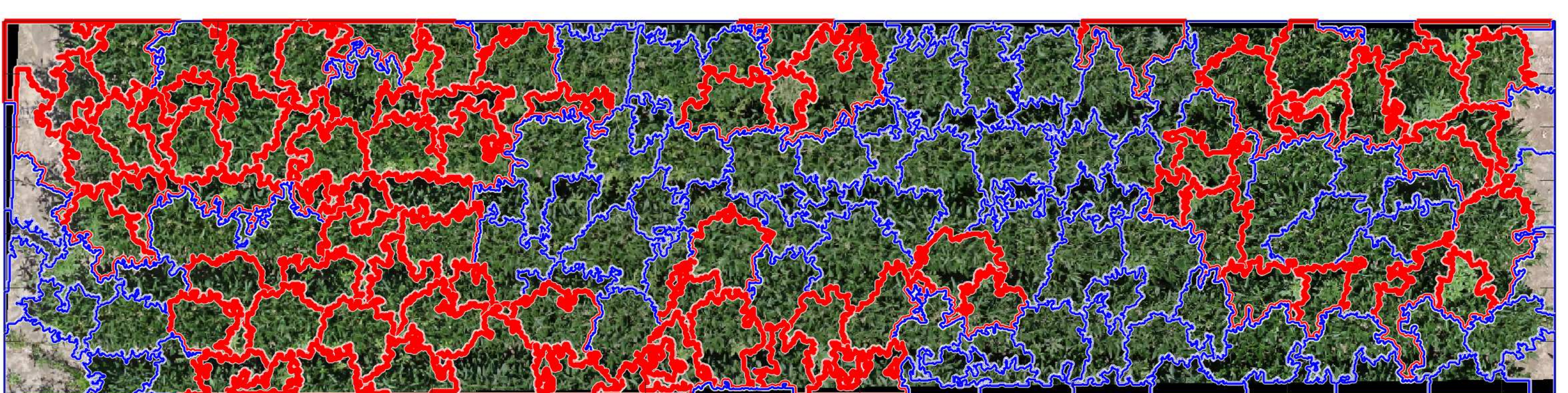}
\caption{ }
\label{fig:soybeanweed_slic}
\end{subfigure}
\begin{subfigure}[t]{\linewidth}
\centering
\includegraphics[width=0.9\textwidth,trim={65mm 85mm 45mm 70mm},clip]{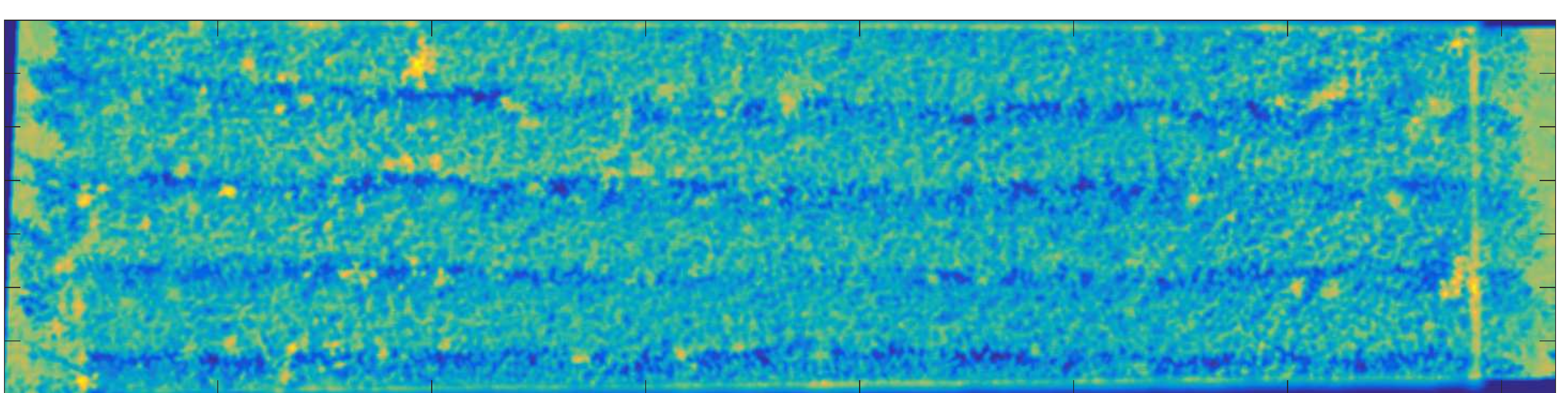}
\caption{ }
\label{fig:TestConfMap_soybean_MRMICI}
\end{subfigure}
\caption{(a) The Ground Truth map of weed in the soybean-weed data. The deep blue is the background (soybean plants) and the yellow marks the target (weed). (b) The SLIC segmentation map of the soybean-weed data. Red marks positive training bags and blue marks negative bags for weed detection experiment. (c) The confidence map obtained from the MIMRF fusion for the soybean-weed data.}
\label{fig:soybeanweed_test}
\end{figure}

\begin{figure}[h]
\centering
\includegraphics[width=0.45\textwidth,trim={0mm 0mm 0mm 0mm},clip]{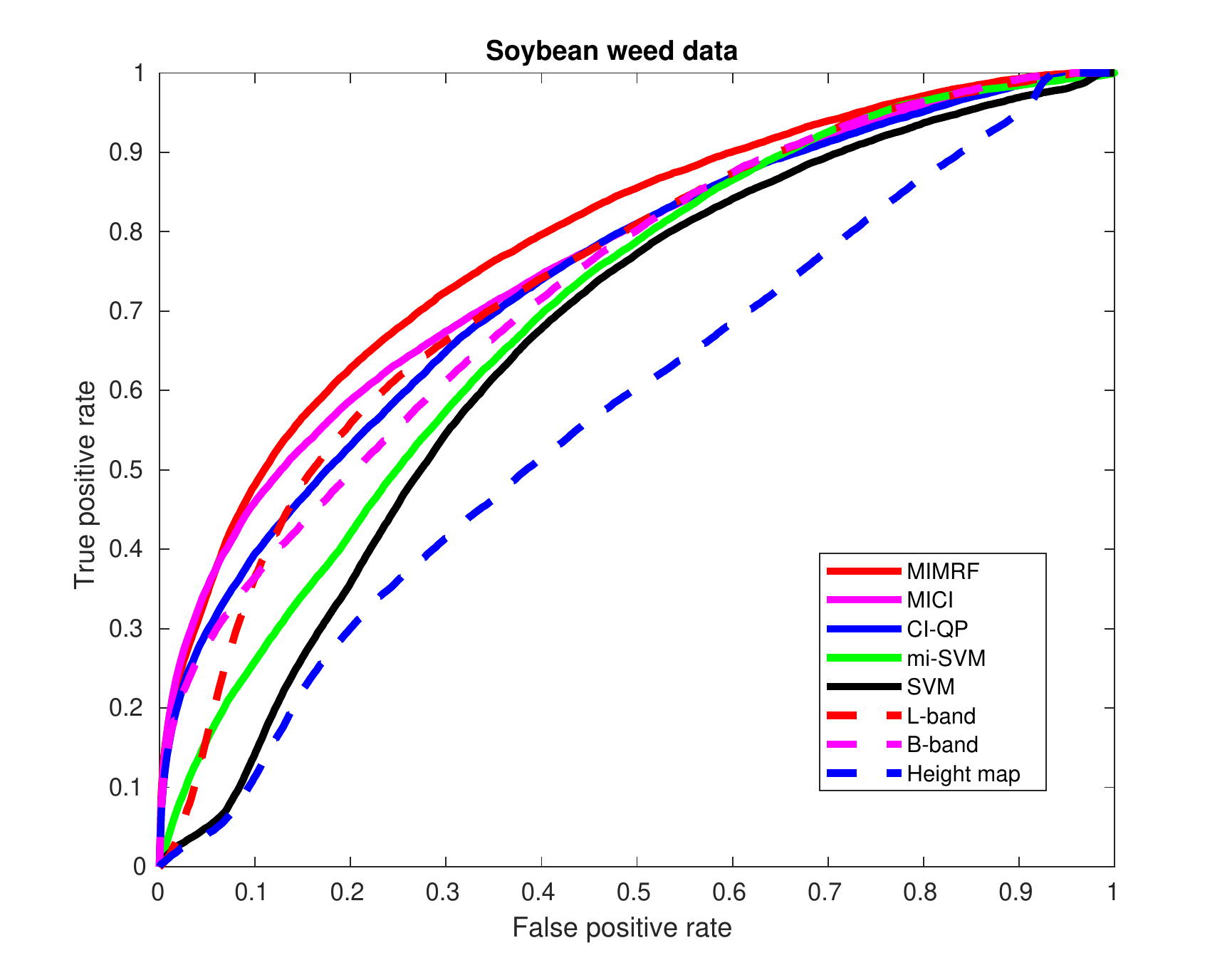}
\caption{The overall ROC curve results for weed detection in the soybean-weed data across comparison methods. }
\label{fig:soybeanweed_roc_all}
\end{figure}

Figure~\ref{fig:soybeanweed_GT} shows the manual ground truth map for weed in this data set. Figure~\ref{fig:soybeanweed_slic} shows the SLIC segmentation results. Similar to Section~\ref{sec:autotrainlabelgen}, each superpixel in this segmentation was regarded as a ``bag'' and the colors mark the bag-level training labels. Figure~\ref{fig:TestConfMap_soybean_MRMICI} shows the fusion results of the proposed MIMRF fusion, where the bright yellow highlights the detected weed. Figure~\ref{fig:soybeanweed_roc_all} shows the overall ROC curve results for all comparison methods. As shown, the proposed MIMRF method can effectively detect weed in the scene and produce an overall better ROC curve performance for weed detection.

\subsection{Discussions}
\label{sec:discussions}

\textcolor{black}{There are a few observations from the experiments regarding the assumptions, limitations, and applications of our proposed MIMRF algorithm. First, our proposed MIMRF algorithm operates under the assumption that there is at least one LiDAR point that corresponds to the hyperspectral pixel, since its purpose was to fuse LiDAR and HSI in the MUUFL Gulfport HSI/LiDAR fusion experiment (if there was zero LiDAR point, there would be nothing to fuse). In our experiments, this assumption is true, since the hyperspectral imagery in the MUUFL Gulfport dataset was collected natively on a pixel grid with an $1m$ ground sample distance, whereas the raw LiDAR data is a point cloud with a higher resolution of 0.60m cross track and 0.78m along track spot spacing. There certainly can exist (rarely, we argue) in other datasets where there is zero LiDAR point in a hyperspectral pixel. In this case, the Choquet integral, by its original definition, would not work since there would not be sufficient sources to ``fuse''. However, we can account for this edge case by either manually defining that if there is zero LiDAR point in a hyperspectral pixel, the Choquet integral ``fusion'' result will be equal to the confidence value of the hyperspectral pixel, or assigning a LiDAR point value to the empty element such as using the average of neighboring LiDAR point values. This way, the math for the MIL bag-level labels in our proposed MIMRF algorithm in \eqref{eq:multiresobjneg}, \eqref{eq:multiresobjpos}, and \eqref{eq:objlnmultiresminmax} can still work in the same manner. }

\textcolor{black}{Second, the MIL framework is useful in modeling problems where the positive bags contain relatively few pixels of target, which is quite common in remote sensing applications. For example, for pixel-size and sub-pixel-size target detection applications, the target size can be only a few pixels or less. In our building detection experiments, some superpixels may contain only a few pixels of building, especially along the edge areas. With the MIL framework, annotators can easily circle a large region and say, ``there is a target in here somewhere but we do not know the exact location.'' In these cases, there will usually be more negative (background) points in a positive bag than the target, and MIL-based methods, such as the proposed MIMRF algorithm in this paper, can easily work with such imprecise labels while non-MIL methods will have difficulty.  Quantitative results on the relationship between bag structure, such as noise, contamination, and percentage of primary instances, and our MIL-based fusion methods were presented in Section V-A and V-B of our previous work \cite{du2018multiple}.}

\textcolor{black}{Finally, it is worth noting that our proposed MIMRF algorithm can act as a method for both tweaking up imprecisely labeled data and being applied out-of-sample. The MIMRF can be used as a method for generating pixel-level labels for imprecisely labeled data, as shown in the soybean-weed dataset experiments. Additionally, it can be applied out-of-sample (e.g., to another patch of weedy soybeans) as long as the data to be fused are from the same set of sensors/classifiers and that the sensors/classifiers have the same relationship as in training. This is because the learned $g^*$ represents the relationship between the sensors/classifiers used. Here is an easy example to illustrate this point. Let us assume for a certain problem the estimated measure $g^*=[g_1,g_2,g_3,g_{12},g_{13},g_{23},g_{123}]=[0,0,0,1,0,0,1]$ for three sources. This means the fusion result is the intersection of the source 1 and source 2, since $g_{12}=1$. Therefore, this $g^*$ can be applied to any other problem as long as the desired fusion result is the intersection of source 1 and source 2. Our MUUFL Gulfport experiments illustrated this point by testing on a different flight -- as long as the fusion sources (HSI and LiDAR confidence maps) have the same relationships as in training, the method can work well.}

\section{Conclusion}
\label{sec:conclusion}

In this paper, we proposed a novel Multiple Instance Multi-Resolution Fusion (MIMRF) framework that can perform multi-resolution, multi-modal sensor fusion on remote sensing data with label uncertainty. The proposed MIMRF is novel and unique in that it directly fuses multi-resolution, multi-modal sensor outputs. Specifically, in the HSI/LiDAR fusion problem,  \textcolor{black}{the proposed MIMRF method can \emph{automatically} select accurate LiDAR points from the raw, non-grid LiDAR point cloud using the Choquet integral and then use such selected points to perform fusion in an end-to-end manner. In a way, such selection of LiDAR points may be viewed as a special, autonomous rasterization technique without enforcing an overall assumption of nearest neighbor.}  Experimental results have shown superior performance especially at edge areas where tradition rasterization methods are likely to be wrong. Additionally, the proposed MIMRF does not require accurate pixel-wise training labels. Instead, the MIMRF method can learn directly from bag-level imprecise labels and produce pixel-level fusion results. Instead of the traditional manual labeling process, we show satisfactory results by simply using publicly available, crowd-sourced data such as the Open Street Map to automatically generate imprecise bag-level labels for the proposed algorithm before fusion.  Thus, we envision others being able to use the proposed self-supervised learning approach to perform pixel-level classification and produce fusion maps with minimal human intervention for a variety of multi-resolution fusion applications. 

Although the proposed MIMRF was originally motivated by the hyperspectral imagery and LiDAR point cloud fusion problem in remote sensing, the method is a general framework can be applied to many multi-resolution and multi-modal fusion applications with uncertian labels, such as precision agriculture. Future work will include investigating and incorporating alternative fusion sources and features, such as texture-based LiDAR features, and applying the proposed MIMRF algorithm to other fusion applications. \textcolor{black}{Additionally, future investigation into other types of fuzzy measures, such as the binary fuzzy measure \cite{anderson2017binary, islam2018efficient, du2019ssci}, can be conducted to compress the number of independent parameters in the fuzzy measure $\mathbf{g}$ for more efficient learning.}

%


%



\ifCLASSOPTIONcaptionsoff
  \newpage
\fi



%

%
\bibliographystyle{IEEEtran} 
\bibliography{sample}

\begin{thebibliography}{10}
\providecommand{\url}[1]{#1}
\csname url@samestyle\endcsname
\providecommand{\newblock}{\relax}
\providecommand{\bibinfo}[2]{#2}
\providecommand{\BIBentrySTDinterwordspacing}{\spaceskip=0pt\relax}
\providecommand{\BIBentryALTinterwordstretchfactor}{4}
\providecommand{\BIBentryALTinterwordspacing}{\spaceskip=\fontdimen2\font plus
\BIBentryALTinterwordstretchfactor\fontdimen3\font minus
  \fontdimen4\font\relax}
\providecommand{\BIBforeignlanguage}[2]{{%
\expandafter\ifx\csname l@#1\endcsname\relax
\typeout{** WARNING: IEEEtran.bst: No hyphenation pattern has been}%
\typeout{** loaded for the language `#1'. Using the pattern for}%
\typeout{** the default language instead.}%
\else
\language=\csname l@#1\endcsname
\fi
#2}}
\providecommand{\BIBdecl}{\relax}
\BIBdecl

\bibitem{gader2004multi}
P.~Gader, A.~Mendez-Vasquez, K.~Chamberlin, J.~Bolton, and A.~Zare,
  ``Multi-sensor and algorithm fusion with the {Choquet} integral: applications
  to landmine detection,'' in \emph{IEEE Int. Geosci. Remote Sens. Symp.
  (IGARSS)}, vol.~3, Sept. 2004, pp. 1605--1608.

\bibitem{gader2013muufl}
P.~Gader, A.~Zare, R.~Close, J.~Aitken, and G.~Tuell, ``{MUUFL Gulfport
  Hyperspectral and LiDAR Airborne Data Set},'' University of Florida,
  Gainesville, FL, Tech. Rep. Rep. REP-2013-570, Oct. 2013.

\bibitem{du2017technical}
X.~Du and A.~Zare, ``{Technical report: scene label ground truth map for MUUFL
  Gulfport Data Set},'' University of Florida, Gainesville, FL., Tech. Rep.
  Tech. Rep. 20170417, 2017.

\bibitem{pohl1998review}
C.~Pohl and J.~L.~V. Genderen, ``Multisensor image fusion in remote sensing:
  Concepts, methods and applications,'' \emph{Int. J. Remote Sens.}, vol.~19,
  no.~5, pp. 823--854, 1998.

\bibitem{liggins2008handbook}
M.~Liggins~II, D.~Hall, and J.~Llinas, \emph{Handbook of multisensor data
  fusion}.\hskip 1em plus 0.5em minus 0.4em\relax CRC Press, 2008.

\bibitem{zhang2010multi}
J.~Zhang, ``Multi-source remote sensing data fusion: status and trends,''
  \emph{Int. J. Image and Data Fusion}, vol.~1, no.~1, pp. 5--24, 2010.

\bibitem{bioucas2013hyperspectral}
J.~M. Bioucas-Dias, A.~Plaza, G.~Camps-Valls, P.~Scheunders, N.~Nasrabadi, and
  J.~Chanussot, ``Hyperspectral remote sensing data analysis and future
  challenges,'' \emph{IEEE Geosci. Remote Sens. Mag.}, vol.~1, no.~2, pp.
  6--36, 2013.

\bibitem{li2010extracting}
Y.~Li and E.~B. Olson, ``Extracting general-purpose features from {LiDAR}
  data,'' in \emph{IEEE Int. Conf. Robotics and Automation (ICRA)}, May 2010,
  pp. 1388--1393.

\bibitem{khodadadzadeh2015fusion}
M.~Khodadadzadeh, J.~Li, S.~Prasad, and A.~Plaza, ``Fusion of hyperspectral and
  {LiDAR} remote sensing data using multiple feature learning,'' \emph{IEEE J.
  Sel. Topics. Appl. Earth Observ.}, vol.~8, no.~6, pp. 2971--2983, 2015.

\bibitem{rasti2017fusion}
B.~Rasti, P.~Ghamisi, J.~Plaza, and A.~Plaza, ``Fusion of hyperspectral and
  {LiDAR} data using sparse and low-rank component analysis,'' \emph{IEEE
  Trans. Geosci. Remote Sens.}, vol.~55, no.~11, pp. 6354--6365, 2017.

\bibitem{luo2017fusion}
R.~Luo, W.~Liao, H.~Zhang, L.~Zhang, P.~Scheunders, Y.~Pi, and W.~Philips,
  ``Fusion of hyperspectral and {LiDAR} data for classification of cloud-shadow
  mixed remote sensed scene,'' \emph{IEEE J. Sel. Topics. Appl. Earth Observ.},
  vol.~10, no.~8, pp. 3768--3781, 2017.

\bibitem{sampath2010segmentation}
A.~Sampath and J.~Shan, ``Segmentation and reconstruction of polyhedral
  building roofs from aerial {LiDAR} point clouds,'' \emph{IEEE Trans. Geosci.
  Remote Sens.}, vol.~48, no.~3, pp. 1554--1567, Mar. 2010.

\bibitem{cao2014astable}
S.~Cao, X.~Zhu, Y.~Pan, and Q.~Yu, ``A stable land cover patches method for
  automatic registration of multitemporal remote sensing images,'' \emph{IEEE
  J. Sel. Topics. Appl. Earth Observ.}, vol.~7, no.~8, pp. 3502--3512, Aug
  2014.

\bibitem{brigot2016adaptation}
G.~Brigot, E.~Colin-Koeniguer, A.~Plyer, and F.~Janez, ``Adaptation and
  evaluation of an optical flow method applied to coregistration of forest
  remote sensing images,'' \emph{IEEE J. Sel. Topics. Appl. Earth Observ.},
  vol.~9, no.~7, pp. 2923--2939, 2016.

\bibitem{li2008adaptive}
Y.~Li and H.~Wu, ``Adaptive building edge detection by combining {LiDAR} data
  and aerial images,'' \emph{The International Archives of the Photogrammetry,
  Remote Sensing and Spatial Information Sciences}, vol.~37, no. Part B1, pp.
  197--202, 2008.

\bibitem{dietterich1997solving}
T.~G. Dietterich, R.~H. Lathrop, and T.~Lozano-P{\'e}rez, ``Solving the
  multiple instance problem with axis-parallel rectangles,'' \emph{Artif.
  Intell.}, vol.~89, no. 1-2, pp. 31--71, Jan. 1997.

\bibitem{yang2010decision}
H.~Yang, Q.~Du, and B.~Ma, ``Decision fusion on supervised and unsupervised
  classifiers for hyperspectral imagery,'' \emph{IEEE Trans. Geosci. Remote
  Sens. Lett.}, vol.~7, no.~4, pp. 875--879, Oct. 2010.

\bibitem{dalponte2008fusion}
M.~Dalponte, L.~Bruzzone, and D.~Gianelle, ``Fusion of hyperspectral and
  {LIDAR} remote sensing data for classification of complex forest areas,''
  \emph{IEEE Trans. Geosci. Remote Sens.}, vol.~46, no.~5, pp. 1416--1427,
  2008.

\bibitem{shen2016an}
H.~Shen, X.~Meng, and L.~Zhang, ``An integrated framework for the
  spatio-temporal-spectral fusion of remote sensing images,'' \emph{IEEE Trans.
  Geosci. Remote Sens.}, vol.~54, no.~12, pp. 7135--7148, 2016.

\bibitem{hinz2003automatic}
S.~Hinz and A.~Baumgartner, ``Automatic extraction of urban road networks from
  multi-view aerial imagery,'' \emph{ISPRS J. Photogrammetry and Remote
  Sensing}, vol.~58, no.~1, pp. 83--98, 2003.

\bibitem{stankov2014detection}
K.~Stankov and D.-C. He, ``Detection of buildings in multispectral very high
  spatial resolution images using the percentage occupancy hit-or-miss
  transform,'' \emph{IEEE J. Sel. Topics. Appl. Earth Observ.}, vol.~7, no.~10,
  pp. 4069--4080, 2014.

\bibitem{king2000challenge}
R.~L. King, ``A challenge for high spatial, spectral, and temporal resolution
  data fusion,'' in \emph{IEEE Int. Geosci. Remote Sens. Symp. (IGARSS)},
  vol.~6, 2000, pp. 2602--2604.

\bibitem{lin2011combining}
A.~Y.-M. Lin, A.~Novo, S.~Har-Noy, N.~D. Ricklin, and K.~Stamatiou, ``Combining
  geoeye-1 satellite remote sensing, uav aerial imaging, and geophysical
  surveys in anomaly detection applied to archaeology,'' \emph{IEEE J. Sel.
  Topics. Appl. Earth Observ.}, vol.~4, no.~4, pp. 870--876, 2011.

\bibitem{ehlers2010multi}
M.~Ehlers, S.~Klonus, P.~Johan~{\AA}strand, and P.~Rosso, ``Multi-sensor image
  fusion for pansharpening in remote sensing,'' \emph{Int. J. Image and Data
  Fusion}, vol.~1, no.~1, pp. 25--45, 2010.

\bibitem{shi2015learning}
C.~Shi, F.~Liu, L.~Li, L.~Jiao, Y.~Duan, and S.~Wang, ``Learning interpolation
  via regional map for pan-sharpening,'' \emph{IEEE Trans. Geosci. Remote
  Sens.}, vol.~53, no.~6, pp. 3417--3431, 2015.

\bibitem{liu2016spatial}
P.~Liu, L.~Xiao, J.~Zhang, and B.~Naz, ``Spatial-hessian-feature-guided
  variational model for pan-sharpening,'' \emph{IEEE Trans. Geosci. Remote
  Sens.}, vol.~54, no.~4, pp. 2235--2253, 2016.

\bibitem{du2007performance}
Q.~Du, N.~H. Younan, R.~King, and V.~P. Shah, ``On the performance evaluation
  of pan-sharpening techniques,'' \emph{IEEE Geosci. Remote Sens. Lett.},
  vol.~4, no.~4, pp. 518--522, 2007.

\bibitem{mckeown1999fusion}
D.~M. McKeown, S.~D. Cochran, S.~J. Ford, J.~C. McGlone, J.~A. Shufelt, and
  D.~A. Yocum, ``Fusion of {HYDICE} hyperspectral data with panchromatic
  imagery for cartographic feature extraction,'' \emph{IEEE Trans. Geosci.
  Remote Sens.}, vol.~37, no.~3, pp. 1261--1277, 1999.

\bibitem{licciardi2012fusion}
G.~A. Licciardi, M.~M. Khan, J.~Chanussot, A.~Montanvert, L.~Condat, and
  C.~Jutten, ``Fusion of hyperspectral and panchromatic images using
  multiresolution analysis and nonlinear {PCA} band reduction,'' \emph{EURASIP
  J. Advances in Signal processing}, vol. 2012, no.~1, pp. 1--17, 2012.

\bibitem{gomez2001wavelet}
R.~B. Gomez, A.~Jazaeri, and M.~Kafatos, ``Wavelet-based hyperspectral and
  multispectral image fusion,'' in \emph{Aerospace/Defense Sensing, Simulation,
  and Controls}.\hskip 1em plus 0.5em minus 0.4em\relax International Society
  for Optics and Photonics, 2001, pp. 36--42.

\bibitem{chen2014fusion}
Z.~Chen, H.~Pu, B.~Wang, and G.-M. Jiang, ``Fusion of hyperspectral and
  multispectral images: A novel framework based on generalization of
  pan-sharpening methods,'' \emph{IEEE Geosci. Remote Sens. Lett.}, vol.~11,
  no.~8, pp. 1418--1422, 2014.

\bibitem{de2017fusion}
V.~De~Silva, J.~Roche, and A.~Kondoz, ``Fusion of {LiDAR} and camera sensor
  data for environment sensing in driverless vehicles,'' \emph{arXiv preprint
  arXiv:1710.06230}, 2017.

\bibitem{maddern2016real}
W.~Maddern and P.~Newman, ``Real-time probabilistic fusion of sparse {3D LIDAR}
  and dense stereo,'' in \emph{IEEE/RSJ Int. Conf. Intelligent Robots and
  Systems (IROS)}, 2016, pp. 2181--2188.

\bibitem{maron1998multipleicml}
O.~Maron and A.~L. Ratan, ``Multiple-instance learning for natural scene
  classification,'' in \emph{Proc. 15th Int. Conf. Mach. Learn.}, 1998, pp.
  341--349.

\bibitem{zhou2006multi}
Z.-H. Zhou and M.-L. Zhang, ``Multi-instance multi-label learning with
  application to scene classification,'' in \emph{Proc. Adv. Neural Inf.
  Process. Syst. (NIPS)}, 2006, pp. 1609--1616.

\bibitem{ali2010human}
S.~Ali and M.~Shah, ``Human action recognition in videos using kinematic
  features and multiple instance learning,'' \emph{IEEE Trans. Pattern Anal.
  Mach. Intell.}, vol.~32, no.~2, pp. 288--303, 2010.

\bibitem{zhang2005multiple}
C.~Zhang, J.~C. Platt, and P.~A. Viola, ``Multiple instance boosting for object
  detection,'' in \emph{Proc. Adv. Neural Inf. Process. Syst. (NIPS)}, 2005,
  pp. 1417--1424.

\bibitem{babenko2009visual}
B.~Babenko, M.-H. Yang, and S.~Belongie, ``Visual tracking with online multiple
  instance learning,'' in \emph{IEEE Conf. Computer Vision and Pattern
  Recognition (CVPR)}, 2009, pp. 983--990.

\bibitem{babenko2011robust}
------, ``Robust object tracking with online multiple instance learning,''
  \emph{IEEE Trans. Pattern Anal. Mach. Intell.}, vol.~33, no.~8, pp.
  1619--1632, 2011.

\bibitem{du2015possibilistic}
X.~Du, A.~Zare, and J.~T. Cobb, ``Possibilistic context identification for
  {SAS} imagery,'' in \emph{Proc. SPIE 9454, Detection and Sensing of Mines,
  Explosive Objects, and Obscured Targets XX}, no. 94541I, May 2015.

\bibitem{torrione2009multiple}
P.~Torrione, C.~Ratto, and L.~M. Collins, ``Multiple instance and context
  dependent learning in hyperspectral data,'' in \emph{1st Workshop on
  Hyperspectral Image and Signal Processing: Evolution in Remote Sensing
  (WHISPERS)}.\hskip 1em plus 0.5em minus 0.4em\relax IEEE, 2009, pp. 1--4.

\bibitem{xu2017weakly}
L.~Xu, D.~A. Clausi, F.~Li, and A.~Wong, ``Weakly supervised classification of
  remotely sensed imagery using label constraint and edge penalty,'' \emph{IEEE
  Trans. Geosci. Remote Sens.}, vol.~55, no.~3, pp. 1424--1436, March 2017.

\bibitem{zare2017discriminative}
A.~Zare, C.~Jiao, and T.~Glenn, ``Discriminative multiple instance
  hyperspectral target characterization,'' \emph{IEEE Trans. Pattern Anal.
  Mach. Intell.}, vol.~40, no.~10, pp. 2342--2354, Oct 2018.

\bibitem{cao2017weakly}
L.~Cao, F.~Luo, L.~Chen, Y.~Sheng, H.~Wang, C.~Wang, and R.~Ji, ``Weakly
  supervised vehicle detection in satellite images via multi-instance
  discriminative learning,'' \emph{Pattern Recognition}, vol.~64, pp. 417 --
  424, 2017.

\bibitem{andrews2002support}
S.~Andrews, ``Support vector machines for mulitple-instance learning,'' in
  \emph{Ann. Conf. Neural Inf. Proc. Systems (NIPS)}, 2002.

\bibitem{choquet1954theory}
G.~Choquet, ``Theory of capacities,'' in \emph{Annales de l'institut Fourier},
  vol.~5, 1954, pp. 131--295.

\bibitem{grabisch1996application}
M.~Grabisch, ``The application of fuzzy integrals in multicriteria decision
  making,'' \emph{European J. Operational Research}, vol.~89, no.~3, pp.
  445--456, Mar. 1996.

\bibitem{grabisch1995anew}
------, ``A new algorithm for identifying fuzzy measures and its application to
  pattern recognition,'' in \emph{Int. Joint Conf. 4th IEEE Int. Conf. Fuzzy
  Systems and 2nd Int. Fuzzy Eng. Symp.}, vol.~1, Mar. 1995, pp. 145--150.

\bibitem{labreuche2003thechoquet}
C.~Labreuche and M.~Grabisch, ``The {Choquet} integral for the aggregation of
  interval scales in multicriteria decision making,'' \emph{Fuzzy Sets and
  Systems}, vol. 137, no.~1, pp. 11 -- 26, 2003.

\bibitem{mendezvazquez2008minimum}
A.~Mendez-Vazquez, P.~D. Gader, J.~M. Keller, and K.~Chamberlin, ``Minimum
  classification error training for {Choquet} integrals with applications to
  landmine detection,'' \emph{IEEE Trans. Fuzzy Systems}, vol.~16, no.~1, pp.
  225--238, Feb 2008.

\bibitem{mendezvazquez2008learning}
A.~Mendez-Vazquez and P.~Gader, ``Learning fuzzy measure parameters by logistic
  {LASSO},'' in \emph{Proc. Annual Conf. North American Fuzzy Information
  Processing Society (NAFIPS)}, May 2008, pp. 1--7.

\bibitem{gader2001recognition}
P.~D. Gader, J.~M. Keller, and B.~N. Nelson, ``Recognition technology for the
  detection of buried land mines,'' \emph{IEEE Trans. Fuzzy Systems}, vol.~9,
  no.~1, pp. 31--43, .Feb 2001.

\bibitem{du2016multiple}
X.~Du, A.~Zare, J.~Keller, and D.~Anderson, ``Multiple instance {Choquet}
  integral for classifier fusion,'' in \emph{IEEE Congr. Evolutionary
  Computation (CEC)}, Vancouver, BC, July 2016, pp. 1054--1061.

\bibitem{wang2015integration}
Q.~Wang, C.~Zheng, H.~Yu, and D.~Deng, ``Integration of heterogeneous
  classifiers based on {Choquet} fuzzy integral,'' in \emph{7th Int. Conf.
  Intelligent Human-Machine Systems and Cybernetics}, vol.~1, Aug. 2015, pp.
  543--547.

\bibitem{fodor1995characterization}
J.~Fodor, J.-L. Marichal, and M.~Roubens, ``Characterization of the ordered
  weighted averaging operators,'' \emph{IEEE Trans. Fuzzy Systems}, vol.~3,
  no.~2, pp. 236--240, May 1995.

\bibitem{maron1998phdthesis}
O.~Maron, ``Learning from ambiguity,'' AI Technical Report 1639, Massachusetts
  Institute of Technology, 1998.

\bibitem{marichal2000an}
J.-L. Marichal, ``An axiomatic approach of the discrete {Choquet} integral as a
  tool to aggregate interacting criteria,'' \emph{IEEE Trans. Fuzzy Systems},
  vol.~8, no.~6, pp. 800--807, Dec 2000.

\bibitem{Sugeno74}
M.~Sugeno, ``Theory of fuzzy integrals and its applications,'' Ph.D.
  dissertation, Tokyo Institute of Technology, 1974.

\bibitem{fitting2003beyond}
M.~Fitting and E.~Orlowska, Eds., \emph{Beyond Two: Theory and Applications of
  Multiple-Valued Logic}.\hskip 1em plus 0.5em minus 0.4em\relax Springer,
  2003.

\bibitem{keller2016fundamentals}
J.~M. Keller, D.~Liu, and D.~B. Fogel, \emph{Fundamentals of computational
  intelligence: Neural networks, fuzzy systems and evolutionary computation},
  1st~ed., ser. IEEE Press Series on Computational Intelligence.\hskip 1em plus
  0.5em minus 0.4em\relax John Wiley \& Sons, Inc., 2016.

\bibitem{mendezvazquez2008info}
A.~Mendez-Vazquez, ``Information fusion and sparsity promotion using {Choquet}
  integrals,'' Ph.D. dissertation, University of Florida, 2008.

\bibitem{nocedal2006numerical}
J.~Nocedal and S.~Wright, \emph{Numerical Optimization}.\hskip 1em plus 0.5em
  minus 0.4em\relax Springer-Verlag New York, 2006.

\bibitem{du2017multiple}
X.~Du, ``Multiple instance {Choquet} integral for multiresolution sensor
  fusion,'' Ph.D. dissertation, University of Missouri, 2017.

\bibitem{du2018multiple}
X.~{Du} and A.~{Zare}, ``{Multiple Instance Choquet Integral Classifier Fusion
  and Regression for Remote Sensing Applications},'' \emph{IEEE Trans. Geosci.
  Remote Sens.}, vol.~57, no.~5, pp. 2741--2753, May 2019.

\bibitem{johnson1994continuous}
N.~Johnson, S.~Kotz, and N.~Balakrishnan, \emph{Continuous Univariate
  Distributions}, 2nd~ed.\hskip 1em plus 0.5em minus 0.4em\relax
  Wiley-Interscience, Oct. 1994, vol.~1.

\bibitem{xiaoxiao_du_2019_2638382}
\BIBentryALTinterwordspacing
X.~Du and A.~Zare, ``{GatorSense/MIMRF: Initial Release},'' 2019. [Online].
  Available: \url{https://doi.org/10.5281/zenodo.2638382}
\BIBentrySTDinterwordspacing

\bibitem{alina_zare_2018_1186326}
A.~Zare, P.~Gader, J.~Aitken, R.~Close, G.~Tuell, T.~Glenn, D.~Dranishnikov,
  and X.~Du, ``{G}ator{S}ense/{MUUFLG}ulfport: {R}elease 01 ({V}ersion v0.1)
  [{D}ata set],'' \url{https://github.com/GatorSense/MUUFLGulfport/tree/v0.1},
  2018, {DOI:} \url{https://doi.org/10.5281/zenodo.1186326}.

\bibitem{achanta2010slictechreport}
R.~Achanta, A.~Shaji, K.~Smith, A.~Lucchi, P.~Fua, and S.~S{\"u}sstrunk,
  ``{SLIC} superpixels,'' \'{E}cole Polytechnique F\'{e}d\'{e}rale de Lausanne
  (EPFL), Lausanne, Switzerland, Tech. Rep., 2010.

\bibitem{achanta2012slic}
------, ``{SLIC} superpixels compared to state-of-the-art superpixel methods,''
  \emph{IEEE Trans. Pattern Anal. Mach. Intell.}, vol.~34, no.~11, pp.
  2274--2282, 2012.

\bibitem{osm1}
\BIBentryALTinterwordspacing
OSMcontributors, ``{Open Street Map},'' 2018. [Online]. Available:
  \url{https://www.openstreetmap.org}
\BIBentrySTDinterwordspacing

\bibitem{schalkoff1989digital}
R.~J. Schalkoff, \emph{Digital image processing and computer vision}.\hskip 1em
  plus 0.5em minus 0.4em\relax Wiley New York, 1989, vol. 286.

\bibitem{scharf1996adaptive}
L.~L. Scharf and L.~T. McWhorter, ``Adaptive matched subspace detectors and
  adaptive coherence estimators,'' in \emph{Proc. 30th Asilomar Conf. Signals
  Syst.}\hskip 1em plus 0.5em minus 0.4em\relax IEEE, Nov. 1996, pp.
  1114--1117.

\bibitem{kraut2005adaptive}
S.~Kraut, L.~L. Scharf, and R.~W. Butler, ``The adaptive coherence estimator: a
  uniformly most-powerful-invariant adaptive detection statistic,'' \emph{IEEE
  Trans. Signal Proc.}, vol.~53, no.~2, pp. 427--438, 2005.

\bibitem{pulsone2000computationally}
N.~Pulsone and M.~A. Zatman, ``A computationally efficient two-step
  implementation of the {GLRT},'' \emph{IEEE Trans. Signal Proc.}, vol.~48,
  no.~3, pp. 609--616, Mar 2000.

\bibitem{anderson2017binary}
D.~T. Anderson, M.~Islam, R.~King, N.~Younan, J.~Fairley, S.~Howington,
  F.~Petry, P.~Elmore, and A.~Zare, ``Binary fuzzy measures and {Choquet}
  integration for multi-source fusion,'' in \emph{Int. Conf. Military
  Technologies (ICMT)}.\hskip 1em plus 0.5em minus 0.4em\relax IEEE, 2017, pp.
  676--681.

\bibitem{islam2018efficient}
M.~A. Islam, D.~T. Anderson, X.~Du, T.~C. Havens, and C.~Wagner, ``Efficient
  binary fuzzy measure representation and {Choquet} integral learning,'' in
  \emph{Int. Conf. Information Processing and Management of Uncertainty in
  Knowledge-Based Systems (IPMU)}.\hskip 1em plus 0.5em minus 0.4em\relax Cham,
  Switzerland: Springer, 2018, pp. 115--126.

\bibitem{du2019ssci}
X.~Du, A.~Zare, and D.~T. Anderson, ``{Multiple Instance Choquet Integral with
  Binary Fuzzy Measures for Remote Sensing Classifier Fusion with Imprecise
  Labels},'' in \emph{{Proc. IEEE Symp. Series on Computational Intelligence
  (SSCI)}}, 2019, Accepted.

\end{thebibliography}

\end{document}